\documentclass{IEEEtran}
\usepackage{fancyhdr}

\usepackage{cite}
\usepackage{graphicx, epstopdf}
\usepackage{amsmath}
\usepackage{array}
\usepackage{epsfig}
\usepackage{amssymb}
\usepackage{subcaption}
\usepackage[ruled,vlined,lined,linesnumbered]{algorithm2e}
\usepackage{color}
\begin{document}\normalsize
\pagestyle{empty} \sloppy

\title{Preselection via Classification: A Case Study on Evolutionary Multiobjective Optimization}

\author{Jinyuan Zhang, Aimin Zhou, Ke Tang, and Guixu Zhang
\thanks{This work is supported by the National Natural Science Foundation of China under Grant No.61673180, and the Science and Technology Commission of Shanghai Municipality under Grant No.14DZ2260800.}
\thanks{J. Zhang, A. Zhou, and G. Zhang are with the Shanghai Key Laboratory of Multidimensional Information Processing, Department of Computer Science and Technology, East China Normal University, 3663 North Zhongshan Road, Shanghai, China. (email: jyzhang@stu.ecnu.edu.cn, \{amzhou,gxzhang\}@cs.ecnu.edu.cn).}
\thanks{K. Tang is with the with the USTC-Birmingham Joint Research Institute in Intelligent Computation and Its Applications (UBRI), School of Computer Science and Technology, University of Science and Technology of China, Hefei, Anhui, China. (email:ketang@ustc.edu.cn).}}

\maketitle
\begin{abstract}
In evolutionary algorithms, a preselection operator aims to select the promising offspring solutions from a candidate offspring set. It is usually based on the estimated or real objective values of the candidate offspring solutions. In a sense, the preselection can be treated as a classification procedure, which classifies the candidate offspring solutions into promising ones and unpromising ones. Following this idea, we propose a \emph{classification based preselection (CPS)} strategy for evolutionary multiobjective optimization. When applying classification based preselection, an evolutionary algorithm maintains two external populations (training data set) that consist of some selected good and bad solutions found so far; then it trains a classifier based on the training data set in each generation. Finally it uses the classifier to filter the unpromising candidate offspring solutions and choose a promising one from the generated candidate offspring set for each parent solution. In such cases, it is not necessary to estimate or evaluate the objective values of the candidate offspring solutions. The classification based preselection is applied to three state-of-the-art \emph{multiobjective evolutionary algorithms (MOEAs)} and is empirically studied on two sets of test instances. The experimental results suggest that classification based preselection can successfully improve the performance of these MOEAs.
\end{abstract}

%%%%%%%%%%%%%%%%%%%%%%%%%%%%%%%%%%%%%%%%%%%%%%%%%%%%%
% SEC1
%%%%%%%%%%%%%%%%%%%%%%%%%%%%%%%%%%%%%%%%%%%%%%%%%%%%%
\section{Introduction}
\label{sec1}

In \emph{evolutionary algorithms (EAs)}, the preselection operator is a component that has different meanings~\cite{1992Manner}. In this paper, we use the term ``preselection" to denote the process that selects the promising offspring solutions from a set of candidate offspring solutions created by the offspring generation operators before the environmental selection procedure.

A key issue in preselection is how to measure the quality of the candidate offspring solutions. The surrogate model (metamodel) is one of the major techniques for this~\cite{2005Jin,2007ZhouONK,2011Jin,2015TabatabaeiHHMS}. The surrogate model is a method to mimic the original optimization surface by finding an alternative mapping, which is computationally cheaper, from the decision variable (model input) to the dependent variable (model output) through a given training data set. Some popular surrogate models include the Kriging~\cite{1989SacksWMW,2002EmmerichGOBG}, Gaussian process~\cite{1998Mackay,2001EIbeltagyK,2008Su,2006EmmerichGN,2014LiuZG}, radial basis function~\cite{2016SunDZJ}, artificial neural networks~\cite{1997HaganDB, 1998Ratle,2000JinOS,2002JinOS}, polynomial response surfaces~\cite{1996Gunst,2009TenneA} and support vector machines~\cite{1998Burges,2009TenneA}. In the community of evolutionary computation, the surrogate model is usually used to replace the original objective function especially when the objective function evaluation is expensive~\cite{2011Jin,2002JinS,2005Jin,2015TabatabaeiHHMS,2007ZhouOLL,2011LuTY,2012LuT,2014LuTSY,2014YuanLWY,2015LiuKZ,2016NguyenZT,2016WangJJ}. In the case of preselection, the surrogate model based preselection strategies have been employed to solve different optimization problems~\cite{2002EmmerichGOBG,2003UlmerSZ,2006HoffmannH,2006OngNY,2012PilatN,2012PilatRN,2014LiaoZZ}. To reduce the time complexity of the surrogate model building, recently we proposed using a ``cheap surrogate model"~\cite{2014GongZC} to estimate the offspring quality.

In EAs, the preselection can be naturally regarded as a classification problem. More precisely, the preselection classifies the candidate offspring solutions into two categories: the selected promising ones, and the discarded unpromising ones. This indicates that what we need to know is whether a candidate offspring solution is good or bad instead of how good it is. Following this idea, a classification method was proposed for expensive optimization problem~\cite{2011LuTY}. This work has also been extended by using both classification and regression methods in preselection~\cite{2012LuT}. A major difference between the regression model based approach and the classification based approach is that the former measures the quality of the candidate offspring solutions precisely while the latter roughly classifies the candidate offspring solutions into several categories. For a candidate offspring solution, an accurate quality measurement might be useful, however in (pre)selection, a label has already offered enough information for making decisions.

It might be more suitable to use classification techniques in multiobjective optimization. The reason is that the solutions in a \emph{multiobjective evolutionary algorithm (MOEA)} are either dominated or nondominated, which forms two classes naturally. In~\cite{2014BandatuND}, the classification algorithms are used to learn Pareto dominance relations. As far as we know, we are the first ones to apply classification to the preselection of MOEAs~\cite{2015ZhangZZ, 2015ZhangAZGZ}. Very recently, a similar idea has been implemented in~\cite{2016LinZK}.

In this paper, we extend our previous work~\cite{2015ZhangZZ, 2015ZhangAZGZ} and propose a general \emph{classification based preselection (CPS)} scheme for evolutionary multiobjective optimization. The major procedure of CPS works as follows: in each generation, first a training data set (external populations) is updated by recently found solutions; secondly a classifier is built according to the training data set; thirdly for each solution, candidate offspring solutions are generated and their labels are predicted by the classifier. Finally for each solution, an offspring solution is chosen according to the predicted labels. The major differences between this paper and the previous work follow.
\begin{itemize}
\item The data preparation strategy is improved: all the nondominated solutions found so far are used to update the negative training set, and the training set contains more points. In~\cite{2015ZhangZZ, 2015ZhangAZGZ}, only the solutions in the previous generation are used to update the negative training set.

\item A general CPS strategy is proposed and applied to the three main MOEA frameworks, i.e., the Pareto domination based MOEA, the indicator based MOEA, and the decomposition based MOEA. In~\cite{2015ZhangZZ, 2015ZhangAZGZ}, CPS is applied to one framework.

\item A systematic empirical study, based on two sets of test instances, has been performed to demonstrate the advantages of CPS.

\end{itemize}

The rest of the paper is organized as follows. Section~\ref{sec2} presents the related work on evolutionary multiobjective optimization and briefly introduces the main MOEA frameworks used in the paper. Section~\ref{sec3} presents the classification based preselection in detail. The proposed CPS is systematically studied on some benchmark problems in Section~\ref{sec4}. Finally, Section~\ref{sec5} concludes this paper with some future work remarks.

%%%%%%%%%%%%%%%%%%%%%%%%%%%%%%%%%%%%%%%%%%%%%%%%%%%%%%
%% SEC2
%%%%%%%%%%%%%%%%%%%%%%%%%%%%%%%%%%%%%%%%%%%%%%%%%%%%%%
\section{Evolutionary Multiobjective Optimization}
\label{sec2}

This paper considers the following continuous \emph{multiobjective optimization problems (MOP)}).
\begin{equation}
\begin{array}{rl}
\mbox{min} & F(x) = (f_1(x), \cdots, f_m(x))^T\\
\mbox{s.t} & x \in \Pi_{i=1}^{n}[a_i, b_i]
\end{array}
\label{MOP}
\end{equation}
where $x=(x_1, \cdots, x_n)^T\in R^n$ is a decision variable vector; $\Pi_{i=1}^{n}[a_i, b_i] \subset R^n$ defines the feasible region of the search space; $a_i<b_i$ ($i=1, \cdots, n)$ are the lower and upper boundaries of the search space respectively, $f_j: R^n \rightarrow R (j=1, \cdots, m)$ is a continuous mapping, and $F(x)$ is an objective vector.

Due to the conflicting nature among the objectives in (\ref{MOP}), there usually does not exist a single solution that can optimize all the objectives at the same time. Instead, a set of tradeoff solutions, named as \emph{Pareto optimal solutions}~\cite{1999Miettinen}, are of interest. In the decision space, the set of all the Pareto optimal solutions is called the \emph{Pareto set (PS)} and in the objective space, it is called the \emph{Pareto front (PF)}.

In practice, the target of different methods is to find an approximation to the PF (PS) of (\ref{MOP}). Among the methods, EAs have become a major method to deal with MOPs due to their population based search property, which makes an MOEA be able to approximate the PF (PS) in a single run. A variety of MOEAs have been proposed in last decades~\cite{2011ZhouQLZSZ}. Most of these algorithms can be classified into three categories: the Pareto based MOEAs~\cite{1998ZitzlerT,2000CorneKO,2001ZitzlerLT,2002DebPAM}, the indicator based MOEAs~\cite{2004ZitzlerK,2007BeumeNE,2011BaderZ,2012AugerBBZ,2013PhanS}, and the decomposition based MOEAs~\cite{1995MurataI,2007ZhangL,2009LiZ}. In this paper, we demonstrate that CPS is able to improve the performance of the three kinds of MOEAs. Three algorithms- the \emph{regularity model based multiobjective estimation of distribution algorithm (RM-MEDA)}~\cite{2008ZhangZJ}, the \emph{hypervolume metric selection based EMOA (SMS-EMOA)}~\cite{2013TrautmannWB}, and the \emph{MOEA based on decomposition with multiple differential evolution mutation operators (MOEA/D-MO)}~\cite{2014LiZZ}, from the three MOEA categories respectively are used as the basic algorithms. These algorithms are briefly introduced as follows.

\subsection{RM-MEDA}

\begin{algorithm}[htbp]
\SetAlgoLined
\LinesNumbered
Initialize the population $P=\{x^1, x^2, \cdots, x^{N}\}$;\\
\While {not terminate}
{
    Build a probabilistic model for modeling the distribution of the solutions in $P$;\\ 	

    Sample a set of offspring solutions $Q$ from the probabilistic model;\\
	
	Set $P = P\cup Q$;\\
	
	Partition $P$ into fronts $P^1, P^2, \cdots$;\\
	
	Find the $k$-th front such that $$\sum_{j=1}^{k-1}|P^j|<N\mbox{ and }\sum_{j=1}^{k}|P^j|\geq N;$$
	
	\While{$\sum_{j=1}^{k}|P^j|> N$}
	{
		Calculate the density of each solution in $P^k$\;
		
		Find the solution with the worst density $x^*\in P^k$;\\
		
		Set $P^k = P^k\setminus\{x^*\}$;\\
	}
	
	Set $P = \cup_{j=1}^{k}P^j$;\\
}
\Return{$P$.}
\caption{RM-MEDA Framework}
\label{alg:rmmeda}
\end{algorithm}

The major contribution of the Pareto domination based MOEAs is the environmental selection operator. It firstly partitions the population into different clusters through the Pareto domination relation. The solutions in the same cluster are nondominated with each other, and a solution in a worse cluster will be dominated by at least one solution in a better cluster. Secondly, it assigns each solution in the same cluster a density value. The extreme solutions and solutions in sparse areas are preferable. The sorting scheme based on both the Pareto domination and density actually defines a full order over the population.

RM-MEDA~\cite{2008ZhangZJ}, shown in Algorithm~\ref{alg:rmmeda}, is a Pareto domination based MOEA. It uses probabilistic models to guide the offspring generation and utilizes a modified nondominated sorting scheme~\cite{2002DebPAM} for environmental selection. RM-MEDA works as follows: the population is initialized in \emph{Line 1}, a probabilistic model is built in \emph{Line 3}, a set of offspring solutions are sampled from the probabilistic model in \emph{Line 4}, and the population is updated through the environmental selection in \emph{Lines 5-13}. Firstly, the offspring population and the current population are merged in \emph{Line 5}. The merged population is partitioned into different clusters in~\emph{Line 6} where a cluster with a low index value is preferable. A key cluster, of which some solutions will be discarded, is found in~\emph{Line 7}. The density values of the solutions in the key cluster are calculated, and the solutions with the worst density values are discarded one by one in~\emph{Lines 8-12}. This step is the major modification where in the original version~\cite{2002DebPAM}, the key cluster is sorted by the density values and the worst ones are discarded directly. The crowding distance is used to estimate the density values.

\subsection{SMS-EMOA}
\label{sec25}

\begin{algorithm}[htbp]
\SetAlgoLined
\LinesNumbered
Initialize the population $P=\{x^1, x^2, \cdots, x^{N}\}$;\\
\While {not terminate}
{
	Generate an offspring solution $y$;\\
		
	Set $P=P\cup\{y\}$;\\
	
	Partition $P$ into fronts $P^1, P^2, \cdots, P^k$ according to fast-nondominated-sort;\\

	Set $x^* = \arg\min_{x\in P^k}\{I_H(P^k)-I_H(P^k\setminus\{x\})\}$;\\

	Set $P = P\setminus \{x^*\}$;\\
		
}
\Return{$P$.}
\caption{SMS-EMOA Framework}
\label{alg:smsemoa}
\end{algorithm}

For performance indicator based MOEAs, a performance indicator is used to measure the population quality and select a promising population to the next generation. The hypervolume indicator is widely used to do so. Some other indicators, for example the R2~\cite{2013TrautmannWB} and the fast hypervolume~\cite{2011BaderZ}, have also been utilized.

SMS-EMOA~\cite{2007BeumeNE} is a typical algorithm in this category, and its framework is shown in Algorithm~\ref{alg:smsemoa}. The algorithm uses the steady state strategy that only one offspring solution is generated in each generation. Similar to the environmental selection in the Pareto domination based approaches, it firstly partitions the combined population into different clusters in \emph{Line 5}. It then finds the solution that has the minimum contribution to the hypervolume ($I_H$) value of the merged population in \emph{Line 6}. Finally, it removes this solution to keep a fixed population size in \emph{Line 7}.

\subsection{MOEA/D-MO}
\label{sec23}

MOEA/D decomposes a MOP into a set of scalar-objective subproblems and solves them simultaneously. The optimal solution of each subproblem will hopefully be a Pareto optimal solution of the original MOP, and a set of well selected subproblems may produce a good approximation to the PS (PF). A key issue in MOEA/D is the subproblem definition. In this paper, we use the following Tchebycheff technique.
\begin{equation}
\min g(x|\lambda, z^*) = \max_{1\leq j\leq m}{\lambda_j|f_j(x)-z_j^*|}
\label{sop}
\end{equation}
where $\lambda=(\lambda_1, \cdots, \lambda_m)^T$ is a weight vector with the subproblem, and $z^*=(z_1^*, \cdots, z_m^*)^T$ is a reference point, i.e., $z_j^*$ is the minimal value of $f_j$ in the search space. For simplicity, we use $g^i(x)$ to denote the $i$-th subproblem. In most cases, two subproblems with close weight vectors also have similar optimal solutions. Based on the distances between the weight vectors, MOEA/D defines the neighborhood of a subproblem, which contains the subproblems with the nearest weight vectors. In MOEA/D, the offspring generation and solution selection are based on the concept of neighborhood.

In MOEA/D, the $i$-th ($i=1, \cdots, N$) subproblem maintains the following information:
\begin{itemize}
\item its weight vector $\lambda^i$ and its objective function $g^i$,

\item its current solution $x^i$ and the objective vector of $x^i$, i.e. $F^i = F(x^i)$, and

\item the index set of its neighboring subproblems, $B^i$.

\end{itemize}

\begin{algorithm} [htbp]
\SetAlgoLined
\LinesNumbered
Initialize a set of subproblems $(x^i, F^i, B^i, g^i)$ for $i=1$, $\cdots$, $N$, initialize the reference point $z^*$ ($j=1$, $\cdots$, $m$) $$z_j^*=\min\limits_{i=1, \cdots, N} f_j(x^i);$$

\While{not terminate}
{
    \ForEach{$i\in \{1, \cdots, N\}$}
    {
    	Set the mating pool as
	$$
	\pi = \left\{\begin{array}{ll}B^i & \mbox{ if }rand()<p_n\\\{1,\cdots,N\} & \mbox{ otherwise}\end{array}\right.
	$$
		
        Generate $M$ offspring solutions $\{y^1,\cdots, y^M\}$ by parents from the mating pool;\\
        		
	   \ForEach{$y\in \{y^1,\cdots, y^M\}$}
        {
            \For{j=1:m}
            {
    	       $$
    		      z^*_j = \left\{\begin{array}{ll}f_j(y) & \mbox{ if }f_j(y) < z^*_j \\ z^*_j & \mbox{ otherwise}\end{array}\right.
    	       $$
            }

    	   Set counter $c = 0$\;
		
    	   \ForEach{$j\in \pi$}
    	   {
        		\If{$g^j(y) < g^j(x^j)$ and $c<C$}
        		{
            		Replace $x^j$ by $y$\;
            		Set $c=c+1$\;
        		}
    	   }
    }
    }
 }

\Return{$P$.}
\caption{Framework of MOEA/D-MO}
\label{alg:moead}
\end{algorithm}

In this paper, we use MOEA/D-MO~\cite{2014LiZZ}, which modifies the offspring generation procedure of MOEA/D-DE~\cite{2009LiZ} with multiple differential evolution (DE) mutation operators, and the algorithm framework is shown in Algorithm~\ref{alg:moead}. In \emph{Line 1}, a set of $N$ subproblems and the reference ideal point are initialized. In \emph{Lines 4-5}, the mating pool is set as the neighborhood solutions with probability $p_n$ and as the population with probability $1-p_n$, and $M$ offspring solutions are generated based on the mating pool. The reference ideal point is then updated in \emph{Line 7-9}. At most $C$ neighboring solutions are replaced by each of the newly generated offspring solution in \emph{Lines 10-16}. It should be noted that the replacement is based on the subproblem objective (\ref{sop}) that is much different from the above two MOEA frameworks.

%%%%%%%%%%%%%%%%%%%%%%%%%%%%%%%%%%%%%%%%%%%%%%%%%%%%%%
%% SEC3
%%%%%%%%%%%%%%%%%%%%%%%%%%%%%%%%%%%%%%%%%%%%%%%%%%%%%%
\section{Classification based Preselection}
\label{sec3}

This section introduces the proposed CPS scheme for multiobjective optimization. To implement CPS, three procedures should be employed: the training data set preparation, the classification model building, and the choosing of promising offspring solutions. The details of these procedures follow.

\subsection{Data Preparation}
\label{sec3a}

There are two issues to be considered for preparing a proper training data set.

The first one is solution labeling. It is not suitable to directly use the objective values to label solutions in the case of multiobjective optimization. Instead, we can use the concept of Pareto domination to label the solutions. The nondominated solutions can be labelled as `positive' training points and the dominated ones can be labelled as `negative' training points.

The second one is data set update. In this paper, we focus on binary classification, which usually requires to balance the number of the positive and negative sample points. To this end, we use two external populations $P_+$ and $P_-$, denoting the positive and negative data sets respectively, to store the training data points. Furthermore, we expect that $P_+$ and $P_-$ have the same size and form a balanced training data set. The two external populations are updated in each generation. The reasons are (a) to reduce the time complexity in the model building step, (b) to represent the current population distribution, and (c) to promote preselection efficiency.

Let $Q = S(P, N)$ denote the nondominated sorting scheme in~\cite{2002DebPAM}, which selects the best $N$ solutions from $P$ and stores the selected ones in $Q$. $Q=NS(P)$ is a procedure to choose the nondominated solutions and store them in $Q$. The data preparation works as follows:
\begin{itemize}
\item In the initialization step: the two external populations are set empty, i.e., $P_+ = P_- = \emptyset$.

\item In each step: let $Q$ be the set of newly generated solutions in each generation, $Q_+$ and $Q_-$ contain the nondominated and dominated solutions in $Q$ respectively. $P_+$ and $P_-$ are updated as
$$P_+ = S\left(NS(P_+\cup Q_+), 5N\right)$$
and
$$P_- = S\left(P_-\cup Q_-\cup P_+\cup Q_+ \setminus NS(P_+\cup Q_+), 5N\right).$$
\end{itemize}

Different to our previous work in~\cite{2015ZhangZZ, 2015ZhangAZGZ}, this new approach uses only nondominated solutions to update $P_+$, and all the dominated solutions found so far to update $P_-$. Furthermore, the external population sizes are expected at most to be $5$ times of the population size $N$. It should be noted that although it is expected that $|P_+|=|P_-|=5N$, in early states or on some complicated problems when it is hard to generate nondominated solutions, $|P_+|$ might be less than $5N$. The influence of the size of the training data set shall be empirically studied shortly in Section IV.

\subsection{Classification Model}
\label{sec3b}

Let $P_+=\{x\}$ and $P_-=\{x\}$ denote the positive and negative training sets respectively where $x$ is the feature vector, i.e., the decision variable vector in our case. $+1$ and $-1$ are the labels of the feature vectors in $P_+$ and $P_-$ respectively. A classification procedure aims to find an approximated relationship between a feature vector $x$ and a label $l\in\{+1,-1\}$
$$
l = \hat{C}(x)
$$
to replace the real relationship $l = C(x)$ based on the given training set.

There exists a variety of classification methods~\cite{2006Bishop}. In this paper, we focus on the following \emph{K-nearest neighbor (KNN)}~\cite{1967CoverH} model.
$$
KNN(x) =\left\{\begin{array}{ll}+1&\mbox{if }\sum\limits_{i=1}^{K}C(x^i)\geq0\\-1&\mbox{otherwise}\end{array}\right.
$$
where $K$ is an odd number, $x^i$ denotes the $i$-th closest feature vector, according to the Euclidean distance, in $P_+\cup P_-$, and $C(x^i)$ is the actual label of $x^i$.

\subsection{Offspring Selection}
\label{sec3c}

Each parent solution $x$ generates $M$ candidate offspring solutions: $Y=\{y_1, y_2, \cdots, y_M\}$. In preselection, the qualities of these candidate solutions are estimated through the classification model, and a promising one will be selected for the real function evaluation.

In this paper, we randomly choose an offspring solution with a predicted label $1$. It works as follows:
\begin{enumerate}
\item Set $Y^*=\{y\in Y| KNN(y) = 1\}$,

\item Reset it as $Y^*=Y$ if $Y^*$ is empty,

\item Randomly choose a solution $y^*\in Y^*$ as the offspring solution.
\end{enumerate}

It should be noted that the procedure may choose an unpromising one when no candidate solutions are labeled as $1$.

\subsection{CPS based MOEA}
\label{sec3d}

\begin{figure}[htbp]
\centering
\includegraphics[width=0.9\columnwidth]{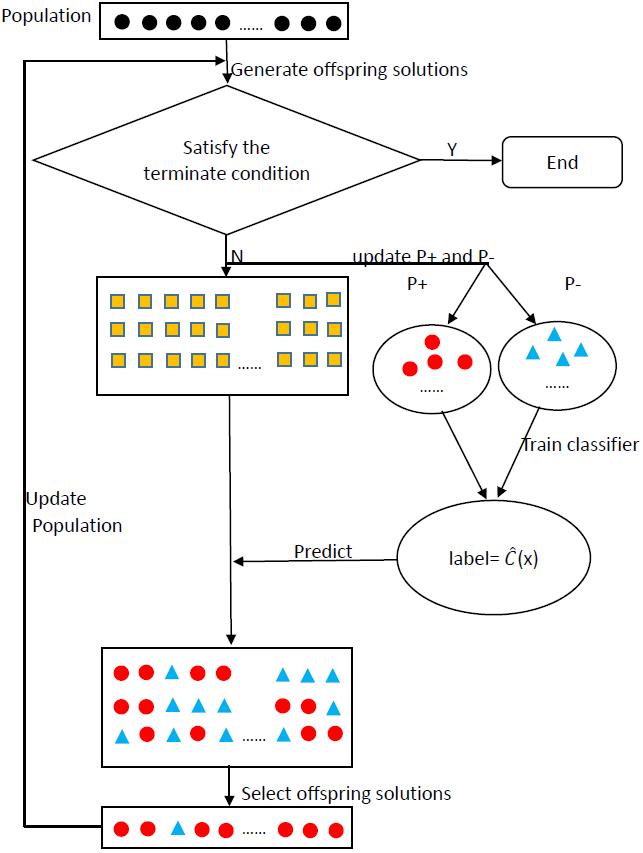}
\caption{An illustration of the flowchart of CPS based MOEA}
\label{fig:flow}
\end{figure}

Fig.~\ref{fig:flow} illustrates a general flowchart of CPS based MOEA.

To apply CPS in MOEAs, the following steps should be added or modified in the original MOEAs.
\begin{itemize}
\item The two external populations $P_+$ and $P_-$ should be set to empty in the initialization step, i.e., \emph{Line 1} in Algorithms~\ref{alg:rmmeda}-\ref{alg:moead}.

\item In each step, the two external populations should be firstly updated as in Section~\ref{sec3a}, i.e., below \emph{Line 2} in Algorithms~\ref{alg:rmmeda}-\ref{alg:moead}.

\item Following the external population update, a classifier is constructed as in Section~\ref{sec3b}, i.e., before \emph{Line 3} in Algorithms~\ref{alg:rmmeda}-\ref{alg:moead}. It should be noted that in this paper, the KNN model is a nonparametric model and this step could be overlooked.

\item The offspring generation step, i.e., \emph{Line 4} in Algorithm~\ref{alg:rmmeda}, and \emph{Line 3} in Algorithm~\ref{alg:smsemoa} should be modified and a set of $M$ candidate offspring solutions are generated for each parent by repeating the offspring generation procedure for $M$ times. A final offspring solution is then chosen  by the selection strategy in Section~\ref{sec3c}. In Algorithm~\ref{alg:moead}, the offspring generation in \emph{Line 5} should not be changed. Then an offspring solution is chosen and only this offspring solution is used to update the reference idea point and to update the neighborhood in \emph{Lines 7-16}.
\end{itemize}

%%%%%%%%%%%%%%%%%%%%%%%%%%%%%%%%%%%%%%%%%%%%%%%%%%%%%%
%% SEC4
%%%%%%%%%%%%%%%%%%%%%%%%%%%%%%%%%%%%%%%%%%%%%%%%%%%%%%

\section{Experimental Study}
\label{sec4}

\subsection{Experimental Settings}
\label{sec41}

In this section, we apply CPS into the variants of the three major MOEA frameworks. The three chosen algorithms are RM-MEDA~\cite{2008ZhangZJ}, SMS-EMOA~\cite{2013TrautmannWB}, and MOEA/D-MO~\cite{2014LiZZ}, respectively. And the three CPS based algorithms are denoted as RM-MEDA-CPS, SMS-EMOA-CPS, and MOEA/D-MO-CPS, respectively. These algorithms are applied to $10$ test instances, ZZJ1-ZZJ10, from~\cite{2008ZhangZJ} in the experiments.

The parameter settings are as follows:
\begin{itemize}
\item The number of decision variables is $n=30$ for all the test instances.

\item The algorithms are executed $30$ times independently on each instance and stop after $20,000$ \emph{function evaluations (FEs)} on ZZJ1, ZZJ2, ZZJ5, and ZZJ6, $40,000$ FEs on ZZJ4 and ZZJ8, as well as $100,000$ FEs on ZZJ3, ZZJ7, ZZJ9, and ZZJ10.

\item The population size is set as $N=100$ on ZZJ1, ZZJ2, ZZJ5, ZZJ6, and $200$ on ZZJ3, ZZJ4 and ZZJ7-ZZJ10 respectively.

\item In CPS, the number of nearest points used in KNN is $K=3$.
\end{itemize}

The other parameter settings of algorithms are referred to related work ~\cite{2008ZhangZJ,2013TrautmannWB,2014LiZZ}. All of the algorithms are implemented in Matlab and are executed on the same computer.

\subsection{Performance Metrics}

We use the \emph{Inverted Generational Distance ($IGD$)} metric~\cite{2005CoelloC,2005ZhouZJTO,2012SchutzeELC} and $I^-_H$ metric~\cite{2003ZitzlerTLFF} to assess the performance of the algorithms in the experimental study.

Let $P^{*}$ be a set of Pareto optimal points to represent the true PF, and $P$ be the set of nondominated solutions found by an algorithm. The $IGD$ and $I^-_H$ metrics are briefly defined as follows:
$$
IGD(P^{*}, P) = \frac{\sum_{x \in P^*}d(x, P)}{|P^{*}|}
$$
where $d(x, P)$ denotes the minimum Euclidean distance between $x$ and any point in $P$, and $|P^{*}|$ denotes the cardinality of $P^*$.

$$
I^{-}_H(P^{*}, P) = I_H(P^{*},z^{*})-I_H(P,z^{*})
$$
where $z^{*}$ is a reference point, and $I_H(P,z^{*})$ denotes the hypervolume of the space covered by the set $P$ and the reference point $z^{*}$.

Both metrics can measure the diversity and convergence of the obtained set $P$. To have a small metric value, the obtained set should be close to the PF and be well-distributed. In our experiments, $10,000$ evenly distributed points in PF are generated as the $P^*$. In the experiments, $z^{*}$ is set to $(1.2,1.2)$ and $(1.2,1.2,1.2)$ for the bi-objective and tri-objective problems respectively.

In order to get statistically conclusions, the Wilcoxon's rank sum test at a $5\%$ significance level is employed to compare the IGD and $I^-_H$ metric values obtained by different algorithms. In the table, $\sim$, $+$, and $-$ denote that the results obtained by the CPS based version are similar to, better than, or worse than that obtained by the original version.

\subsection{Comparison Study}

\subsubsection{RM-MEDA-CPS vs. RM-MEDA}

\begin{table*}[htbp]
\scriptsize
\centering \caption{The statistical results of IGD and $I^-_H$ metric values obtained by RM-MEDA-CPS and RM-MEDA on ZZJ1-ZZJ10}\label{tab:rmcompare}
\begin{tabular}{l|c|cccc|cccc}\hline\hline
instance&\multicolumn{1}{c|}{metric}&\multicolumn{4}{c}{RM-MEDA-CPS}&\multicolumn{4}{|c}{RM-MEDA}\\
&&mean&std.&min&max&mean&std.&min&max\\\hline
$ZZJ1$	&$IGD$	&4.19e-03(+)	&8.61e-05	&4.07e-03	&4.42e-03	&4.28e-03	&1.11e-04	&4.10e-03	&4.49e-03	\\
&$I^-_H$&5.73e-03(+)	&2.74e-04	&5.36e-03	&6.45e-03	&6.03e-03	&3.72e-04	&5.39e-03	&6.76e-03	\\
$ZZJ2$	&$IGD$	&4.13e-03($\sim$)	&8.41e-05	&3.97e-03	&4.29e-03	&4.15e-03	&7.72e-05	&3.98e-03	&4.32e-03	\\
&$I^-_H$&5.78e-03($\sim$)	&4.02e-04	&5.08e-03	&6.97e-03	&5.93e-03	&3.72e-04	&5.28e-03	&6.77e-03	\\
$ZZJ3$	&$IGD$	&7.33e-03(+)	&1.44e-03	&5.21e-03	&1.21e-02	&1.02e-02	&3.26e-03	&5.89e-03	&2.52e-02	\\
&$I^-_H$&9.80e-03(+)	&1.79e-03	&7.16e-03	&1.56e-02	&1.35e-02	&3.91e-03	&8.10e-03	&3.15e-02	\\
$ZZJ4$	&$IGD$	&4.68e-02(+)	&9.16e-04	&4.43e-02	&4.87e-02	&4.80e-02	&1.06e-03	&4.60e-02	&5.06e-02	\\
&$I^-_H$&5.75e-02(+)	&2.19e-03	&5.19e-02	&6.11e-02	&6.13e-02	&3.28e-03	&5.55e-02	&6.80e-02	\\
$ZZJ5$	&$IGD$	&4.99e-03(+)	&4.38e-04	&4.62e-03	&7.11e-03	&5.16e-03	&5.04e-04	&4.62e-03	&7.36e-03	\\
&$I^-_H$&8.06e-03(+)	&1.26e-03	&6.91e-03	&1.38e-02	&8.64e-03	&1.46e-03	&7.30e-03	&1.49e-02	\\
$ZZJ6$	&$IGD$	&5.97e-03(+)	&5.99e-04	&5.01e-03	&8.14e-03	&8.76e-03	&3.29e-03	&5.71e-03	&2.17e-02	\\
&$I^-_H$&1.24e-02(+)	&1.93e-03	&8.81e-03	&1.76e-02	&2.31e-02	&1.12e-02	&1.30e-02	&6.24e-02	\\
$ZZJ7$	&$IGD$	&8.75e-02(+)	&1.15e-02	&3.11e-02	&9.72e-02	&1.08e-01	&1.71e-02	&3.98e-02	&1.23e-01	\\
&$I^-_H$&1.11e-01(+)	&1.09e-02	&6.13e-02	&1.23e-01	&1.37e-01	&1.55e-02	&7.45e-02	&1.55e-01	\\
$ZZJ8$	&$IGD$	&5.79e-02(+)	&2.26e-03	&5.45e-02	&6.60e-02	&6.31e-02	&4.09e-03	&5.84e-02	&7.55e-02	\\
&$I^-_H$&8.93e-02(+)	&4.26e-03	&8.28e-02	&1.01e-01	&1.01e-01	&8.54e-03	&8.94e-02	&1.27e-01	\\
$ZZJ9$	&$IGD$	&4.62e-03($\sim$)	&1.29e-03	&3.20e-03	&7.57e-03	&4.77e-03	&2.82e-03	&3.36e-03	&1.84e-02	\\
&$I^-_H$&8.25e-03($\sim$)	&2.31e-03	&5.70e-03	&1.34e-02	&8.50e-03	&4.73e-03	&6.02e-03	&3.10e-02	\\
$ZZJ10$	&$IGD$	&1.34e+02($\sim$)	&8.77e+00	&1.03e+02	&1.48e+02	&1.32e+02	&7.01e+00	&1.18e+02	&1.44e+02	\\
&$I^-_H$&1.11e+00($\sim$)	&2.26e-16	&1.11e+00	&1.11e+00	&1.11e+00	&2.26e-16	&1.11e+00	&1.11e+00	\\
\hline
$+/-/\sim$ &$IGD$	&7/0/3	&&&&&&\\
$+/-/\sim$ &$I^-_H$ &7/0/3		&&&&&&\\
\hline\hline
\end{tabular}
\end{table*}

\begin{figure*}[htbp]
\centering
\begin{subfigure}[t]{0.38\columnwidth}
    \includegraphics[ width=\columnwidth]{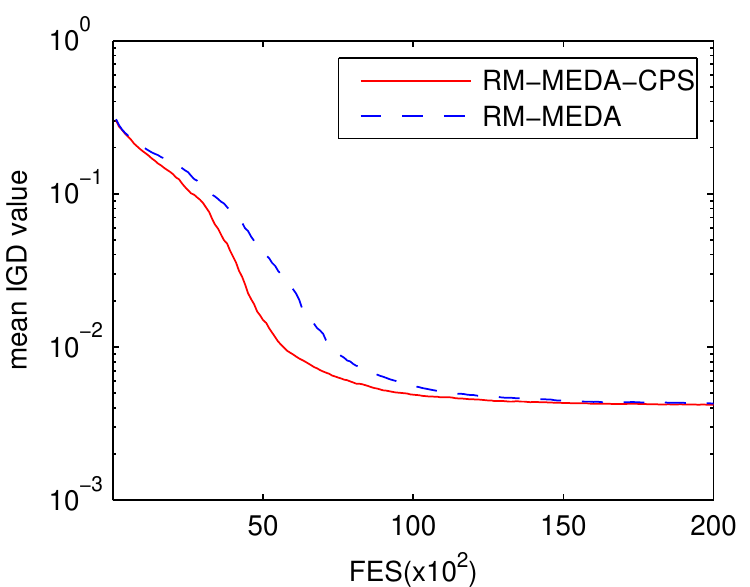}
    \subcaption{ZZJ1}
\end{subfigure}
\begin{subfigure}[t]{0.38\columnwidth}
    \includegraphics[ width=\columnwidth]{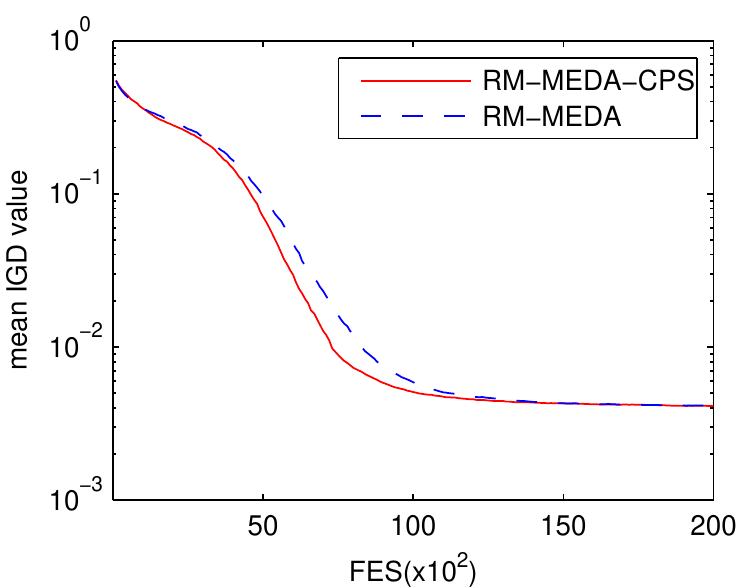}
    \subcaption{ZZJ2}
\end{subfigure}
\begin{subfigure}[t]{0.38\columnwidth}
    \includegraphics[ width=\columnwidth]{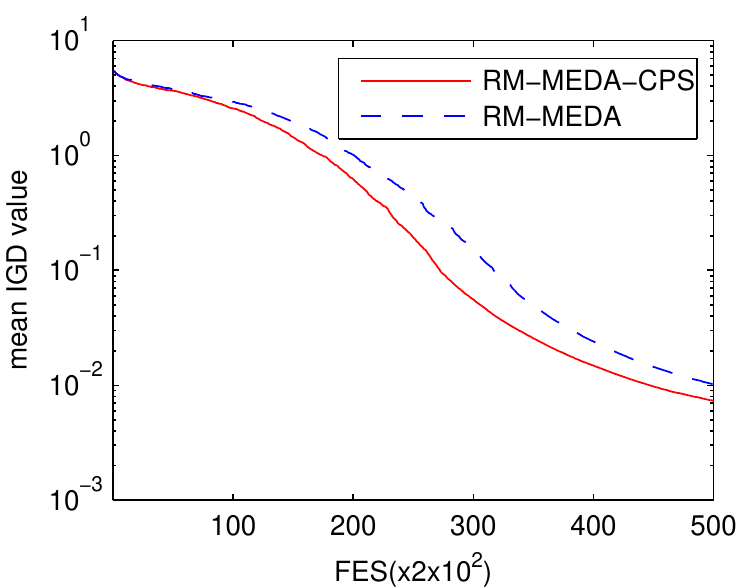}
    \subcaption{ZZJ3}
\end{subfigure}
\begin{subfigure}[t]{0.38\columnwidth}
    \includegraphics[ width=\columnwidth]{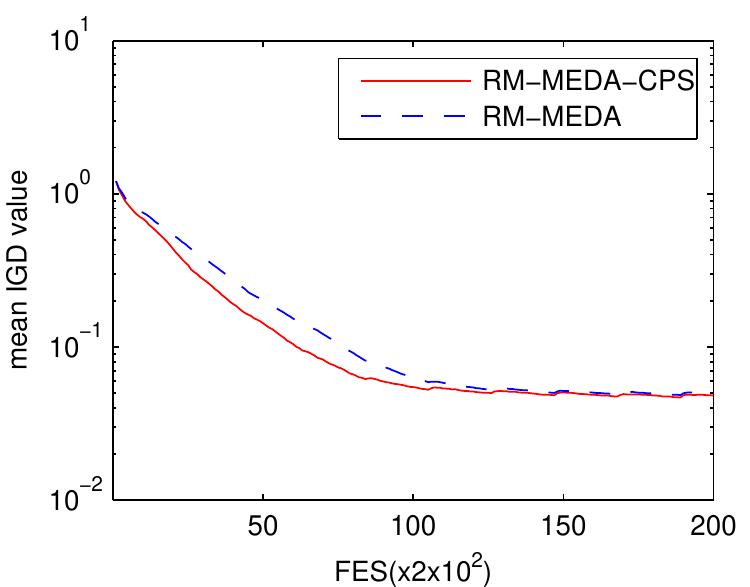}
    \subcaption{ZZJ4}
\end{subfigure}
\begin{subfigure}[t]{0.38\columnwidth}
    \includegraphics[ width=\columnwidth]{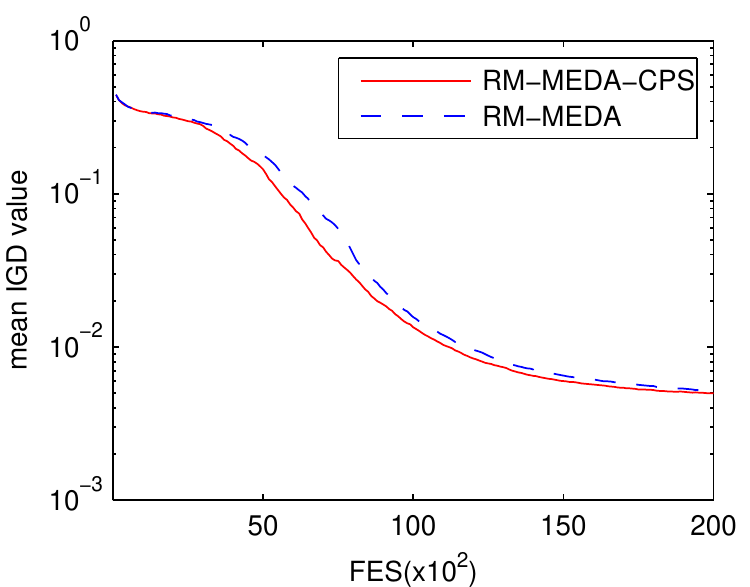}
    \subcaption{ZZJ5}
\end{subfigure}
\begin{subfigure}[t]{0.38\columnwidth}
    \includegraphics[ width=\columnwidth]{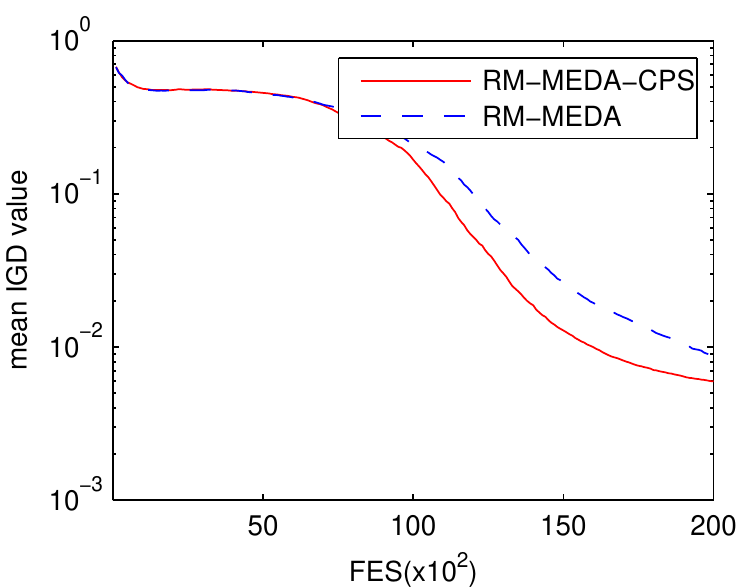}
    \subcaption{ZZJ6}
\end{subfigure}
\begin{subfigure}[t]{0.38\columnwidth}
    \includegraphics[ width=\columnwidth]{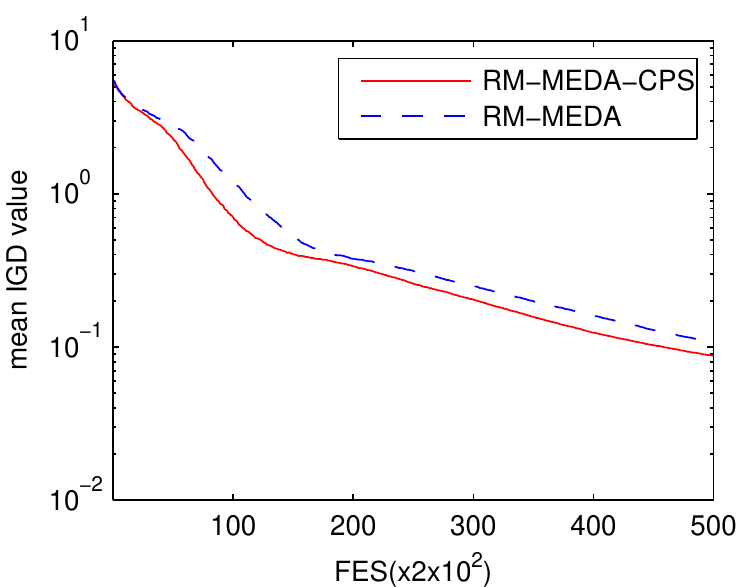}
    \subcaption{ZZJ7}
\end{subfigure}
\begin{subfigure}[t]{0.38\columnwidth}
    \includegraphics[ width=\columnwidth]{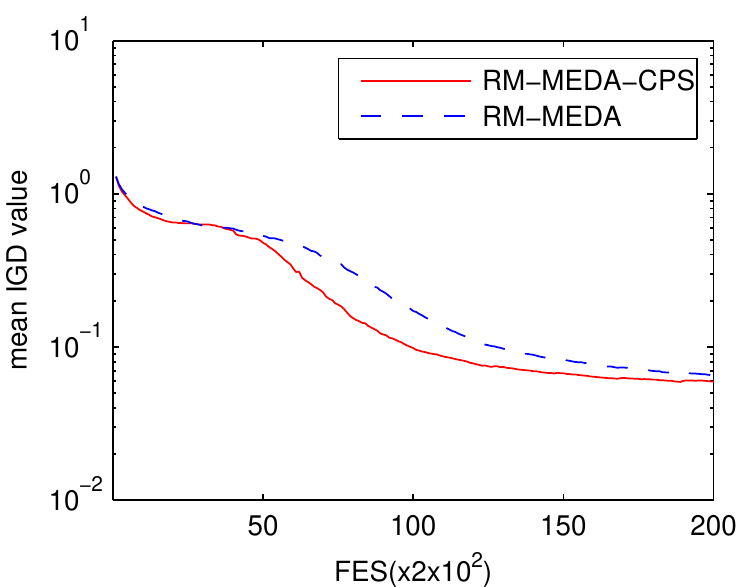}
    \subcaption{ZZJ8}
\end{subfigure}
\begin{subfigure}[t]{0.38\columnwidth}
    \includegraphics[ width=\columnwidth]{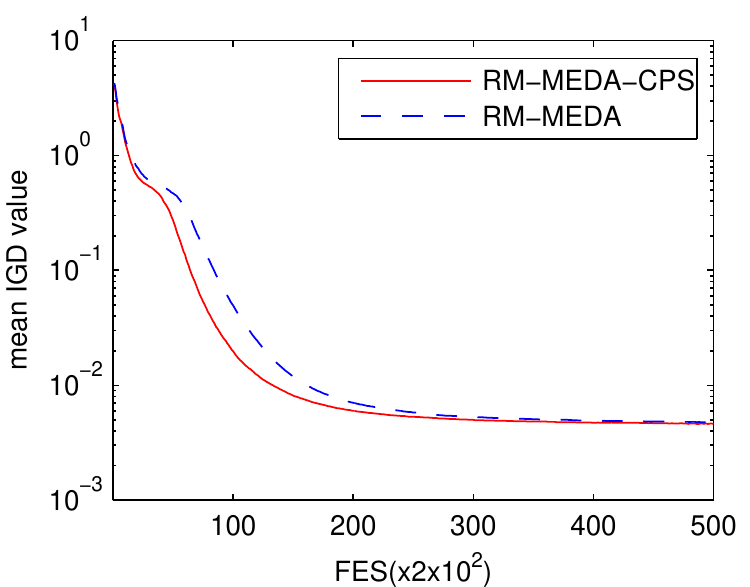}
    \subcaption{ZZJ9}
\end{subfigure}
\begin{subfigure}[t]{0.38\columnwidth}
    \includegraphics[ width=\columnwidth]{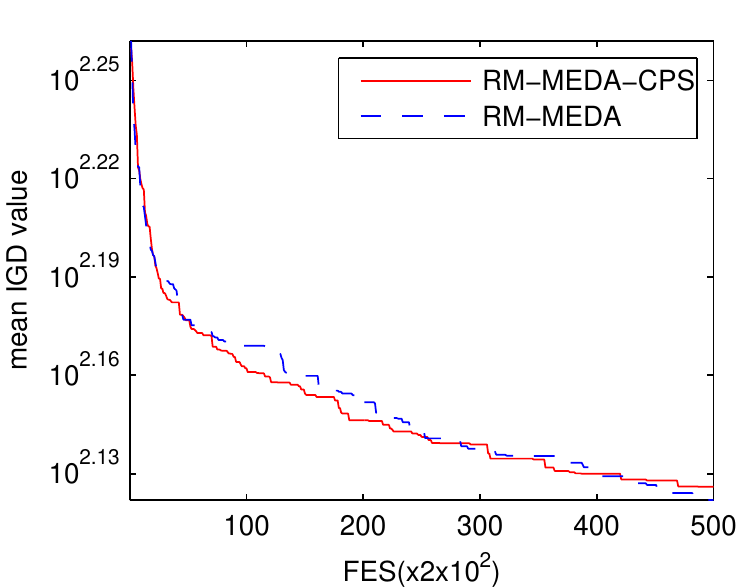}
    \subcaption{ZZJ10}
\end{subfigure}
\caption {The mean IGD values versus FEs obtained by RM-MEDA-CPS and RM-MEDA over 30 runs}
\label{fig:rmcompare}
\end{figure*}

\begin{figure*}[htbp]
\centering
\begin{subfigure}[t]{0.38\columnwidth}
    \includegraphics[ width=\columnwidth]{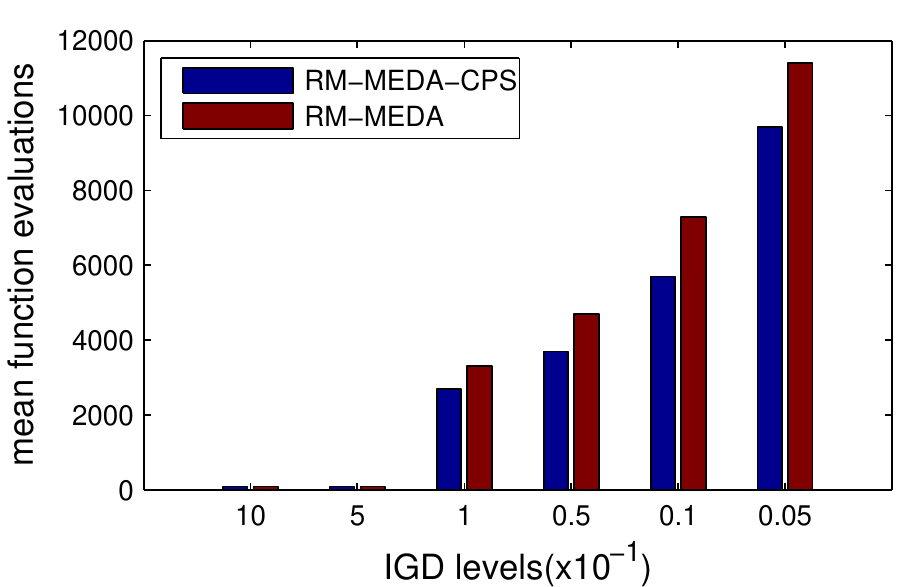}
    \subcaption{ZZJ1}
\end{subfigure}
\begin{subfigure}[t]{0.38\columnwidth}
    \includegraphics[ width=\columnwidth]{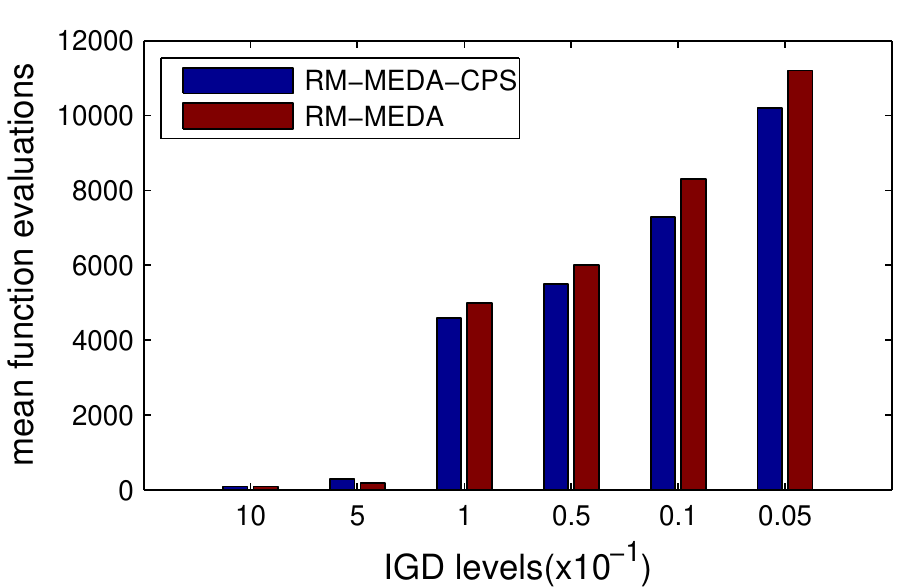}
    \subcaption{ZZJ2}
\end{subfigure}
\begin{subfigure}[t]{0.38\columnwidth}
    \includegraphics[ width=\columnwidth]{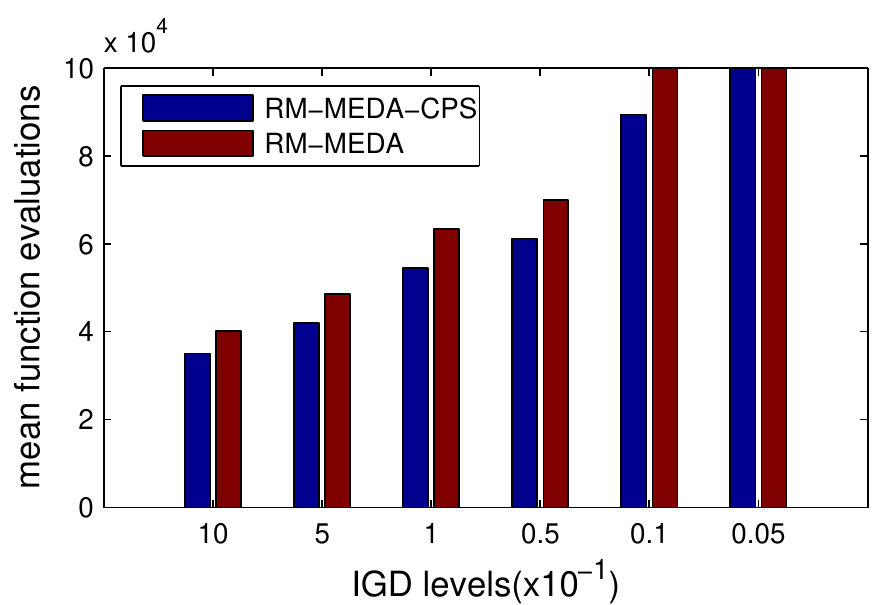}
    \subcaption{ZZJ3}
\end{subfigure}
\begin{subfigure}[t]{0.38\columnwidth}
    \includegraphics[ width=\columnwidth]{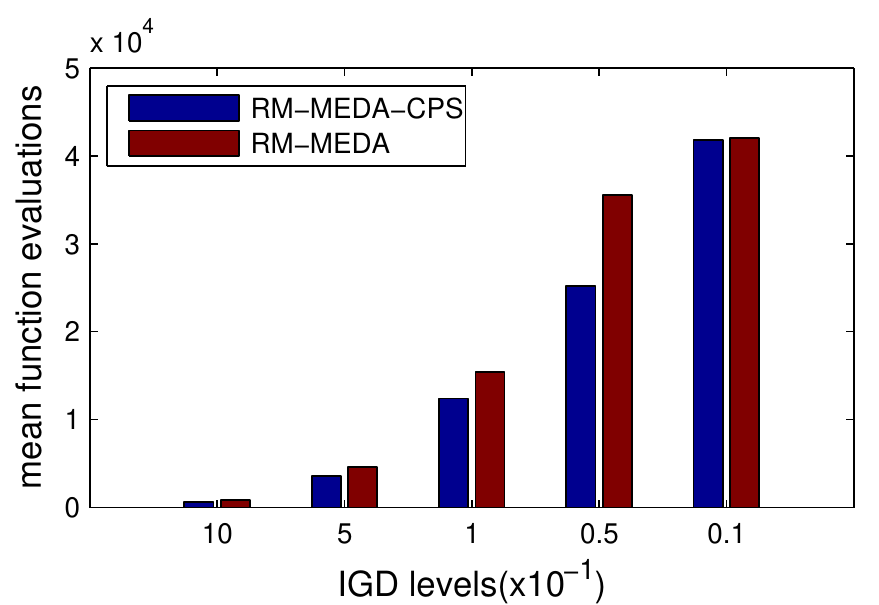}
    \subcaption{ZZJ4}
\end{subfigure}
\begin{subfigure}[t]{0.38\columnwidth}
    \includegraphics[ width=\columnwidth]{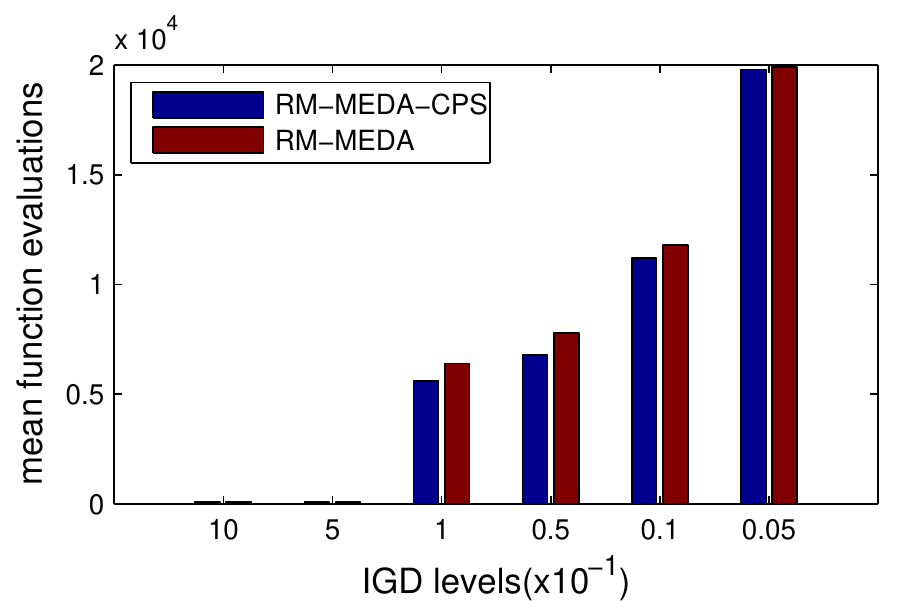}
    \subcaption{ZZJ5}
\end{subfigure}
\begin{subfigure}[t]{0.38\columnwidth}
    \includegraphics[ width=\columnwidth]{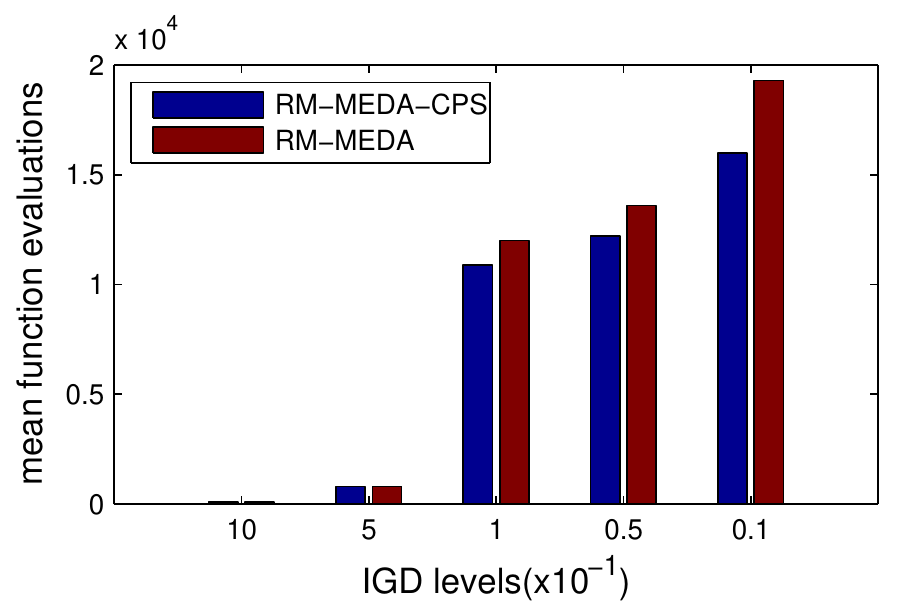}
    \subcaption{ZZJ6}
\end{subfigure}
\begin{subfigure}[t]{0.38\columnwidth}
    \includegraphics[ width=\columnwidth]{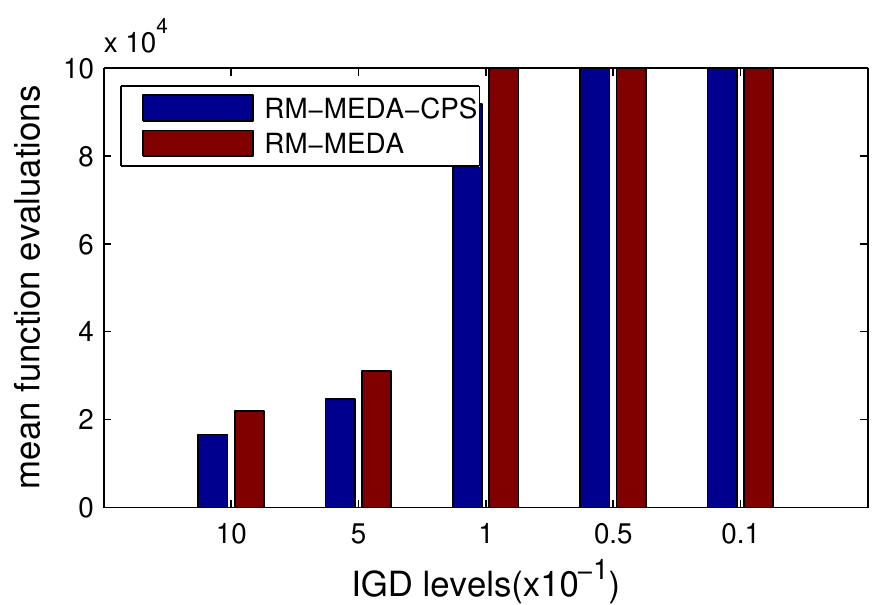}
    \subcaption{ZZJ7}
\end{subfigure}
\begin{subfigure}[t]{0.38\columnwidth}
    \includegraphics[ width=\columnwidth]{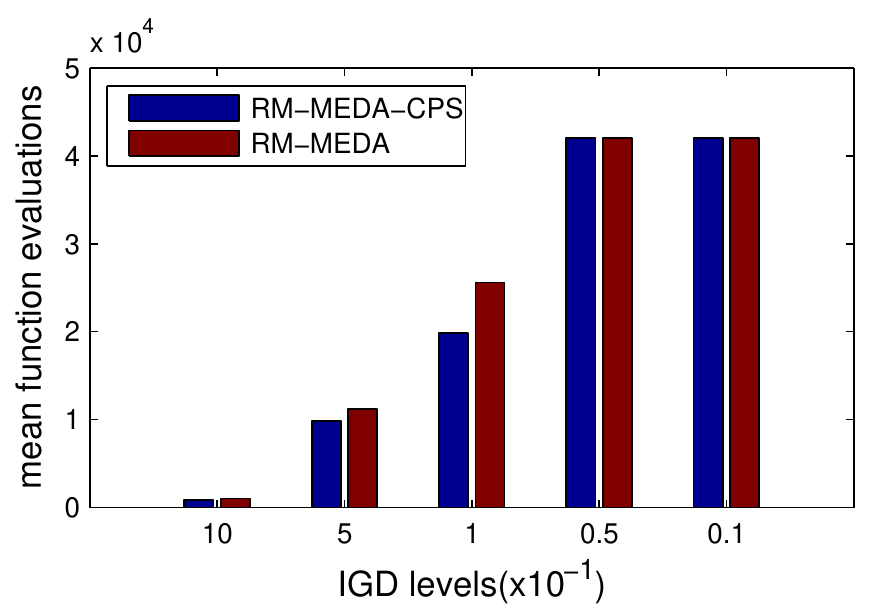}
    \subcaption{ZZJ8}
\end{subfigure}
\begin{subfigure}[t]{0.38\columnwidth}
    \includegraphics[ width=\columnwidth]{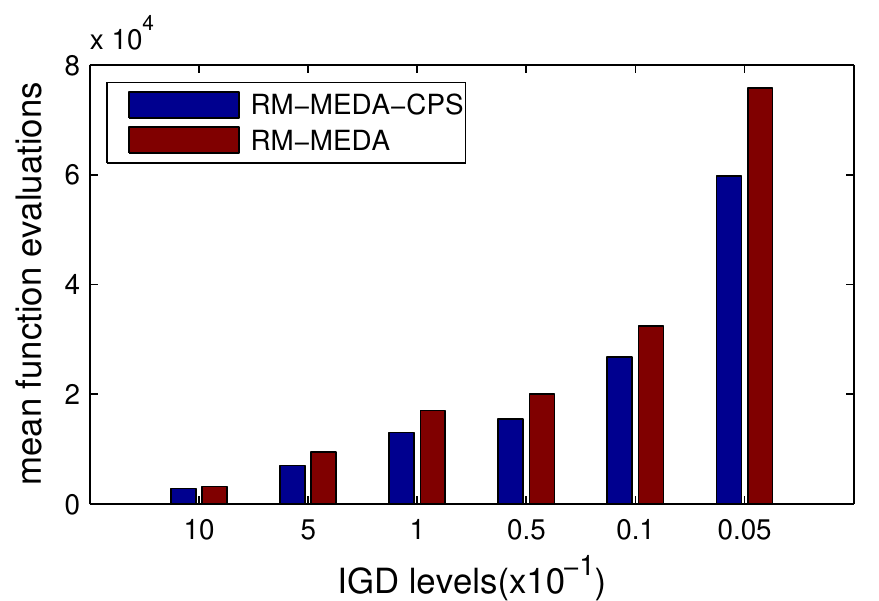}
    \subcaption{ZZJ9}
\end{subfigure}
\begin{subfigure}[t]{0.38\columnwidth}
    \includegraphics[ width=\columnwidth]{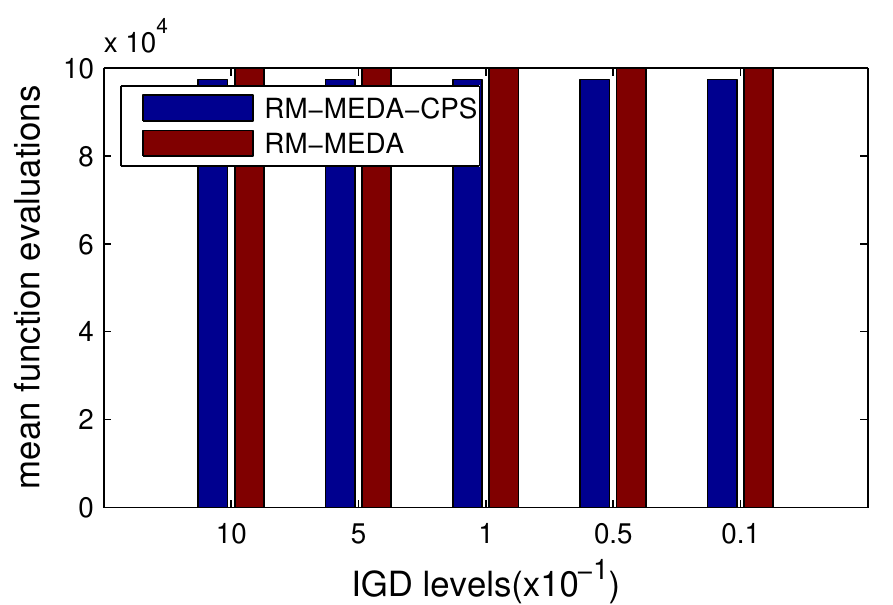}
    \subcaption{ZZJ10}
\end{subfigure}
\caption {The mean FEs required by RM-MEDA-CPS and RM-MEDA to obtain different IGD values over 30 runs}
\label{fig:rmcomparebar}
\end{figure*}

\begin{figure*}[htbp]
\centering
\begin{subfigure}[t]{0.64\columnwidth}
    \includegraphics[ width=0.50\columnwidth,height=2.3cm]{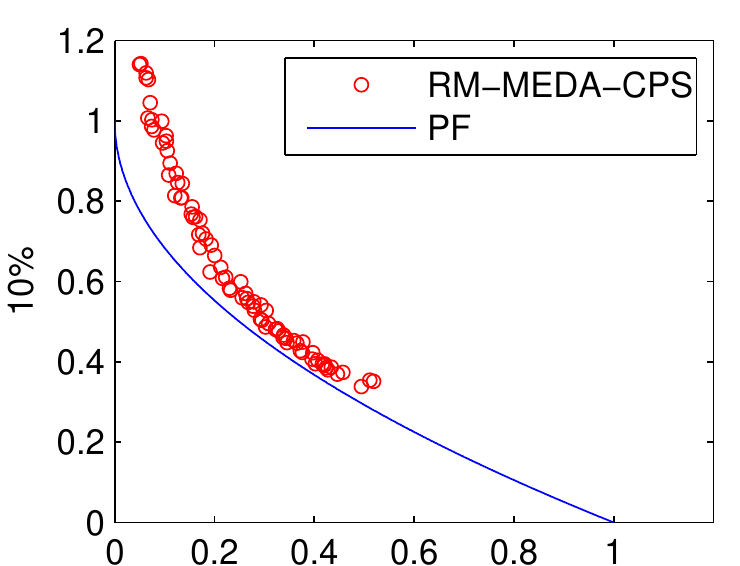}
    \includegraphics[ width=0.46\columnwidth,height=2.3cm]{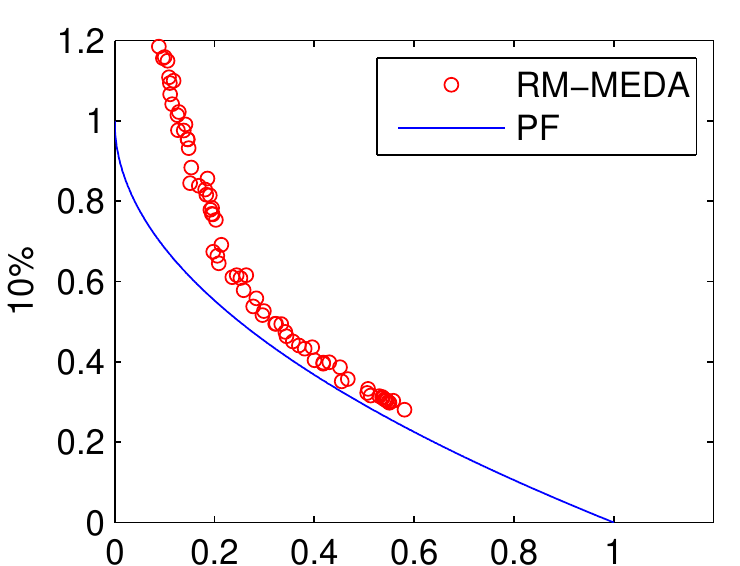}

    \includegraphics[ width=0.50\columnwidth,height=2.3cm]{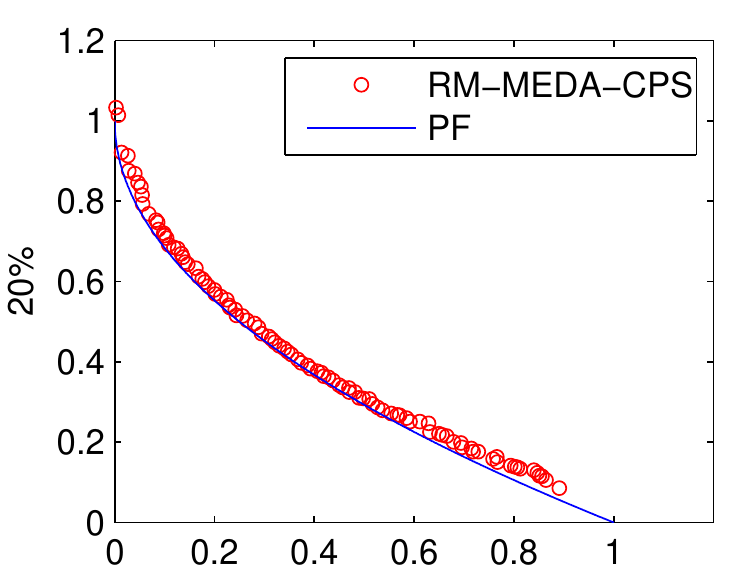}
    \includegraphics[ width=0.46\columnwidth,height=2.3cm]{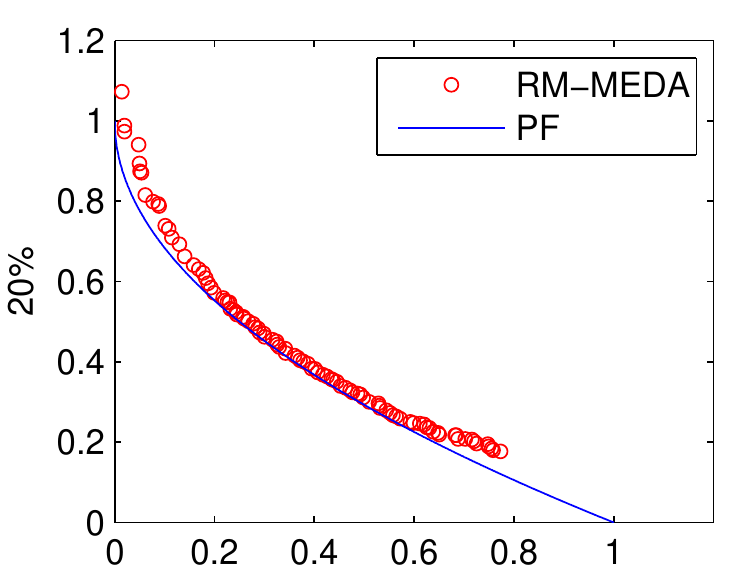}

    \includegraphics[ width=0.50\columnwidth,height=2.3cm]{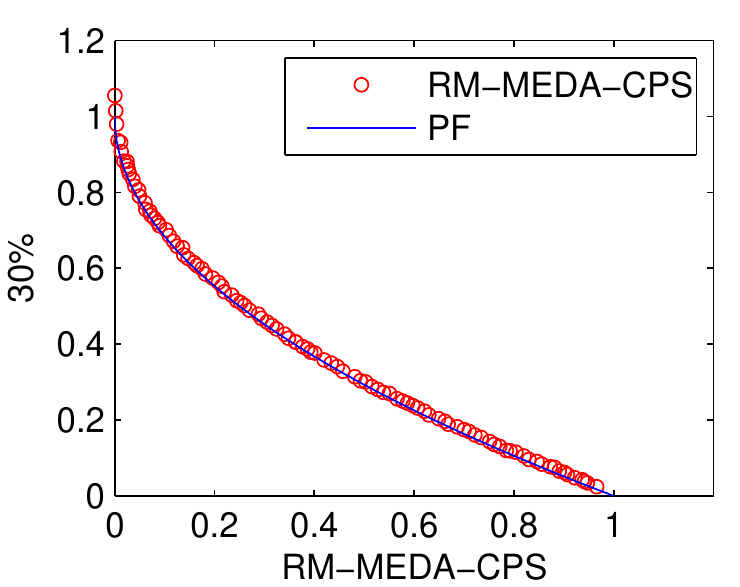}
    \includegraphics[ width=0.46\columnwidth,height=2.3cm]{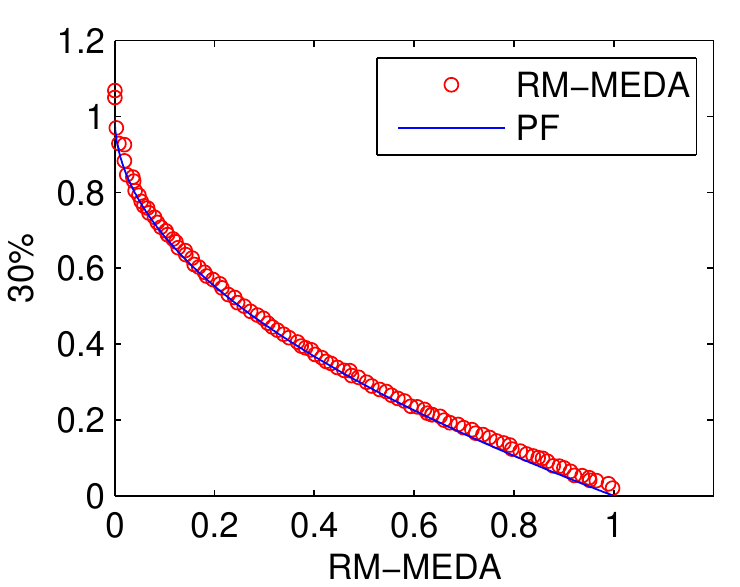}
    \subcaption{ZZJ1}
\end{subfigure}
\begin{subfigure}[t]{0.64\columnwidth}
    \includegraphics[ width=0.50\columnwidth,height=2.3cm]{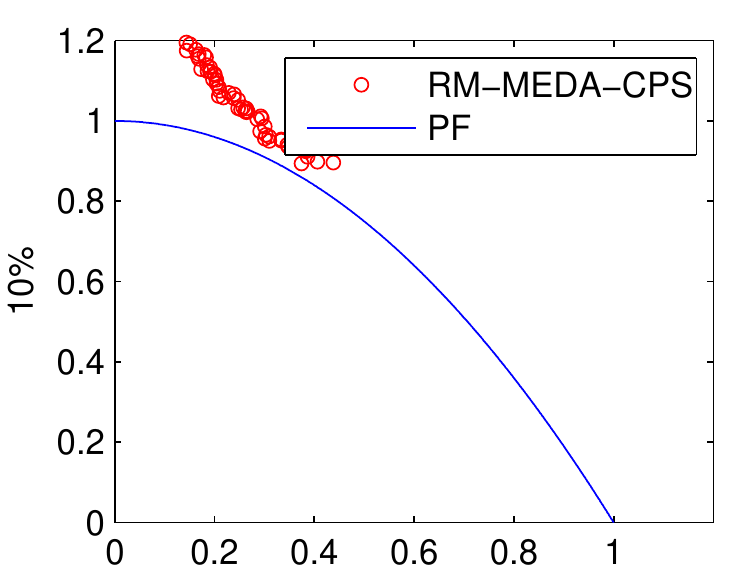}
    \includegraphics[ width=0.46\columnwidth,height=2.3cm]{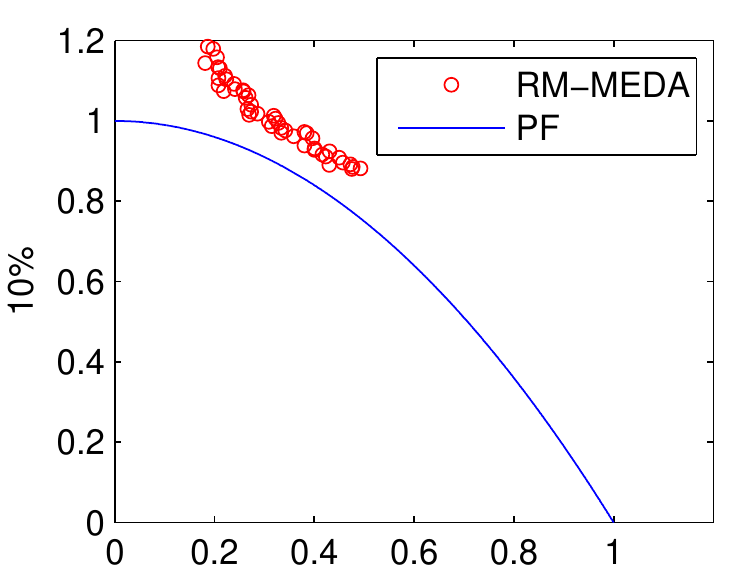}

    \includegraphics[ width=0.50\columnwidth,height=2.3cm]{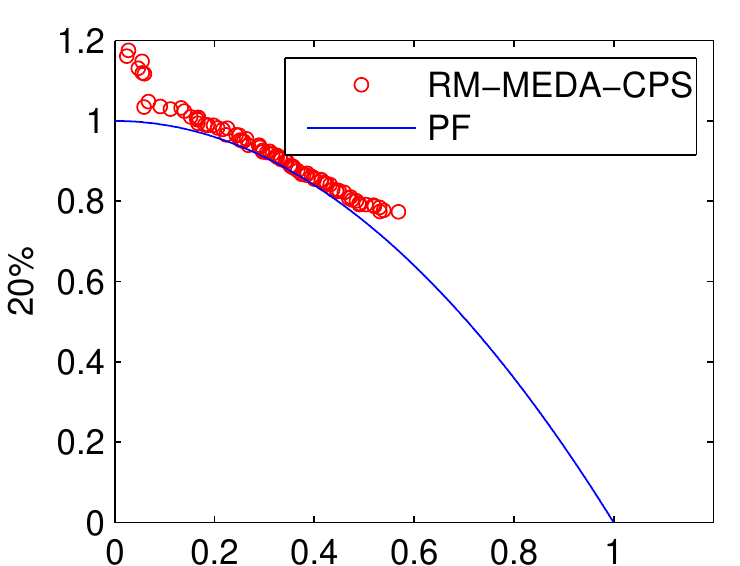}
    \includegraphics[ width=0.46\columnwidth,height=2.3cm]{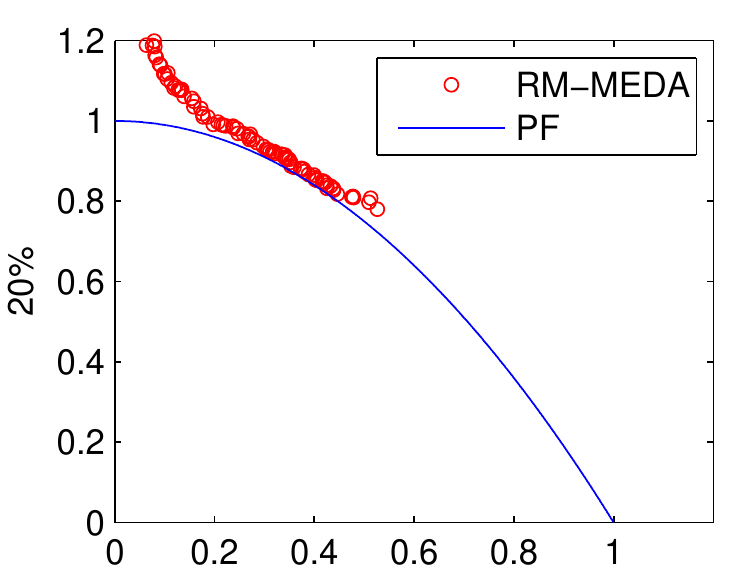}

    \includegraphics[ width=0.50\columnwidth,height=2.3cm]{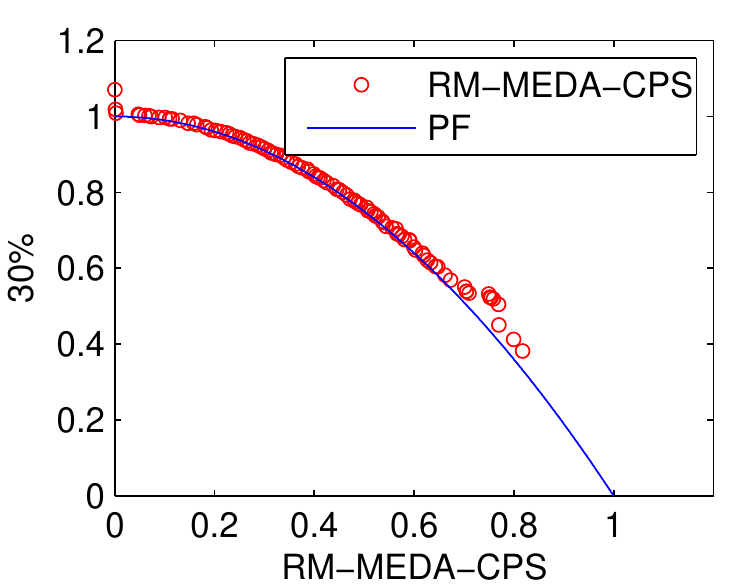}
    \includegraphics[ width=0.46\columnwidth,height=2.3cm]{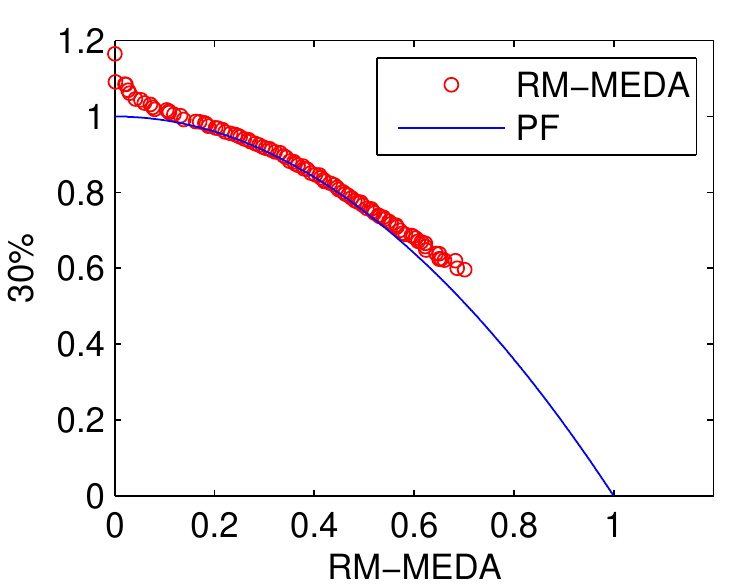}
    \subcaption{ZZJ2}
\end{subfigure}
\begin{subfigure}[t]{0.64\columnwidth}
    \includegraphics[ width=0.50\columnwidth,height=2.3cm]{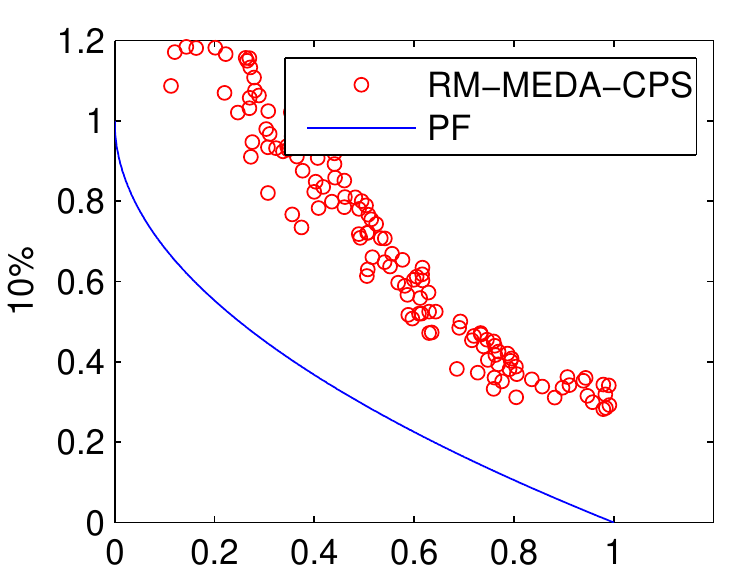}
    \includegraphics[ width=0.46\columnwidth,height=2.3cm]{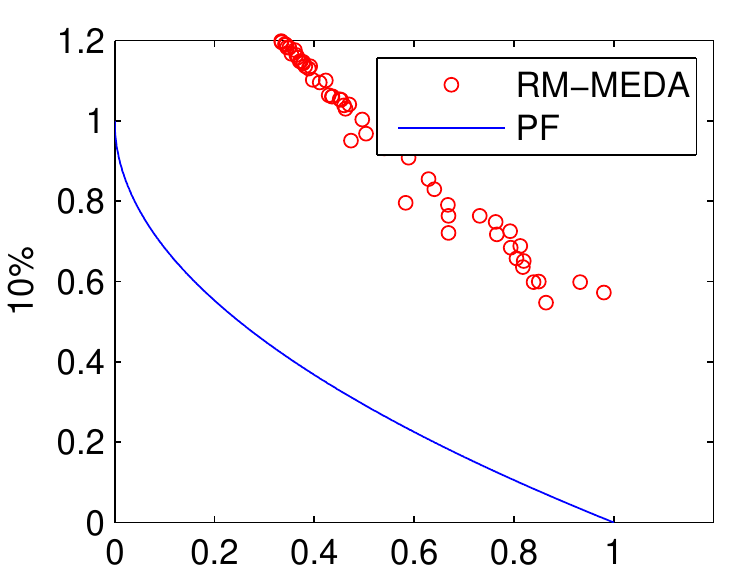}

    \includegraphics[ width=0.50\columnwidth,height=2.3cm]{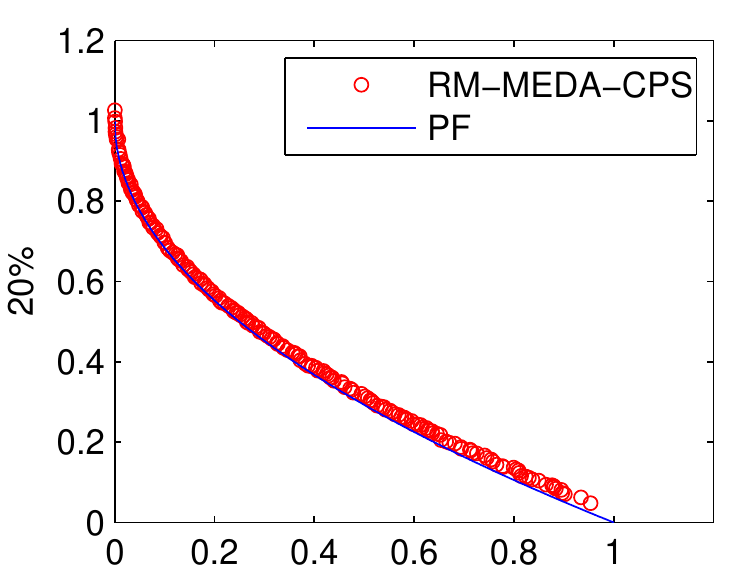}
    \includegraphics[ width=0.46\columnwidth,height=2.3cm]{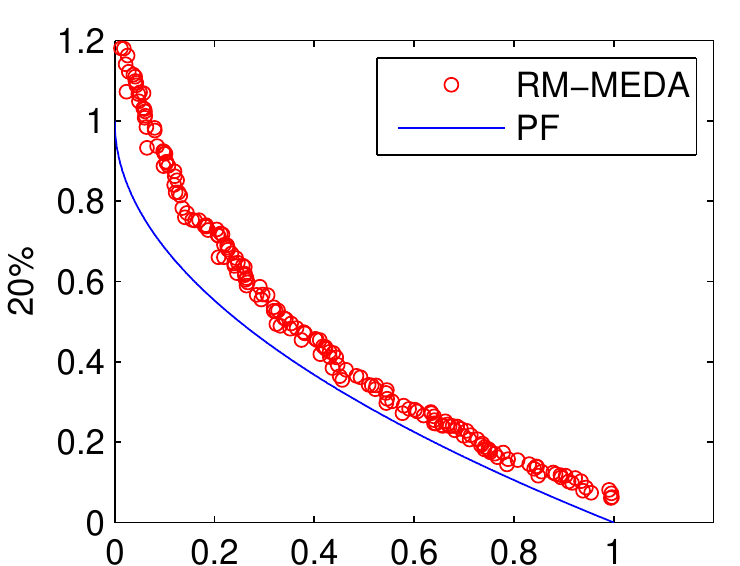}

    \includegraphics[ width=0.48\columnwidth,height=2.3cm]{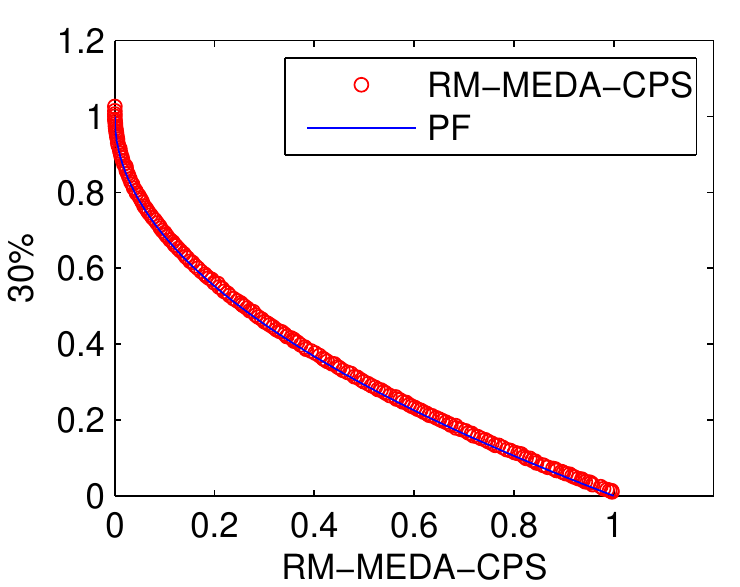}
    \includegraphics[ width=0.48\columnwidth,height=2.3cm]{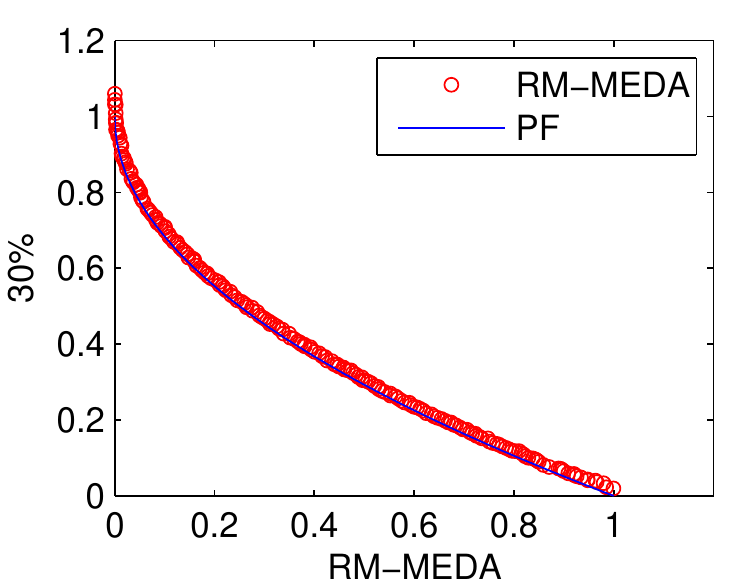}
    \subcaption{ZZJ9}
\end{subfigure}
\caption {The median (according to the IGD metric values) approximations obtained by RM-MEDA-CPS and RM-MEDA after 10\%, 20\%, 30\% of the max FEs on (a) ZZJ1, (b) ZZJ2, and (c) ZZJ9}
\label{fig:rmcompareplot}
\end{figure*}

Table~\ref{tab:rmcompare} shows the statistical results of the IGD and $I^-_H$ metric values obtained by RM-MEDA-CPS and RM-MEDA on ZZJ1-ZZJ10 over 30 runs. The statistical test indicates that according to both the two metrics, RM-MEDA-CPS outperforms RM-MEDA on 7 out of 10 instances, and on the other 3 instances, the two algorithms have similar performances. This results also indicates that given the same FEs, RM-MEDA-CPS works no worse than RM-MEDA. Since the only difference between the two algorithms is on the use of CPS, the experimental results suggest that CPS can successfully improve the performance of RM-MEDA.

Fig.~\ref{fig:rmcompare} plots the run time performance in terms of the IGD values obtained by the two algorithms on the 10 instances. The curves obtained by the two algorithms in Fig.~\ref{fig:rmcompare} show that on all test instances, RM-MEDA-CPS converges faster than RM-MEDA. Fig.~\ref{fig:rmcomparebar} plots the mean FEs required to obtain different levels of the IGD values. It suggests that to obtain the same IGD values, RM-MEDA-CPS uses fewer computational resources than RM-MEDA does. The experimental results in Figs.~\ref{fig:rmcompare} and~\ref{fig:rmcomparebar} suggest that CPS can help to speed up the convergence of RM-MEDA on most of the instances.

In order to visualize the final obtained solutions, we take ZZJ1, ZZJ2 and ZZJ9 as examples and plot the final obtained populations of the median runs according to the IGD values after 10\%, 20\%, and 30\% of the max FEs in Fig.~\ref{fig:rmcompareplot}. It is clear that RM-MEDA-CPS is able to achieve better results than RM-MEDA with the same computational costs.

{\subsubsection{SMS-EMOA-CPS vs. SMS-EMOA}

\begin{table*}[htbp]
\scriptsize
\centering \caption{The statistical results of IGD and $I^-_H$ metric values obtained by SMS-EMOA-CPS and SMS-EMOA on ZZJ1-ZZJ10}\label{tab:sms}
\begin{tabular}{l|c|cccc|cccc}\hline\hline
instance&\multicolumn{1}{c|}{metric}&\multicolumn{4}{c}{SMS-EMOA-CPS}&\multicolumn{4}{|c}{SMS-EMOA}\\
&&mean&std.&min&max&mean&std.&min&max\\\hline
$ZZJ1$	&$IGD$	&3.67e-03($\sim$)	&2.29e-05	&3.62e-03	&3.71e-03	&3.67e-03	&1.85e-05	&3.64e-03	&3.69e-03	\\
&$I^-_H$&4.34e-03(+)	&6.12e-05	&4.26e-03	&4.48e-03	&4.37e-03	&5.70e-05	&4.28e-03	&4.55e-03	\\
$ZZJ2$	&$IGD$	&4.53e-03($\sim$)	&1.85e-04	&4.29e-03	&5.22e-03	&4.53e-03	&1.26e-04	&4.32e-03	&4.76e-03	\\
&$I^-_H$&4.40e-03(+)	&6.38e-05	&4.29e-03	&4.57e-03	&4.46e-03	&7.73e-05	&4.29e-03	&4.62e-03	\\
$ZZJ3$	&$IGD$	&1.55e-01(+)	&1.62e-02	&1.28e-01	&1.86e-01	&2.06e-01	&2.60e-02	&1.43e-01	&2.53e-01	\\
&$I^-_H$&2.00e-01(+)	&2.22e-02	&1.63e-01	&2.45e-01	&2.69e-01	&3.44e-02	&1.86e-01	&3.26e-01	\\
$ZZJ4$	&$IGD$	&5.32e-02($\sim$)	&4.63e-04	&5.22e-02	&5.41e-02	&5.32e-02	&8.29e-04	&5.08e-02	&5.42e-02	\\
&$I^-_H$&2.40e-02($\sim$)	&1.02e-04	&2.38e-02	&2.42e-02	&2.40e-02	&9.36e-05	&2.38e-02	&2.42e-02	\\
$ZZJ5$	&$IGD$	&1.10e-02($\sim$)	&2.19e-02	&5.05e-03	&1.27e-01	&7.15e-03	&1.14e-03	&5.42e-03	&9.94e-03	\\
&$I^-_H$&1.55e-02($\sim$)	&2.57e-02	&7.78e-03	&1.51e-01	&1.11e-02	&1.67e-03	&8.45e-03	&1.52e-02	\\
$ZZJ6$	&$IGD$	&2.14e-01($\sim$)	&2.87e-01	&5.50e-03	&6.10e-01	&2.29e-01	&2.95e-01	&5.86e-03	&6.10e-01	\\
&$I^-_H$&1.95e-01($\sim$)	&2.50e-01	&7.13e-03	&5.33e-01	&2.03e-01	&2.56e-01	&7.80e-03	&5.33e-01	\\
$ZZJ7$	&$IGD$	&3.40e-01(+)	&7.96e-03	&3.28e-01	&3.56e-01	&3.52e-01	&7.34e-03	&3.36e-01	&3.64e-01	\\
&$I^-_H$&4.47e-01(+)	&6.00e-03	&4.33e-01	&4.60e-01	&4.57e-01	&4.62e-03	&4.47e-01	&4.65e-01	\\
$ZZJ8$	&$IGD$	&4.29e-01(-)	&6.93e-02	&3.82e-01	&6.22e-01	&3.44e-01	&1.13e-01	&5.78e-02	&3.88e-01	\\
&$I^-_H$&3.13e-01(-)	&7.06e-02	&2.52e-01	&4.10e-01	&2.25e-01	&7.20e-02	&4.04e-02	&2.55e-01	\\
$ZZJ9$	&$IGD$	&1.27e-02(+)	&7.08e-03	&4.39e-03	&3.34e-02	&1.61e-02	&6.66e-03	&8.38e-03	&3.17e-02	\\
&$I^-_H$&2.23e-02(+)	&1.16e-02	&8.30e-03	&5.55e-02	&2.89e-02	&1.11e-02	&1.55e-02	&5.59e-02	\\
$ZZJ10$	&$IGD$	&1.83e+01($\sim$)	&4.15e+00	&7.16e+00	&2.92e+01	&1.88e+01	&4.38e+00	&2.29e+00	&2.53e+01	\\
&$I^-_H$&1.11e+00($\sim$)	&2.26e-16	&1.11e+00	&1.11e+00	&1.11e+00	&2.26e-16	&1.11e+00	&1.11e+00	\\
\hline
$+/-/\sim$ &$IGD$	&3/1/6	&&&&&&\\
$+/-/\sim$ &$I^-_H$ &5/1/4		&&&&&&\\
\hline\hline
\end{tabular}
\end{table*}

\begin{figure*}[htbp]
\centering
\begin{subfigure}[t]{0.38\columnwidth}
    \includegraphics[ width=\columnwidth]{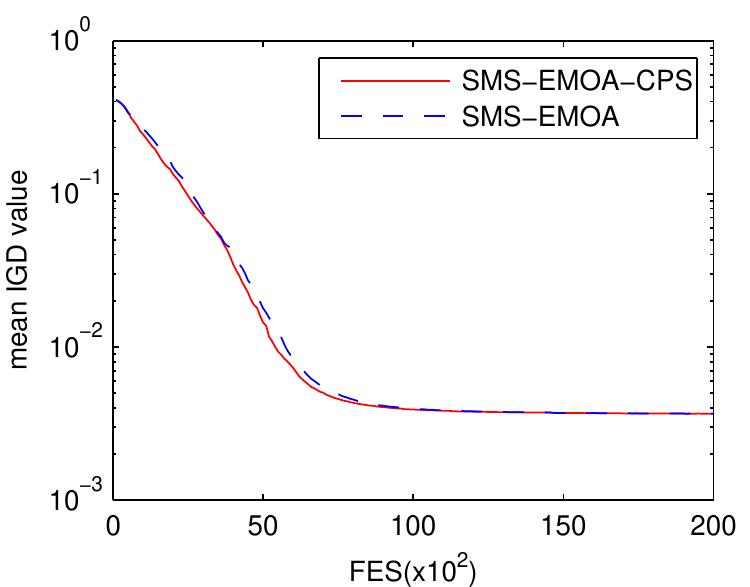}
    \subcaption{ZZJ1}
\end{subfigure}
\begin{subfigure}[t]{0.38\columnwidth}
    \includegraphics[ width=\columnwidth]{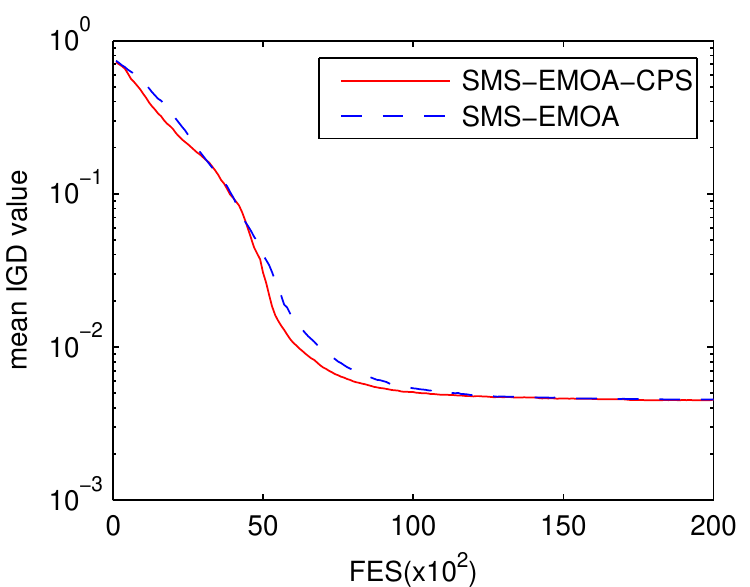}
    \subcaption{ZZJ2}
\end{subfigure}
\begin{subfigure}[t]{0.38\columnwidth}
    \includegraphics[ width=\columnwidth]{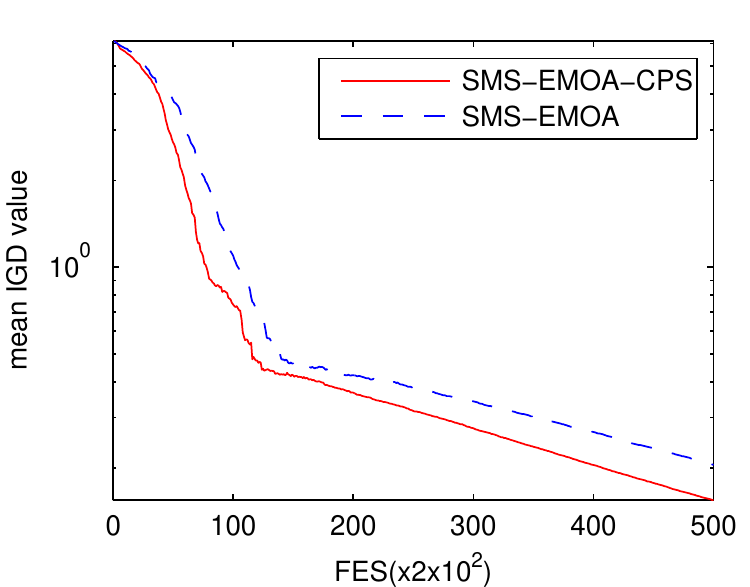}
    \subcaption{ZZJ3}
\end{subfigure}
\begin{subfigure}[t]{0.38\columnwidth}
    \includegraphics[ width=\columnwidth]{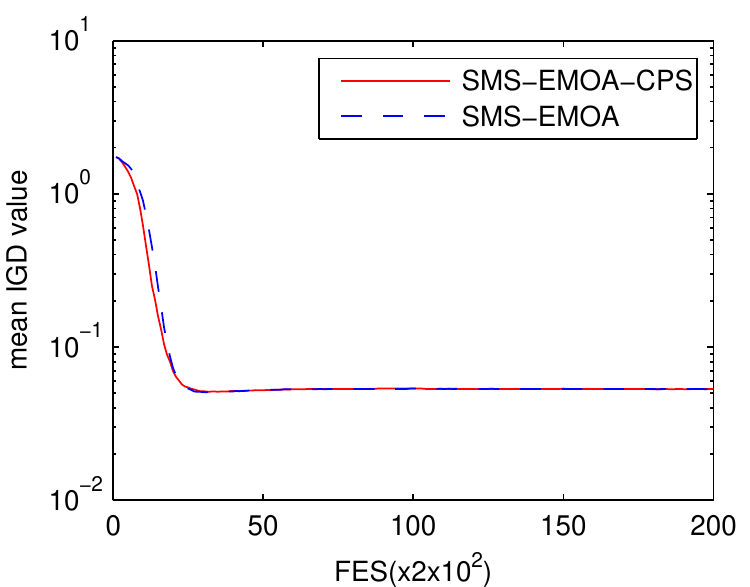}
    \subcaption{ZZJ4}
\end{subfigure}
\begin{subfigure}[t]{0.38\columnwidth}
    \includegraphics[ width=\columnwidth]{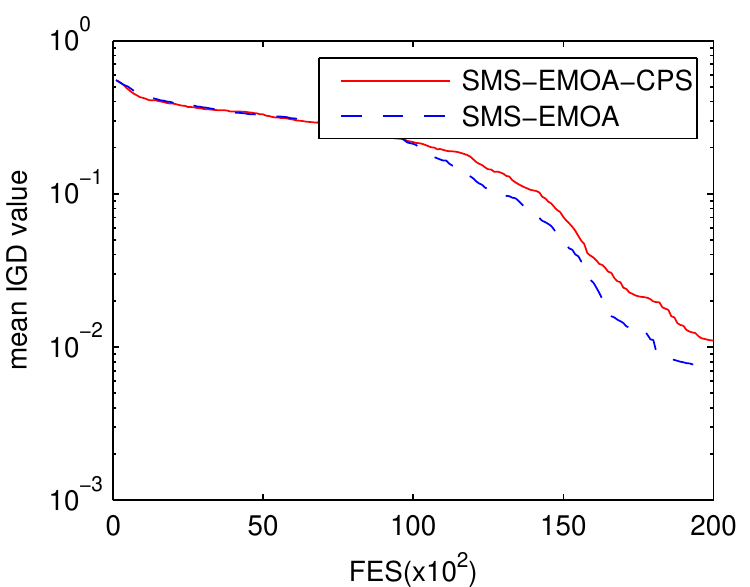}
    \subcaption{ZZJ5}
\end{subfigure}
\begin{subfigure}[t]{0.38\columnwidth}
    \includegraphics[ width=\columnwidth]{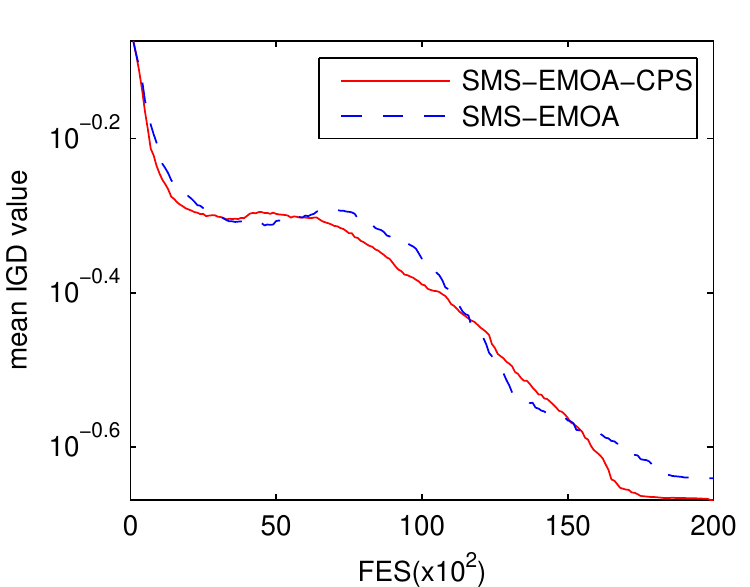}
    \subcaption{ZZJ6}
\end{subfigure}
\begin{subfigure}[t]{0.38\columnwidth}
    \includegraphics[ width=\columnwidth]{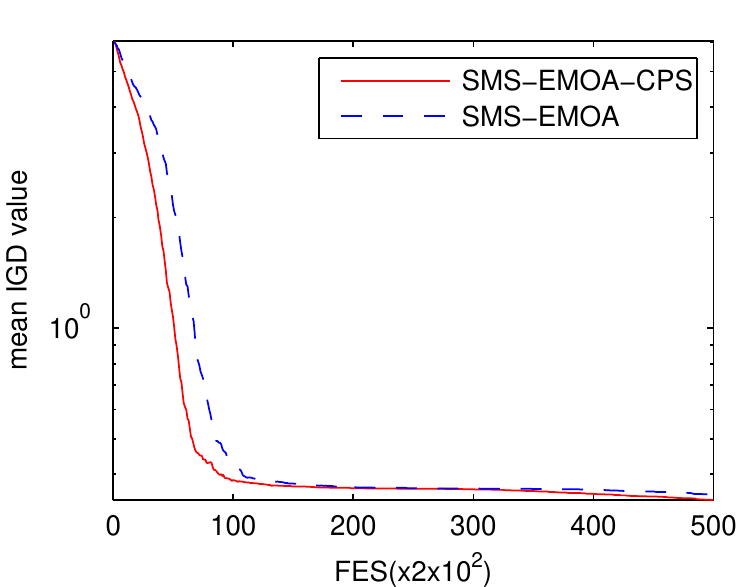}
    \subcaption{ZZJ7}
\end{subfigure}
\begin{subfigure}[t]{0.38\columnwidth}
    \includegraphics[ width=\columnwidth]{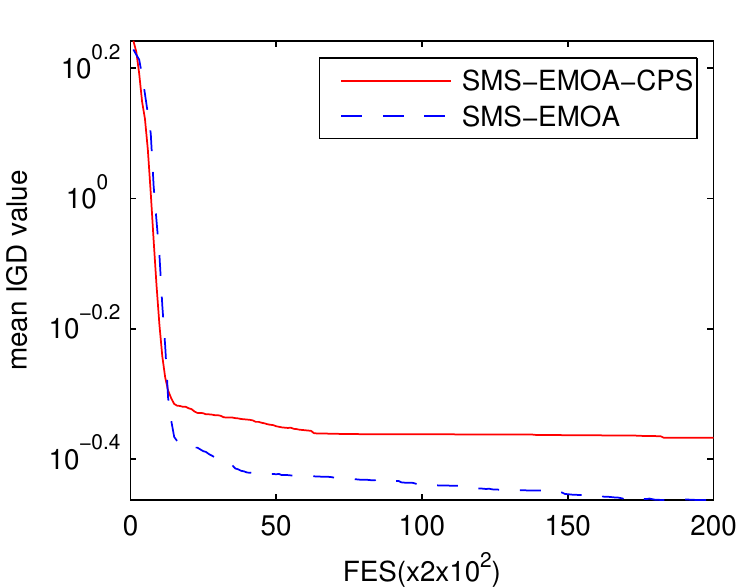}
    \subcaption{ZZJ8}
\end{subfigure}
\begin{subfigure}[t]{0.38\columnwidth}
    \includegraphics[ width=\columnwidth]{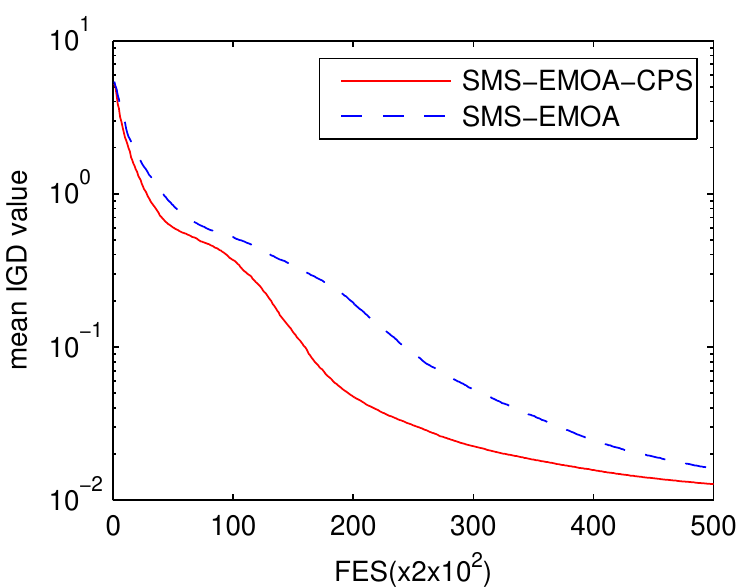}
    \subcaption{ZZJ9}
\end{subfigure}
\begin{subfigure}[t]{0.38\columnwidth}
    \includegraphics[ width=\columnwidth]{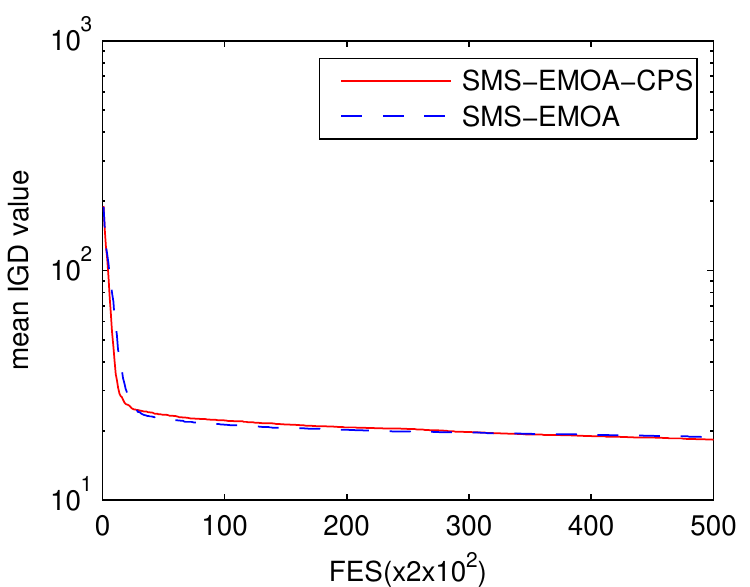}
    \subcaption{ZZJ10}
\end{subfigure}
\caption {The mean IGD values versus FEs obtained by SMS-EMOA-CPS and SMS-EMOA over 30 runs}
\label{fig:sms}
\end{figure*}

\begin{figure*}[htbp]
\centering
\begin{subfigure}[t]{0.38\columnwidth}
    \includegraphics[ width=\columnwidth]{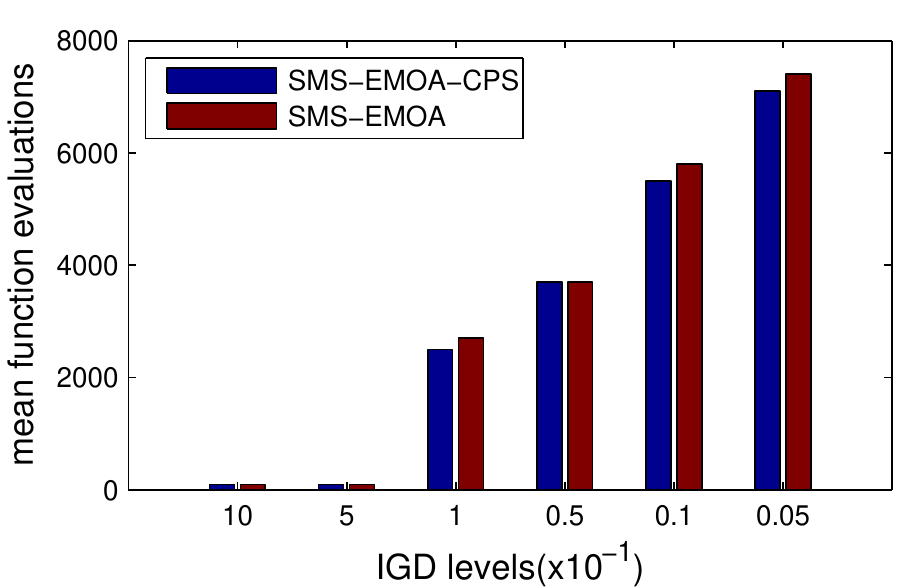}
    \subcaption{ZZJ1}
\end{subfigure}
\begin{subfigure}[t]{0.38\columnwidth}
    \includegraphics[ width=\columnwidth]{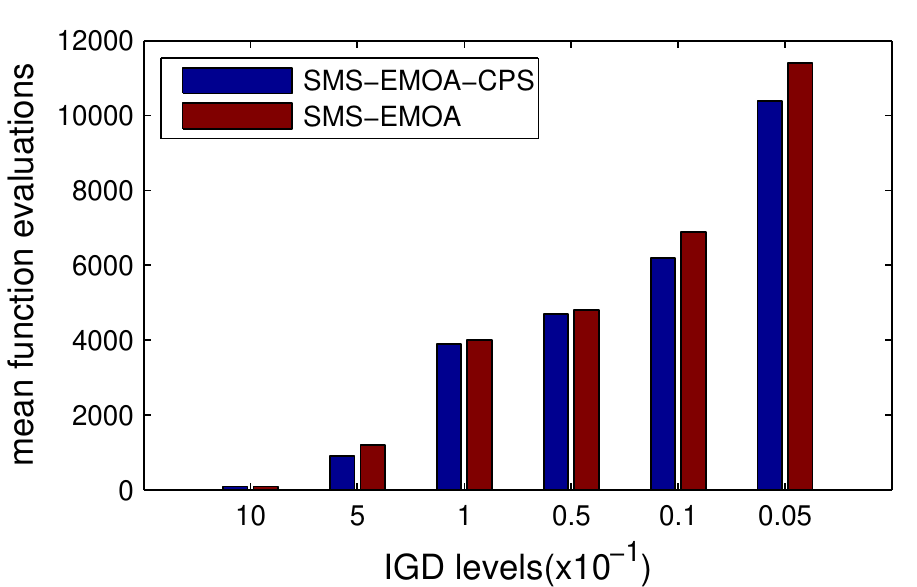}
    \subcaption{ZZJ2}
\end{subfigure}
\begin{subfigure}[t]{0.38\columnwidth}
    \includegraphics[ width=\columnwidth]{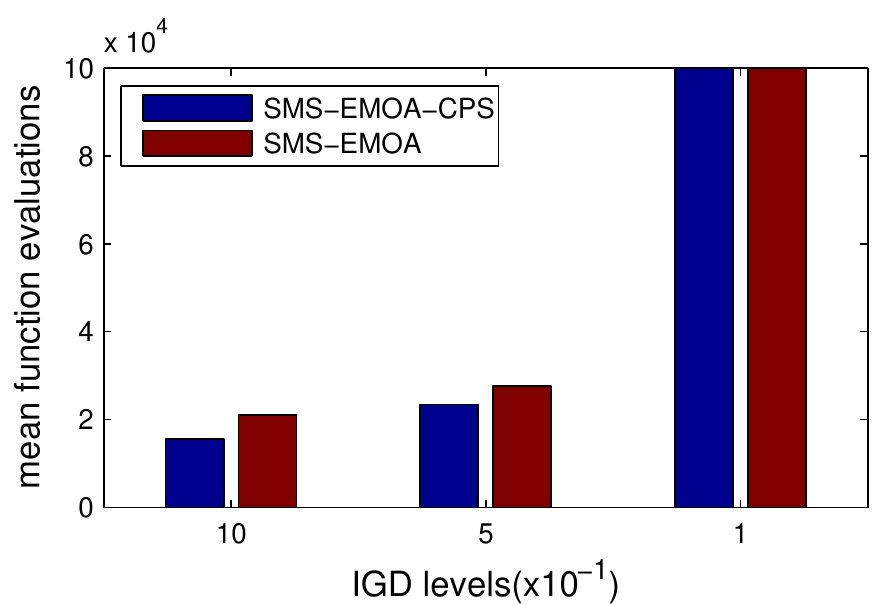}
    \subcaption{ZZJ3}
\end{subfigure}
\begin{subfigure}[t]{0.38\columnwidth}
    \includegraphics[ width=\columnwidth]{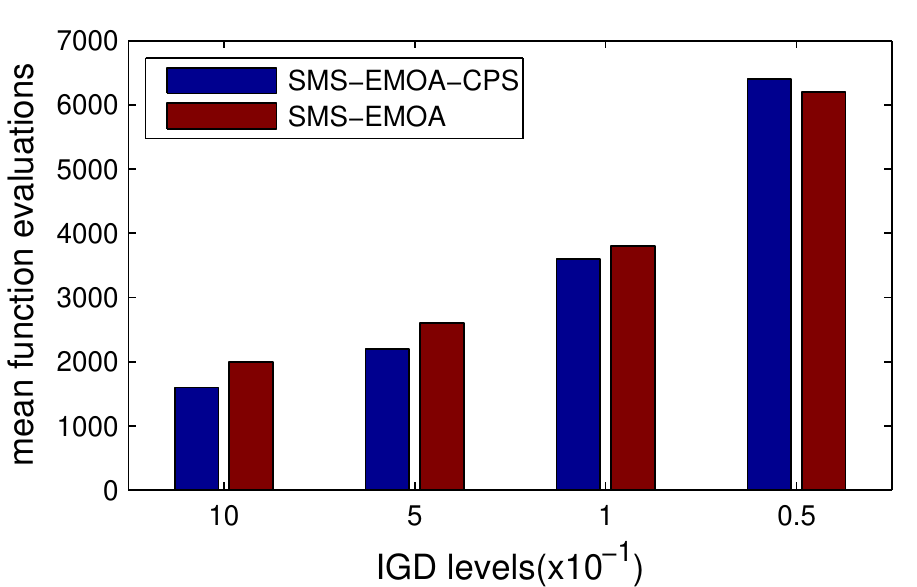}
    \subcaption{ZZJ4}
\end{subfigure}
\begin{subfigure}[t]{0.38\columnwidth}
    \includegraphics[ width=\columnwidth]{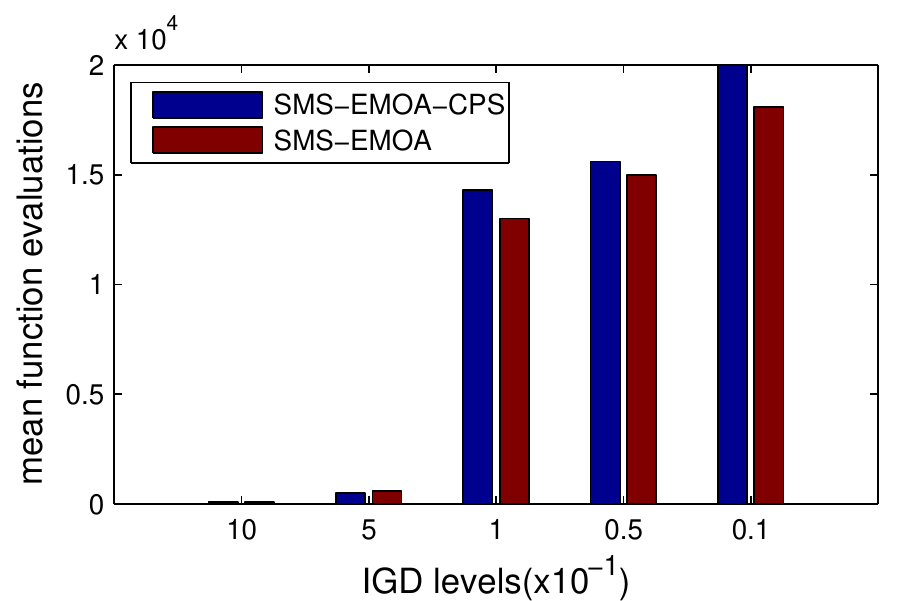}
    \subcaption{ZZJ5}
\end{subfigure}
\begin{subfigure}[t]{0.38\columnwidth}
    \includegraphics[ width=\columnwidth]{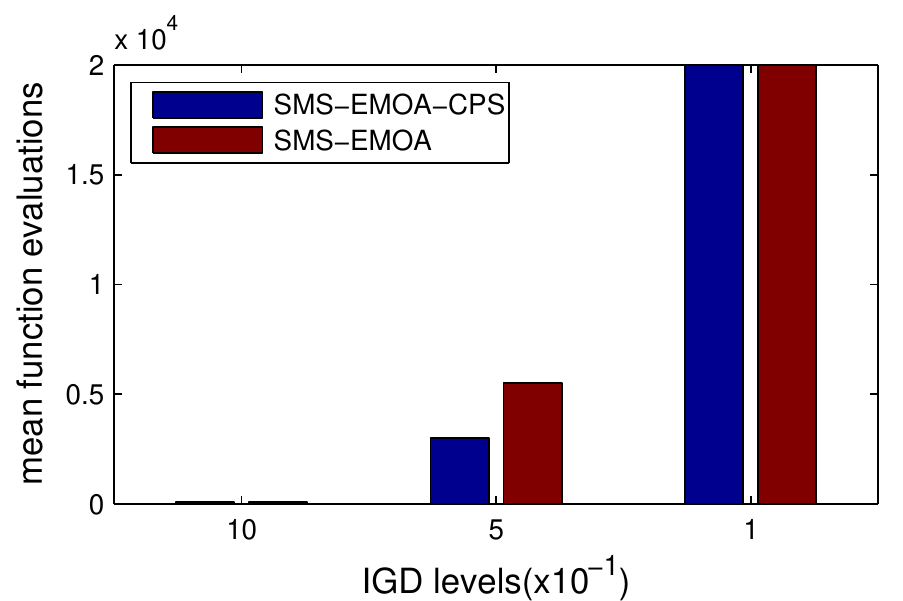}
    \subcaption{ZZJ6}
\end{subfigure}
\begin{subfigure}[t]{0.38\columnwidth}
    \includegraphics[ width=\columnwidth]{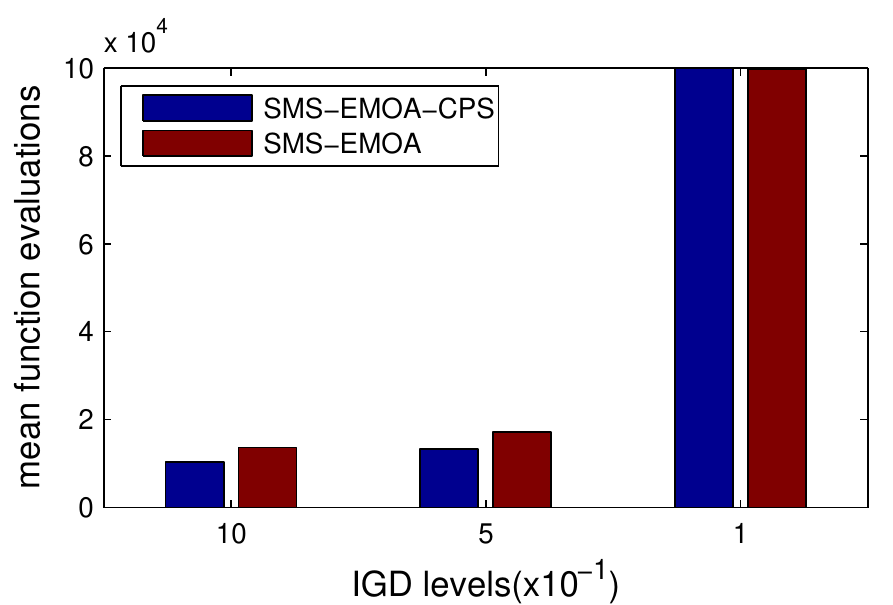}
    \subcaption{ZZJ7}
\end{subfigure}
\begin{subfigure}[t]{0.38\columnwidth}
    \includegraphics[ width=\columnwidth]{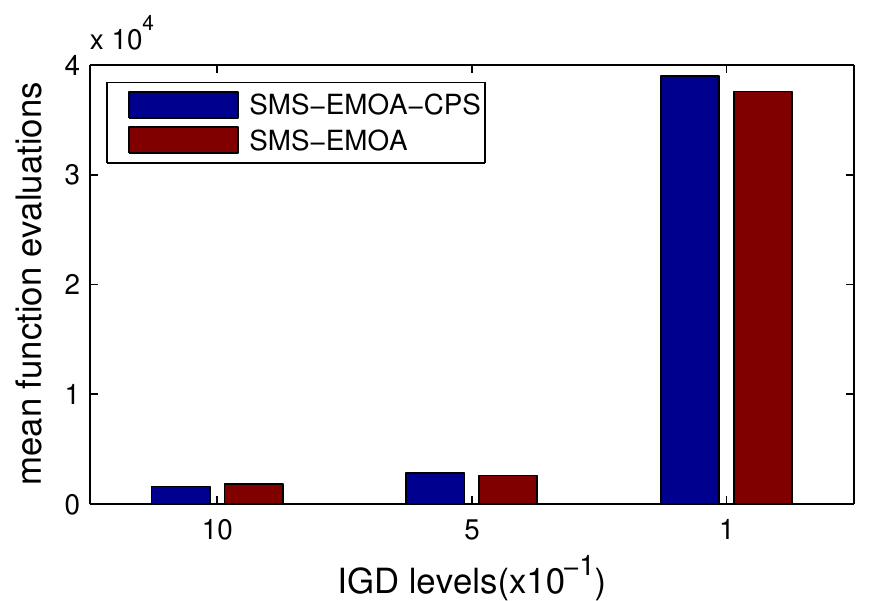}
    \subcaption{ZZJ8}
\end{subfigure}
\begin{subfigure}[t]{0.38\columnwidth}
    \includegraphics[ width=\columnwidth]{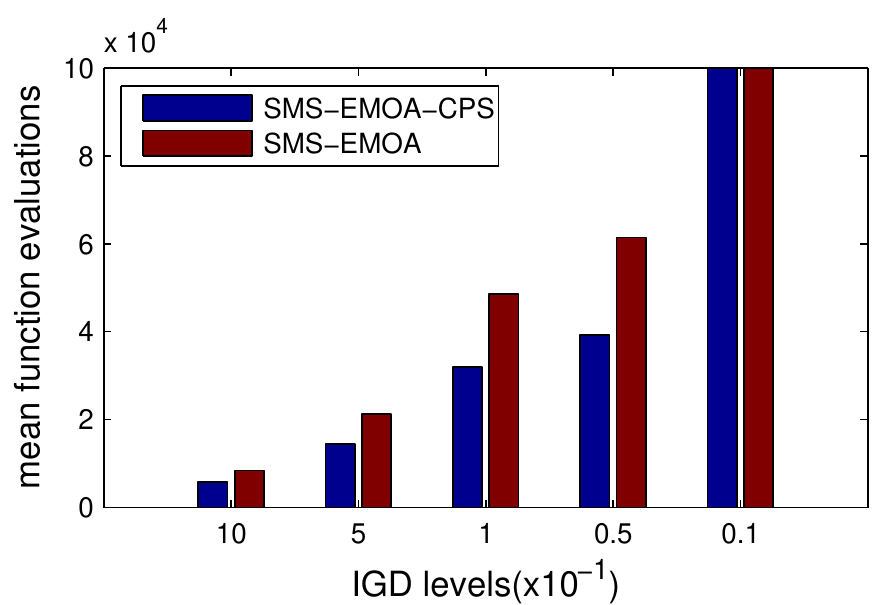}
    \subcaption{ZZJ9}
\end{subfigure}
\begin{subfigure}[t]{0.38\columnwidth}
    \includegraphics[ width=\columnwidth]{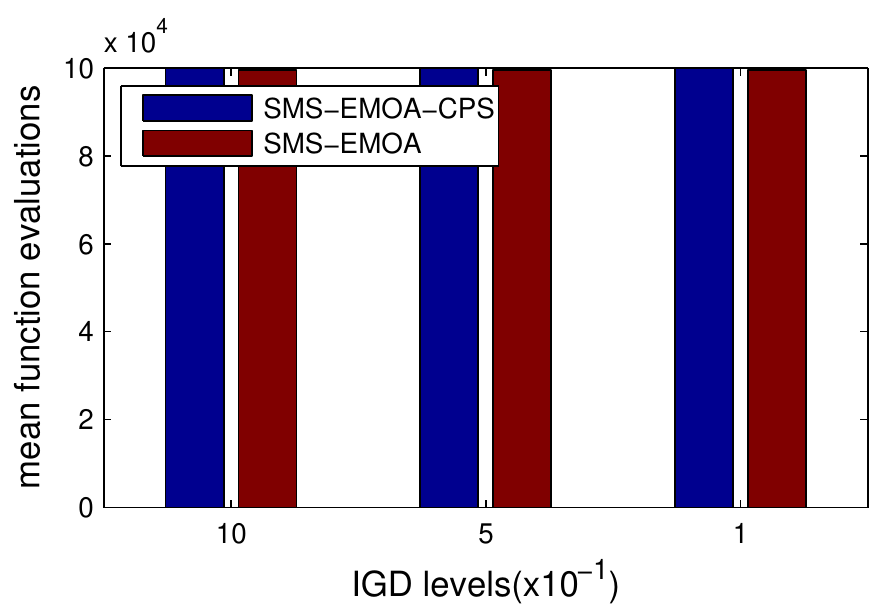}
    \subcaption{ZZJ10}
\end{subfigure}
\caption {The mean FEs required by SMS-EMOA-CPS and SMS-EMOA to obtain different IGD values over 30 runs}
\label{fig:smsbar}
\end{figure*}

\begin{figure*}[htbp]
\centering
\begin{subfigure}[t]{0.64\columnwidth}
    \includegraphics[ width=0.50\columnwidth,height=2.3cm]{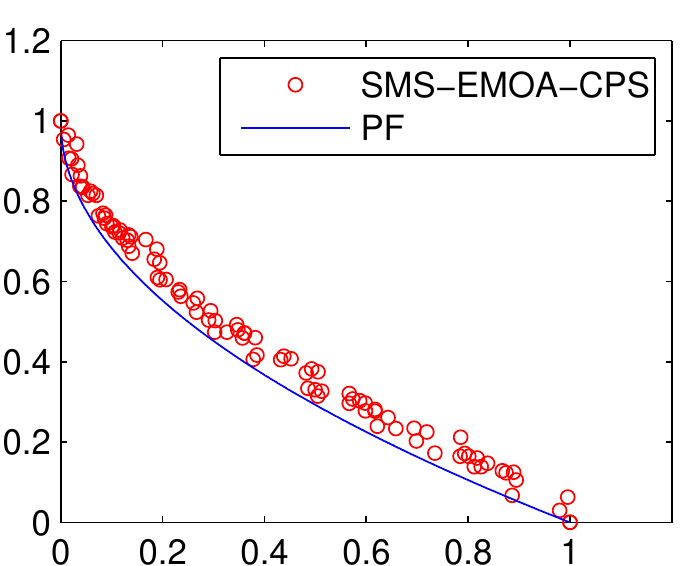}
    \includegraphics[ width=0.46\columnwidth,height=2.3cm]{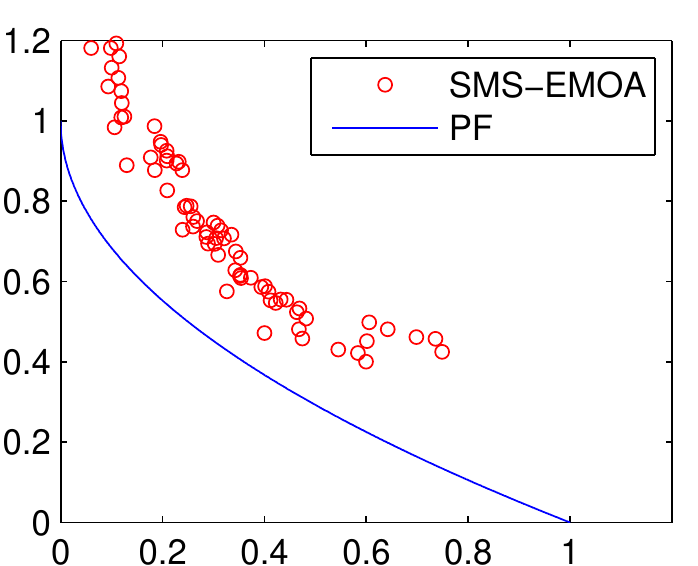}

    \includegraphics[ width=0.50\columnwidth,height=2.3cm]{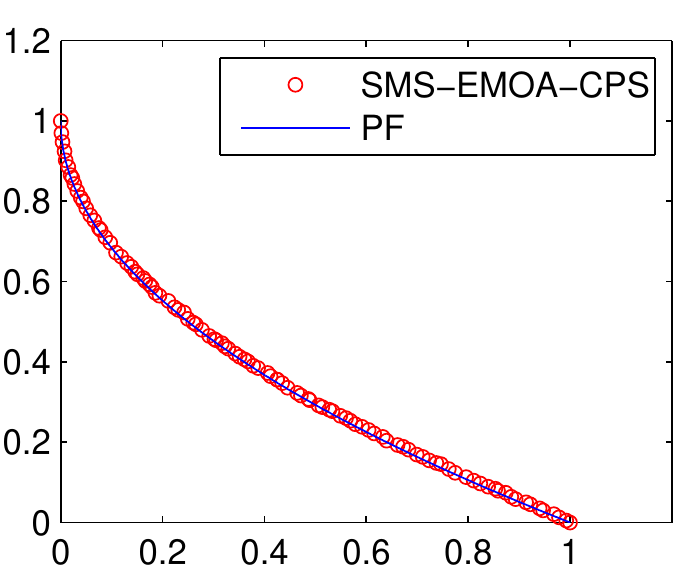}
    \includegraphics[ width=0.46\columnwidth,height=2.3cm]{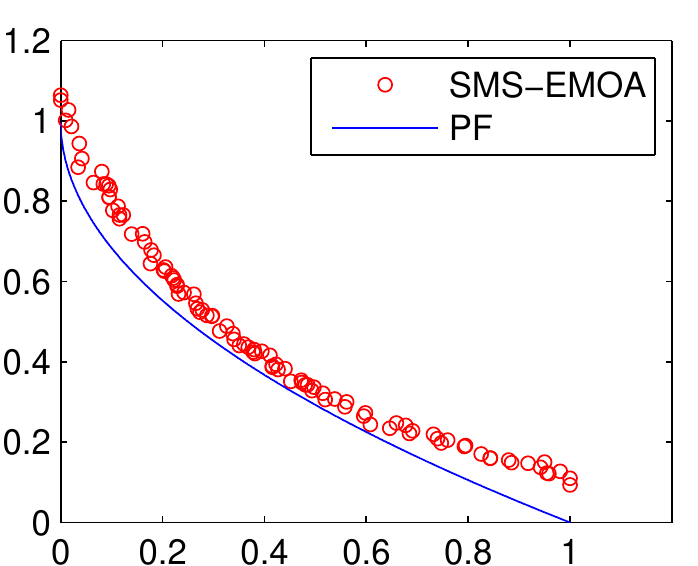}

    \includegraphics[ width=0.50\columnwidth,height=2.3cm]{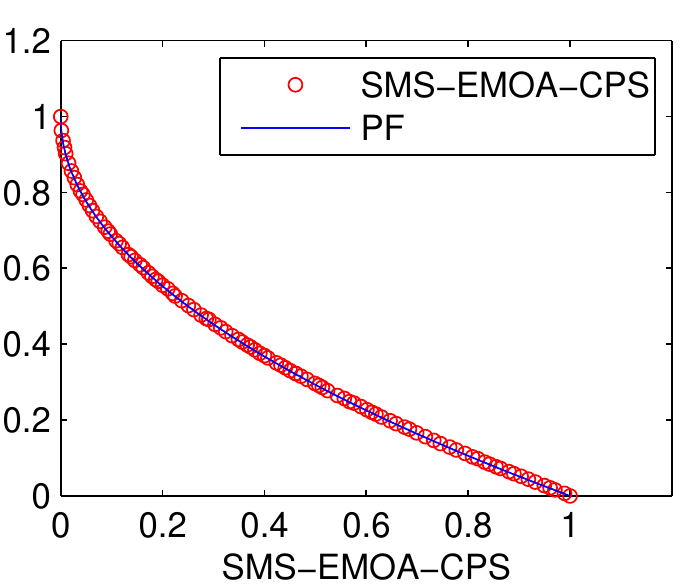}
    \includegraphics[ width=0.46\columnwidth,height=2.3cm]{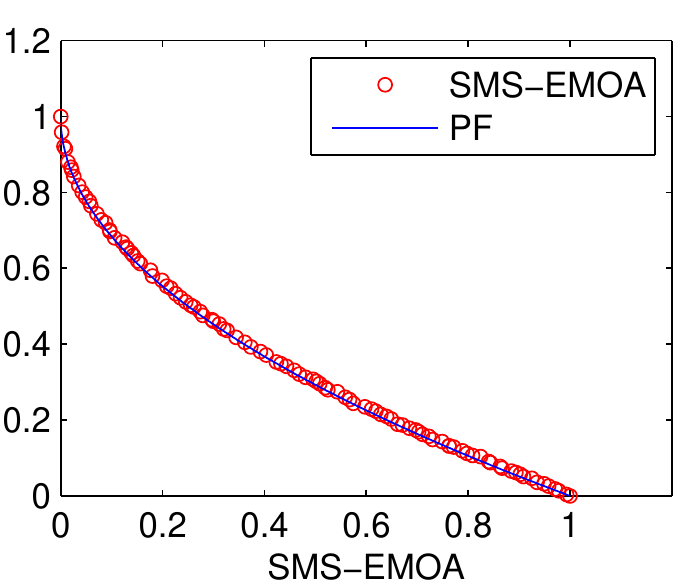}
    \subcaption{ZZJ1}
\end{subfigure}
\begin{subfigure}[t]{0.64\columnwidth}
    \includegraphics[ width=0.50\columnwidth,height=2.3cm]{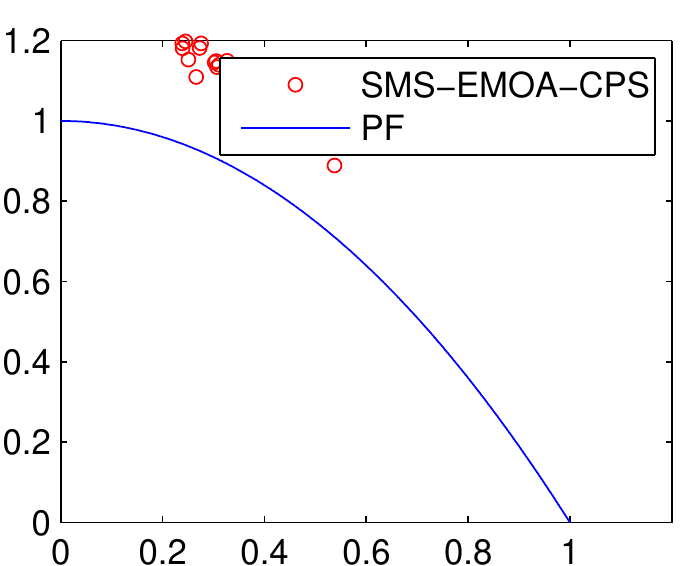}
    \includegraphics[ width=0.46\columnwidth,height=2.3cm]{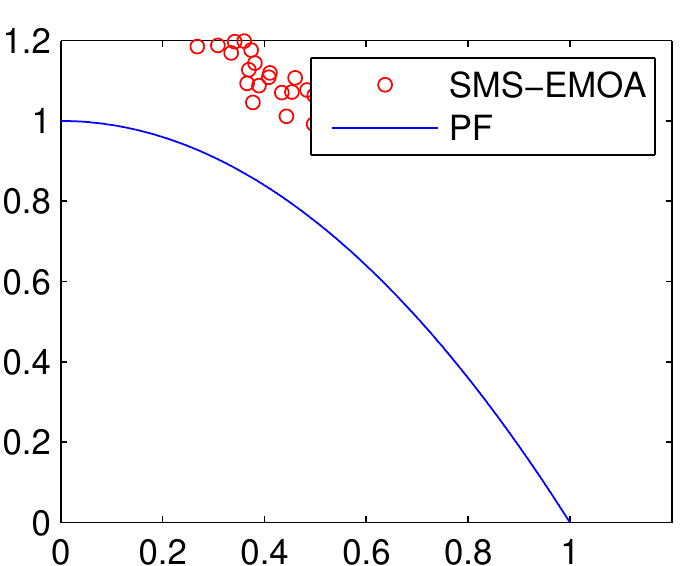}

    \includegraphics[ width=0.50\columnwidth,height=2.3cm]{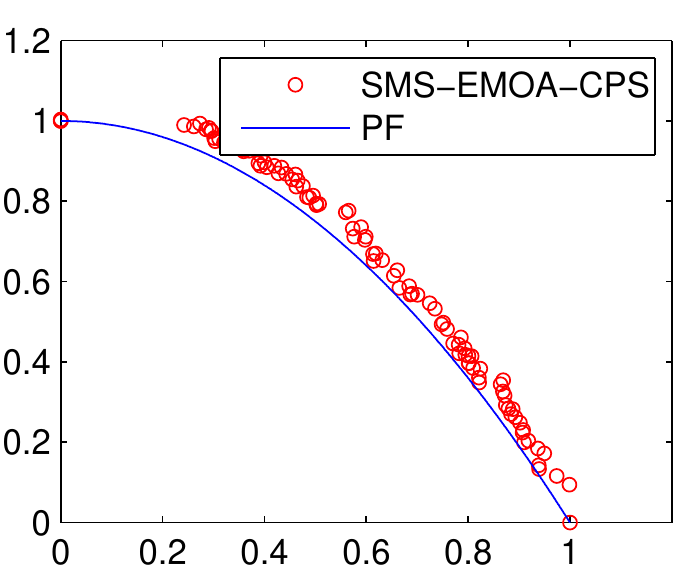}
    \includegraphics[ width=0.46\columnwidth,height=2.3cm]{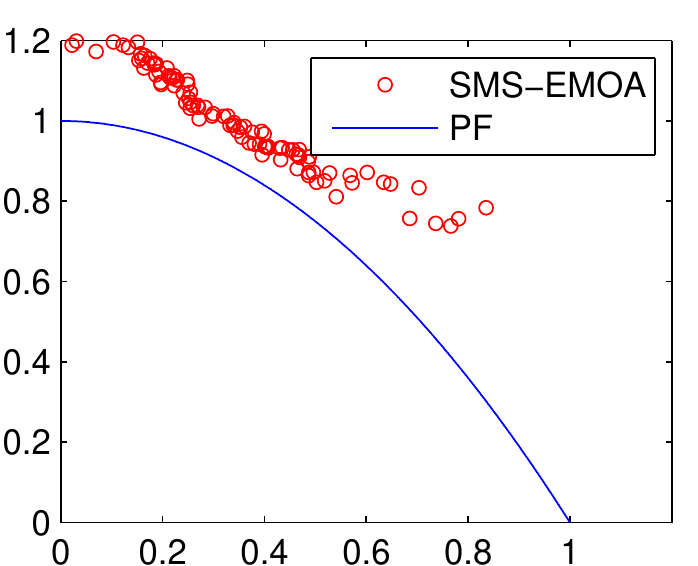}

    \includegraphics[ width=0.50\columnwidth,height=2.3cm]{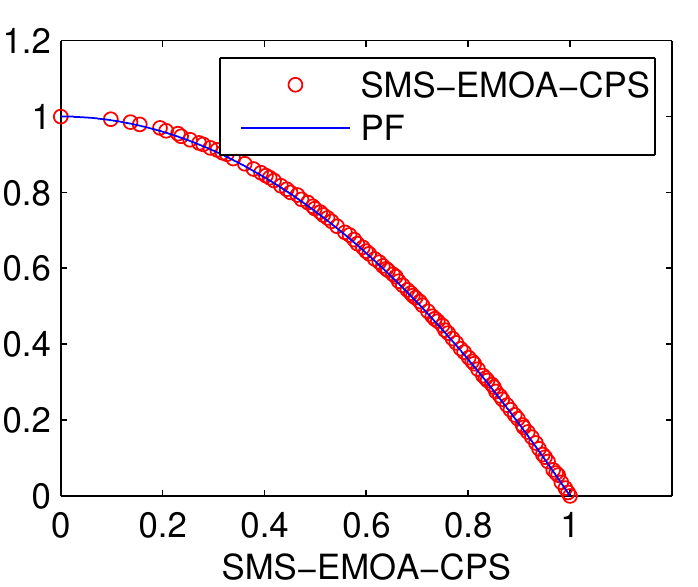}
    \includegraphics[ width=0.46\columnwidth,height=2.3cm]{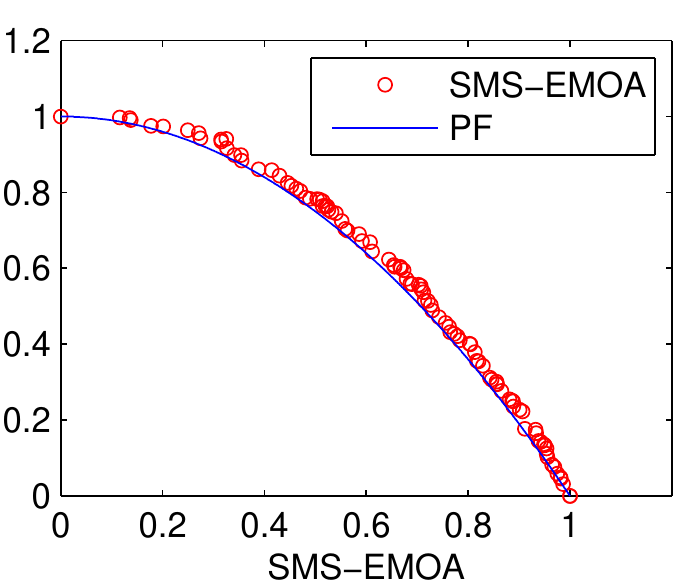}
    \subcaption{ZZJ2}
\end{subfigure}
\begin{subfigure}[t]{0.64\columnwidth}
    \includegraphics[ width=0.50\columnwidth,height=2.3cm]{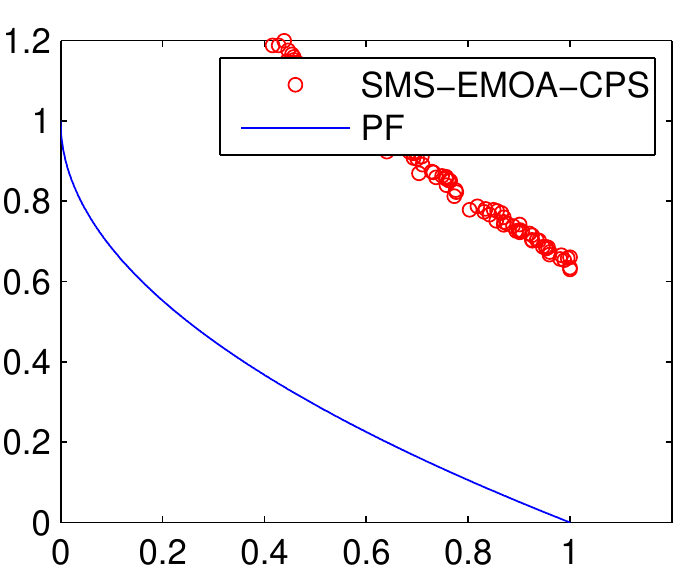}
    \includegraphics[ width=0.46\columnwidth,height=2.3cm]{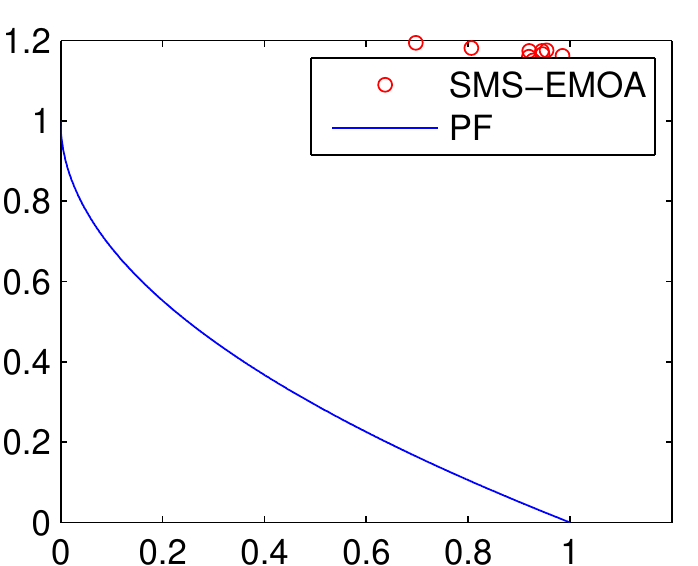}

    \includegraphics[ width=0.50\columnwidth,height=2.3cm]{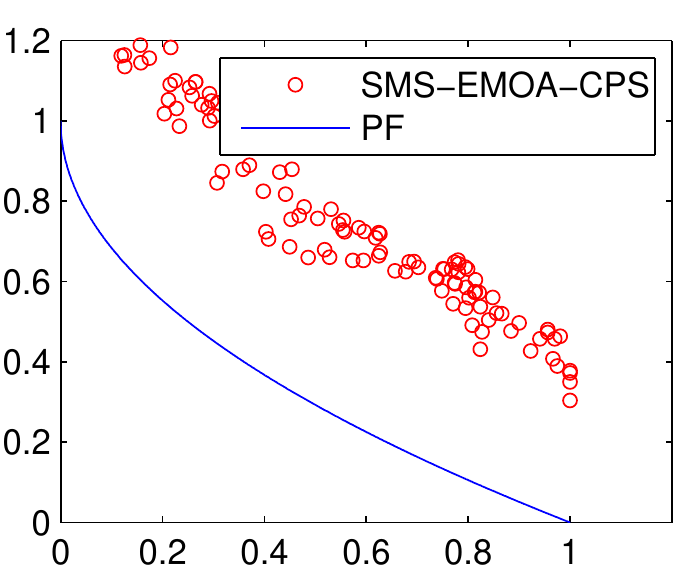}
    \includegraphics[ width=0.46\columnwidth,height=2.3cm]{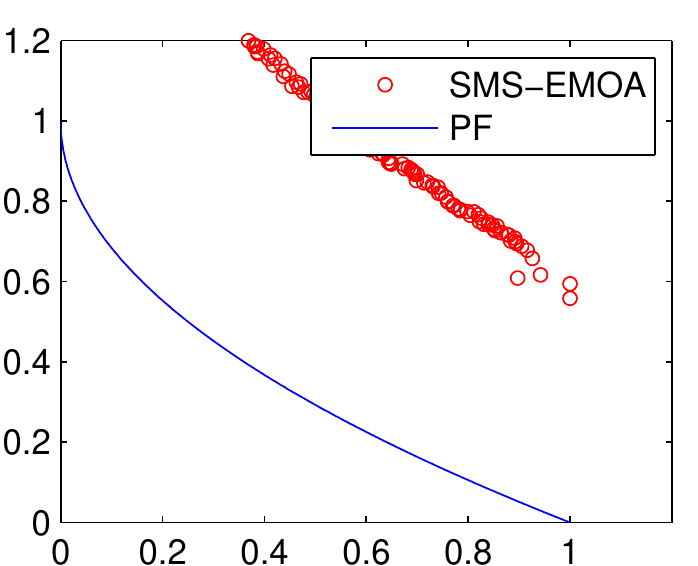}

    \includegraphics[ width=0.48\columnwidth,height=2.3cm]{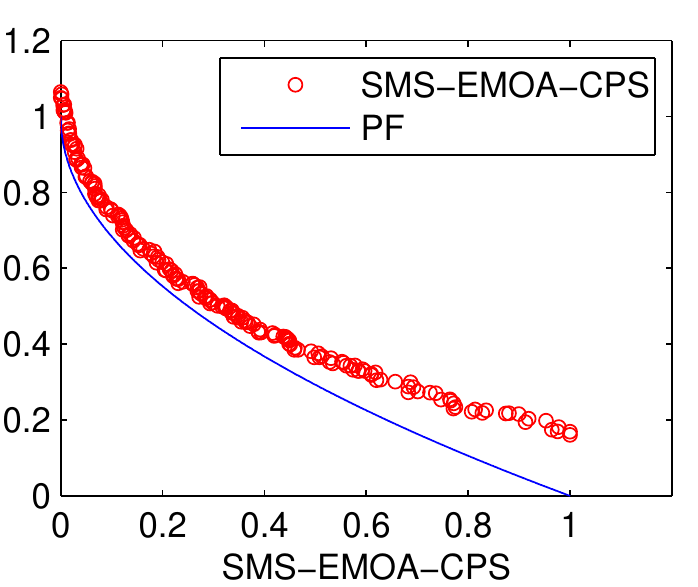}
    \includegraphics[ width=0.48\columnwidth,height=2.3cm]{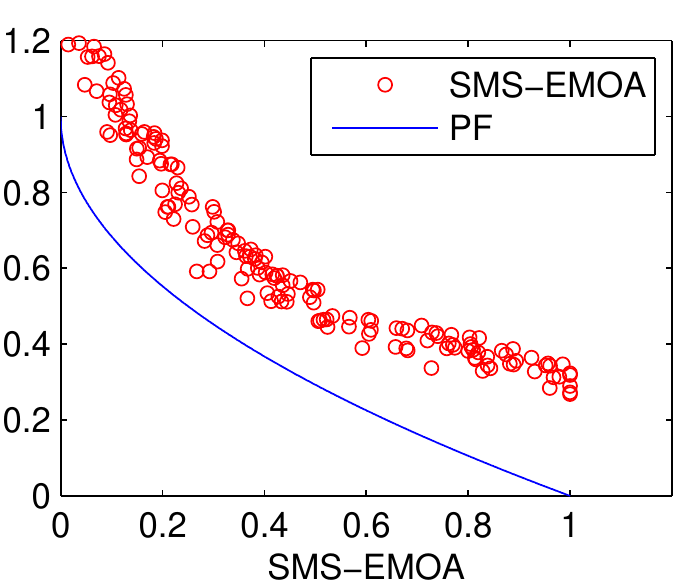}
    \subcaption{ZZJ9}
\end{subfigure}
\caption {The median (according to the IGD metric values) approximations obtained by SMS-EMOA-CPS and SMS-EMOA after 10\%, 20\%, 30\% of the max FEs on (a) ZZJ1, (b) ZZJ2, and (c) ZZJ9}
\label{fig:smsplot}
\end{figure*}

Table~\ref{tab:sms} presents the statistical results of IGD and $I^-_H$ metric values obtained by SMS-EMOA-CPS and SMS-EMOA on ZZJ1-ZZJ10 over 30 runs. The statistical test shows different results according to different performance metrics. According to the IGD metric, SMS-EMOA-CPS works better than SMS-EMOA on 3 instances, works worse than SMS-EMOA on 1 instance, and works similar to SMS-EMOA on the other 6 instances. According to the $I^-_H$ metric, SMS-EMOA-CPS shows better performance on 5 instances, worse performance on 1 instance, and similar performance on the other 4 instances than SMS-EMOA. Even so, we can still say that SMS-EMOA-CPS is not worse than SMS-EMOA on most of the given instances according to the final obtained results.

Fig.~\ref{fig:sms} presents the run time performance in terms of the IGD values obtained by the two algorithms on the 10 instances. It shows that on ZZJ5 and ZZJ8, SMS-EMOA-CPS converges slower than SMS-EMOA, and on all the other instances, SMS-EMOA-CPS converges faster than or similar to SMS-EMOA. Fig.~\ref{fig:smsbar} plots the FEs required by the two algorithms to obtain some levels of IGD values. This figure suggests that to obtain the same IGD values, SMS-EMOA-CPS takes fewer computational resources than SMS-EMOA does blue, especially in the early stages on most of the instances. Figs.~\ref{fig:sms} and~\ref{fig:smsbar} suggest that CPS can improve the convergence speed of SMS-EMOA on most of the instances.

In order to visualize the final obtained solutions, we draw the final obtained populations of the median runs according to the IGD values after 10\%, 20\%, and 30\% of the max FEs in Fig.~\ref{fig:smsplot} for ZZJ1, ZZJ2 and ZZJ9. The figure shows that the SMS-EMOA-CPS can achieve better results than SMS-EMOA with the same computational costs on most of the instances.

\subsubsection{MOEA/D-MO-CPS vs. MOEA/D-MO}

\begin{table*}[htbp]
\scriptsize
\centering \caption{The statistical results of IGD and $I^-_H$ metric values obtained by MOEA/D-MO-CPS and MOEA/D-MO on ZZJ1-ZZJ10}\label{tab:moead}
\begin{tabular}{l|c|cccc|cccc}\hline\hline
instance&\multicolumn{1}{c|}{metric}&\multicolumn{4}{c}{MOEA/D-MO-CPS}&\multicolumn{4}{|c}{MOEA/D-MO}\\
&&mean&std.&min&max&mean&std.&min&max\\\hline
$ZZJ1$	&$IGD$	&4.31e-03(+)	&9.29e-05	&4.17e-03	&4.50e-03	&4.44e-03	&1.18e-04	&4.26e-03	&4.81e-03	\\
&$I^-_H$&6.44e-03(+)	&2.34e-04	&6.09e-03	&6.99e-03	&6.77e-03	&2.76e-04	&6.33e-03	&7.55e-03	\\
$ZZJ2$	&$IGD$	&4.19e-03(+)	&7.37e-05	&4.06e-03	&4.35e-03	&4.32e-03	&1.19e-04	&4.18e-03	&4.72e-03	\\
&$I^-_H$&6.24e-03(+)	&2.36e-04	&5.78e-03	&6.77e-03	&6.67e-03	&3.45e-04	&6.18e-03	&7.74e-03	\\
$ZZJ3$	&$IGD$	&1.47e-01(+)	&2.63e-02	&9.65e-02	&1.91e-01	&1.94e-01	&2.88e-02	&1.08e-01	&2.46e-01	\\
&$I^-_H$&2.13e-01(+)	&4.72e-02	&1.21e-01	&2.82e-01	&2.82e-01	&4.58e-02	&1.38e-01	&3.53e-01	\\
$ZZJ4$	&$IGD$	&4.15e-02(+)	&7.28e-04	&4.02e-02	&4.31e-02	&4.25e-02	&6.35e-04	&4.13e-02	&4.39e-02	\\
&$I^-_H$&4.03e-02(+)	&7.39e-04	&3.89e-02	&4.19e-02	&4.09e-02	&9.75e-04	&3.90e-02	&4.28e-02	\\
$ZZJ5$	&$IGD$	&5.61e-03(+)	&4.23e-04	&4.86e-03	&6.62e-03	&6.01e-03	&5.07e-04	&5.34e-03	&7.78e-03	\\
&$I^-_H$&8.99e-03(+)	&7.14e-04	&7.67e-03	&1.06e-02	&9.66e-03	&8.09e-04	&8.56e-03	&1.24e-02	\\
$ZZJ6$	&$IGD$	&1.38e-01($\sim$)	&2.45e-01	&5.88e-03	&6.10e-01	&1.00e-01	&1.97e-01	&5.88e-03	&6.10e-01	\\
&$I^-_H$&1.31e-01($\sim$)	&2.16e-01	&1.02e-02	&5.33e-01	&1.10e-01	&1.92e-01	&1.01e-02	&5.33e-01	\\
$ZZJ7$	&$IGD$	&1.87e-01($\sim$)	&8.20e-02	&1.35e-01	&5.80e-01	&1.77e-01	&1.82e-02	&1.51e-01	&2.29e-01	\\
&$I^-_H$&2.49e-01($\sim$)	&7.09e-02	&1.85e-01	&5.27e-01	&2.47e-01	&2.75e-02	&2.06e-01	&3.22e-01	\\
$ZZJ8$	&$IGD$	&1.19e-01($\sim$)	&1.23e-01	&5.19e-02	&3.88e-01	&1.32e-01	&1.30e-01	&5.28e-02	&3.87e-01	\\
&$I^-_H$&1.08e-01($\sim$)	&6.90e-02	&5.28e-02	&2.56e-01	&1.17e-01	&7.22e-02	&6.11e-02	&2.56e-01	\\
$ZZJ9$	&$IGD$	&1.12e-02($\sim$)	&9.16e-03	&3.56e-03	&3.62e-02	&1.06e-02	&6.36e-03	&3.51e-03	&2.57e-02	\\
&$I^-_H$&1.98e-02($\sim$)	&1.47e-02	&6.51e-03	&5.95e-02	&1.91e-02	&1.03e-02	&6.65e-03	&4.35e-02	\\
$ZZJ10$	&$IGD$	&9.11e+00($\sim$)	&5.75e+00	&1.34e+00	&2.40e+01	&9.91e+00	&4.97e+00	&2.66e+00	&1.86e+01	\\
&$I^-_H$&1.11e+00($\sim$)	&2.26e-16	&1.11e+00	&1.11e+00	&1.11e+00	&2.26e-16	&1.11e+00	&1.11e+00	\\
\hline
$+/-/\sim$ &$IGD$	&5/0/5	&&&&&&\\
$+/-/\sim$ &$I^-_H$ &5/0/5		&&&&&&\\
\hline\hline
\end{tabular}
\end{table*}

\begin{figure*}[htbp]
\centering
\begin{subfigure}[t]{0.38\columnwidth}
    \includegraphics[ width=\columnwidth]{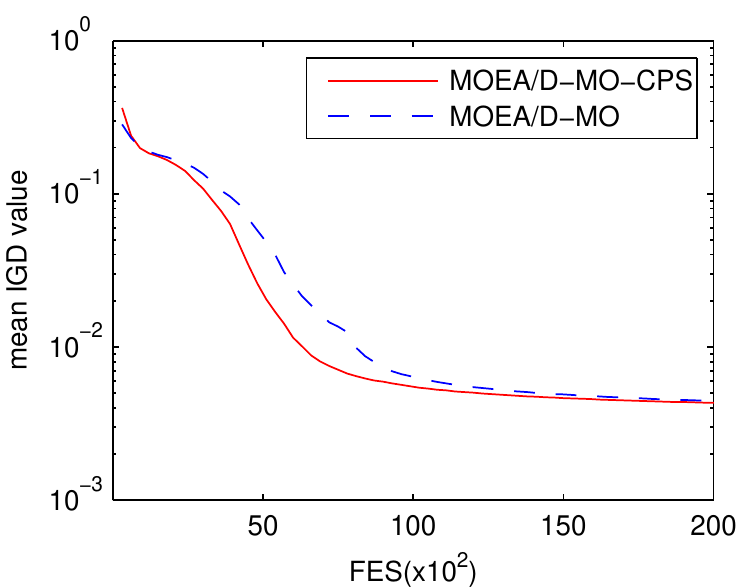}
    \subcaption{ZZJ1}
\end{subfigure}
\begin{subfigure}[t]{0.38\columnwidth}
    \includegraphics[ width=\columnwidth]{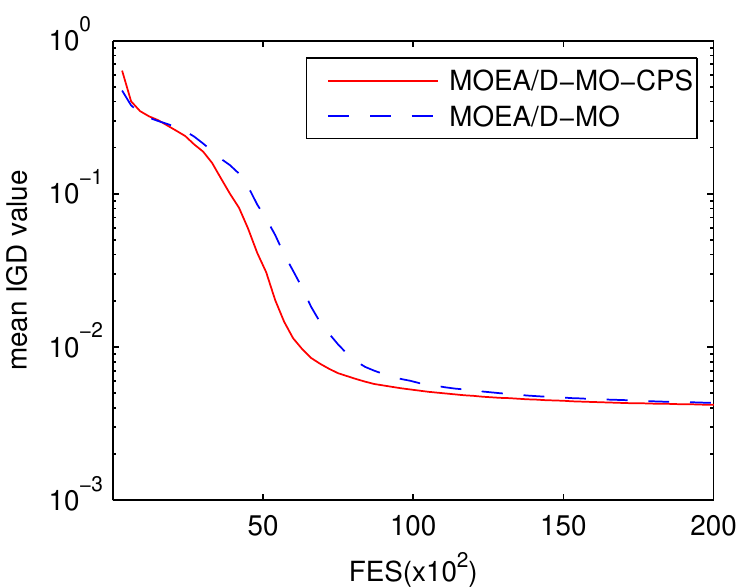}
    \subcaption{ZZJ2}
\end{subfigure}
\begin{subfigure}[t]{0.38\columnwidth}
    \includegraphics[ width=\columnwidth]{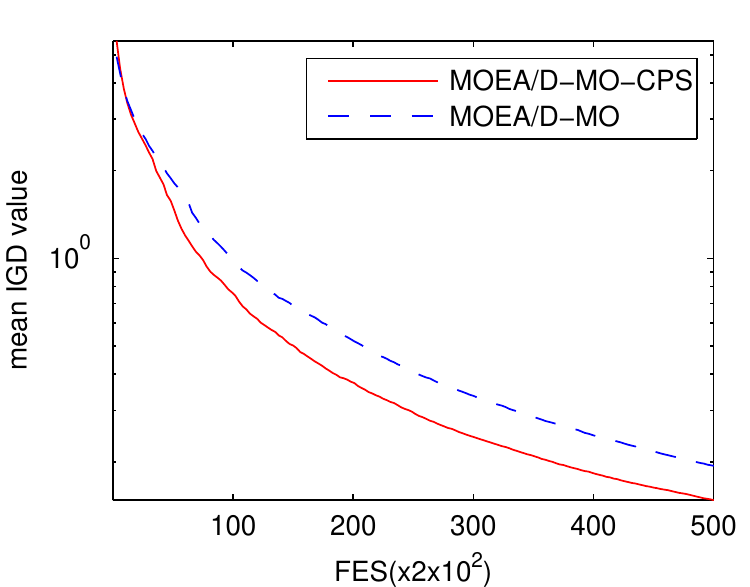}
    \subcaption{ZZJ3}
\end{subfigure}
\begin{subfigure}[t]{0.38\columnwidth}
    \includegraphics[ width=\columnwidth]{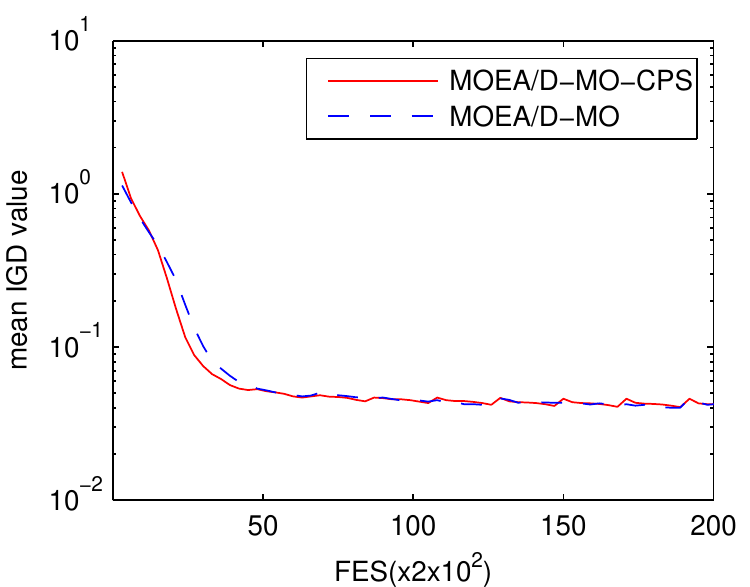}
    \subcaption{ZZJ4}
\end{subfigure}
\begin{subfigure}[t]{0.38\columnwidth}
    \includegraphics[ width=\columnwidth]{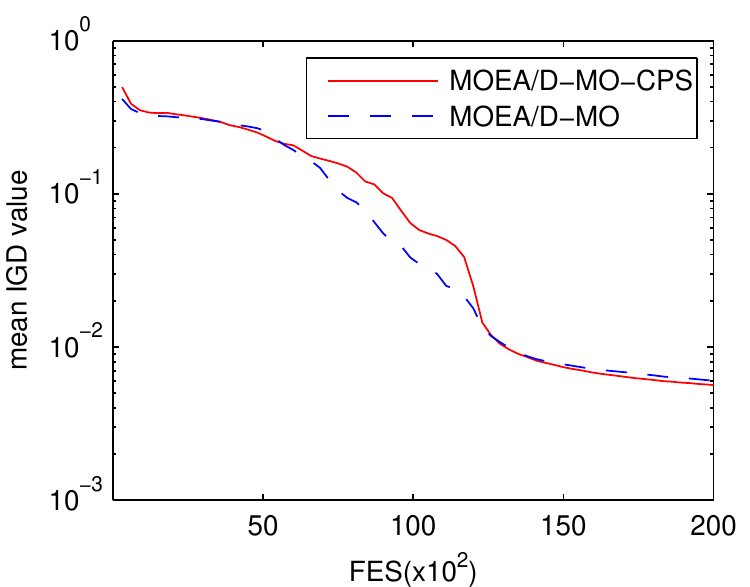}
    \subcaption{ZZJ5}
\end{subfigure}
\begin{subfigure}[t]{0.38\columnwidth}
    \includegraphics[ width=\columnwidth]{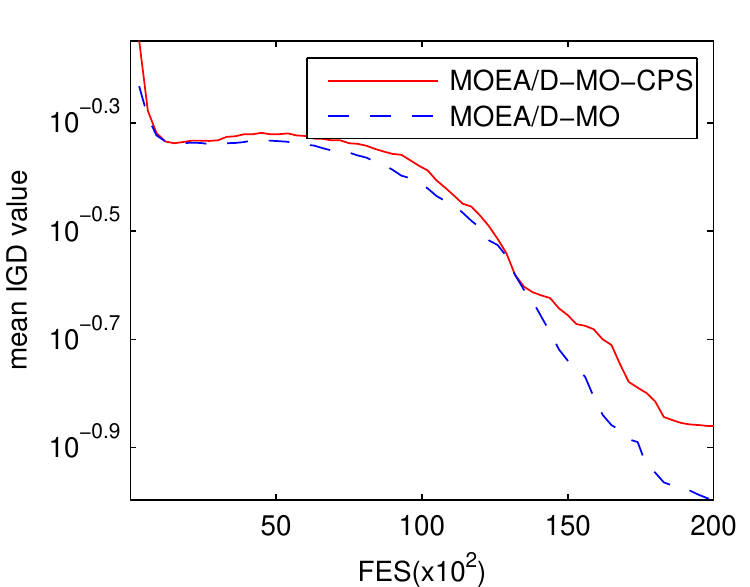}
    \subcaption{ZZJ6}
\end{subfigure}
\begin{subfigure}[t]{0.38\columnwidth}
    \includegraphics[ width=\columnwidth]{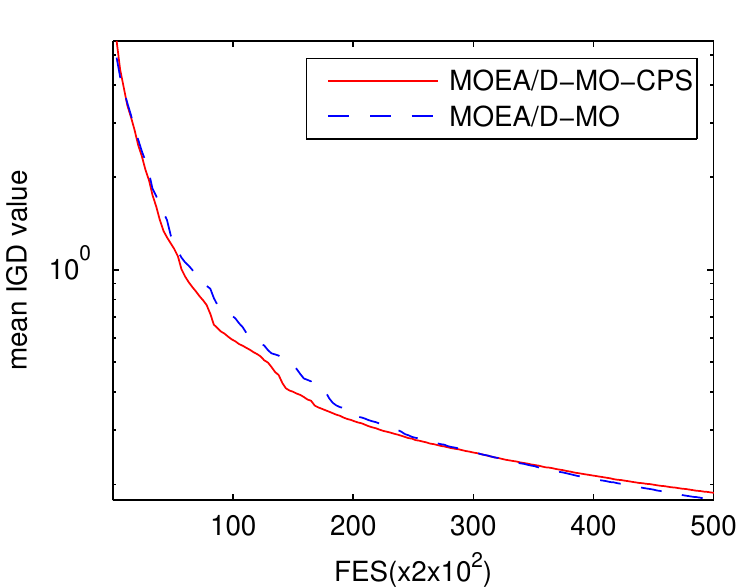}
    \subcaption{ZZJ7}
\end{subfigure}
\begin{subfigure}[t]{0.38\columnwidth}
    \includegraphics[ width=\columnwidth]{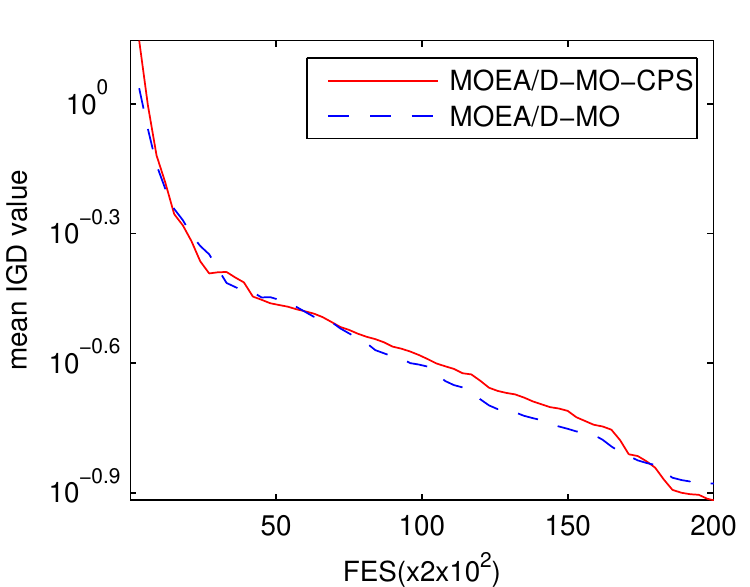}
    \subcaption{ZZJ8}
\end{subfigure}
\begin{subfigure}[t]{0.38\columnwidth}
    \includegraphics[ width=\columnwidth]{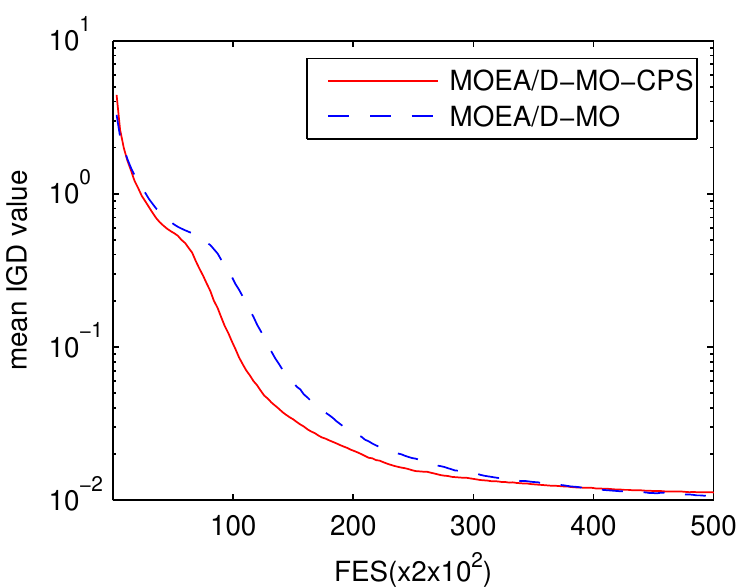}
    \subcaption{ZZJ9}
\end{subfigure}
\begin{subfigure}[t]{0.38\columnwidth}
    \includegraphics[ width=\columnwidth]{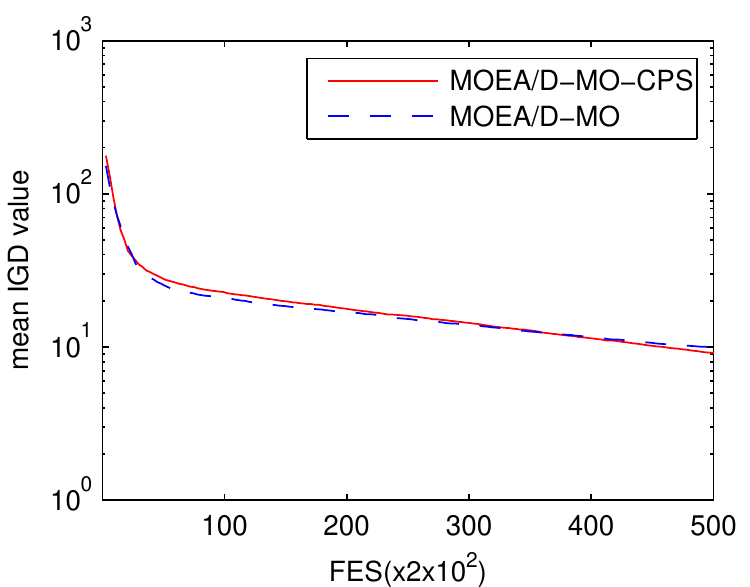}
    \subcaption{ZZJ10}
\end{subfigure}
\caption {The mean IGD values versus FEs obtained by MOEA/D-MO-CPS and MOEA/D-MO over 30 runs}
\label{fig:moead}
\end{figure*}

\begin{figure*}[htbp]
\centering
\begin{subfigure}[t]{0.38\columnwidth}
    \includegraphics[ width=\columnwidth]{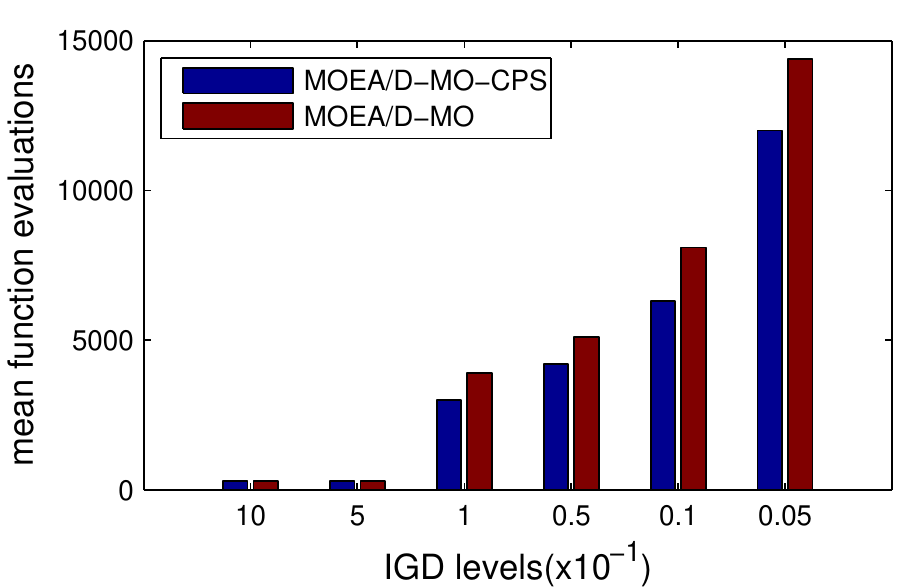}
    \subcaption{ZZJ1}
\end{subfigure}
\begin{subfigure}[t]{0.38\columnwidth}
    \includegraphics[ width=\columnwidth]{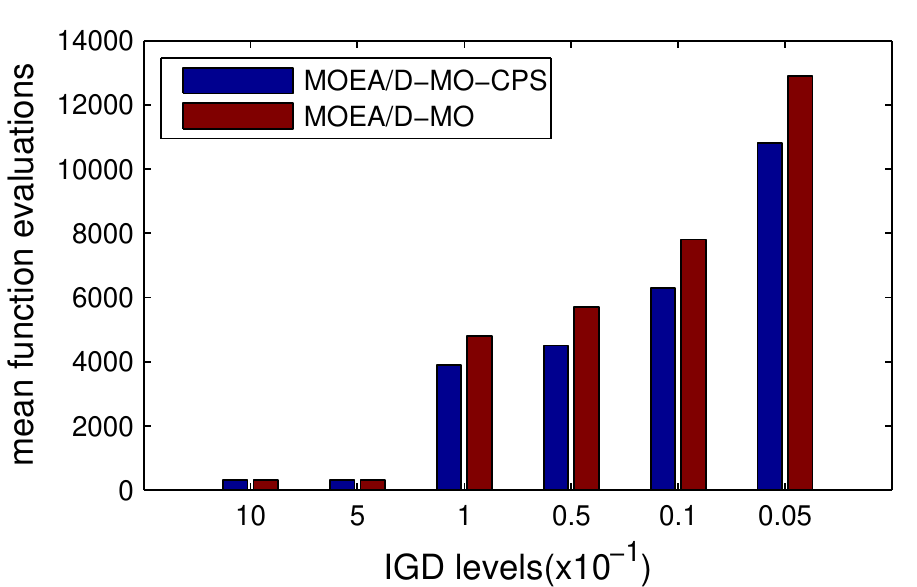}
    \subcaption{ZZJ2}
\end{subfigure}
\begin{subfigure}[t]{0.38\columnwidth}
    \includegraphics[ width=\columnwidth]{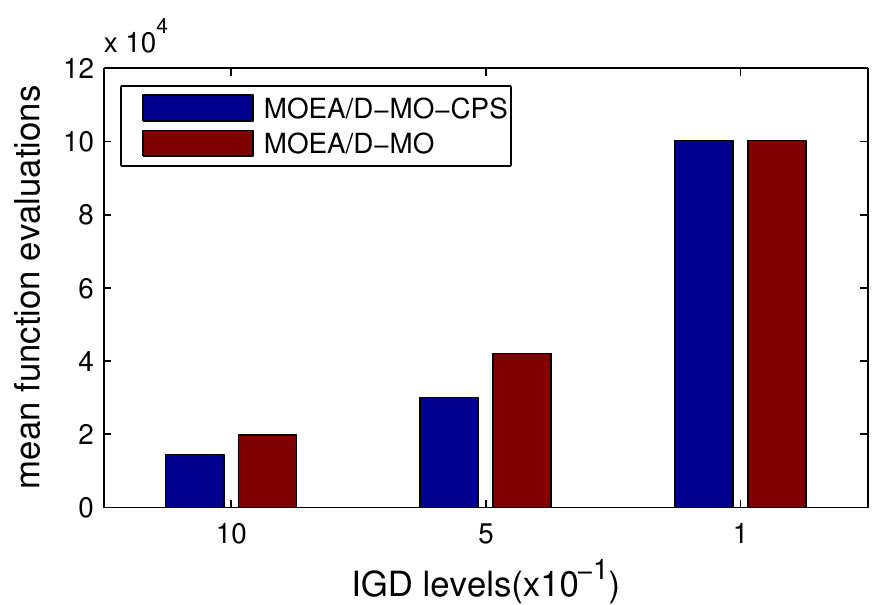}
    \subcaption{ZZJ3}
\end{subfigure}
\begin{subfigure}[t]{0.38\columnwidth}
    \includegraphics[ width=\columnwidth]{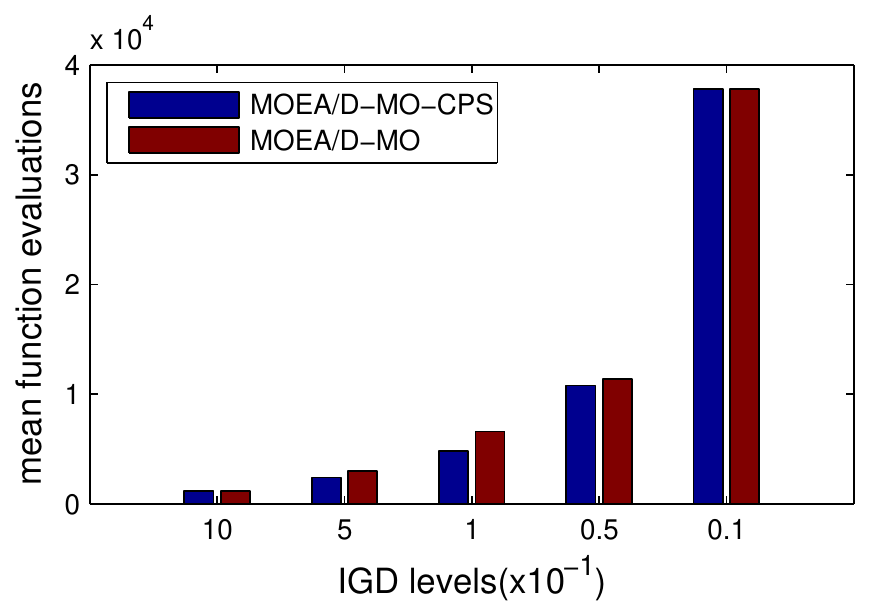}
    \subcaption{ZZJ4}
\end{subfigure}
\begin{subfigure}[t]{0.38\columnwidth}
    \includegraphics[ width=\columnwidth]{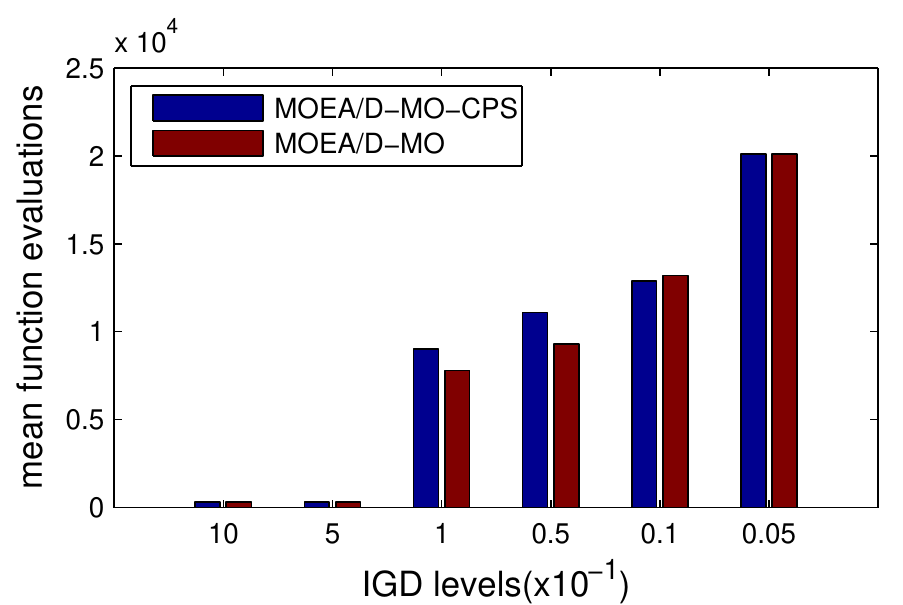}
    \subcaption{ZZJ5}
\end{subfigure}
\begin{subfigure}[t]{0.38\columnwidth}
    \includegraphics[ width=\columnwidth]{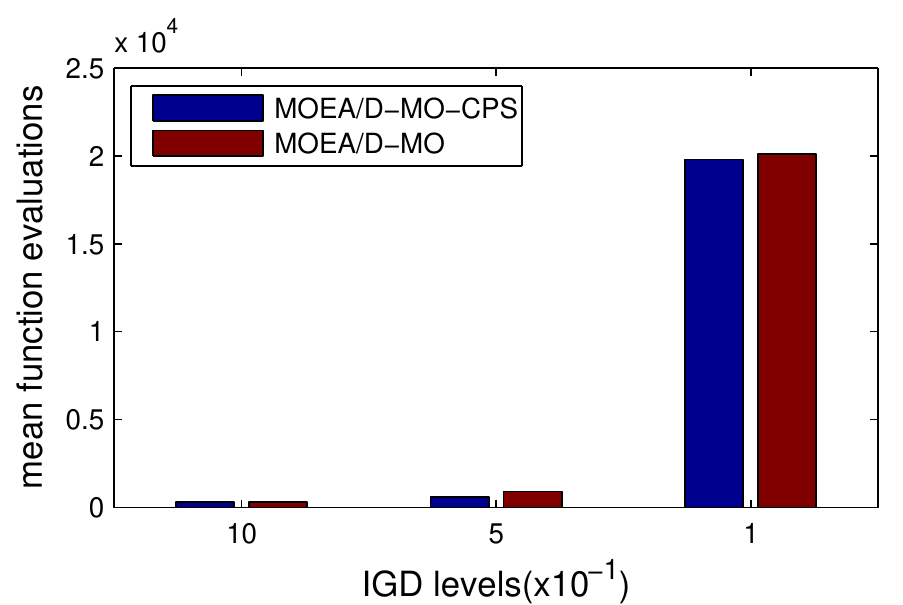}
    \subcaption{ZZJ6}
\end{subfigure}
\begin{subfigure}[t]{0.38\columnwidth}
    \includegraphics[ width=\columnwidth]{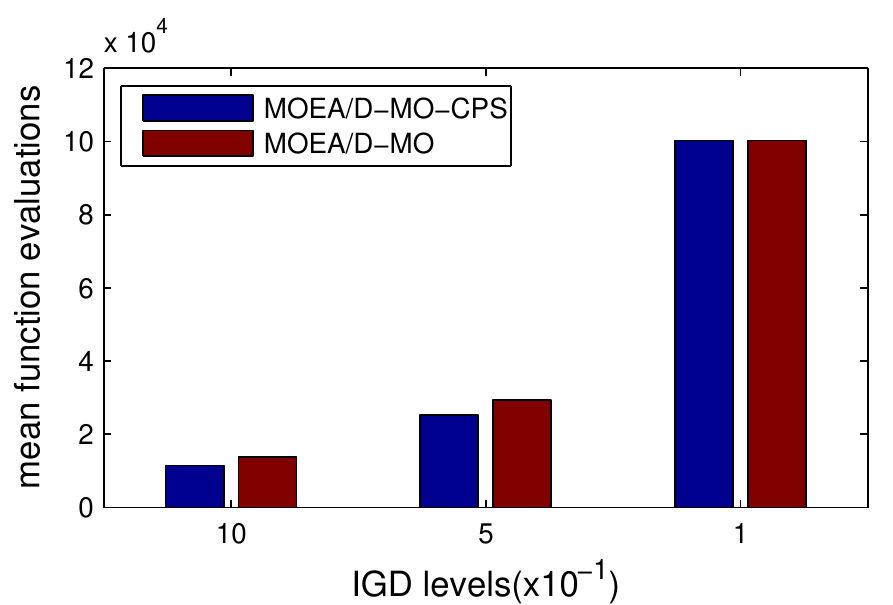}
    \subcaption{ZZJ7}
\end{subfigure}
\begin{subfigure}[t]{0.38\columnwidth}
    \includegraphics[ width=\columnwidth]{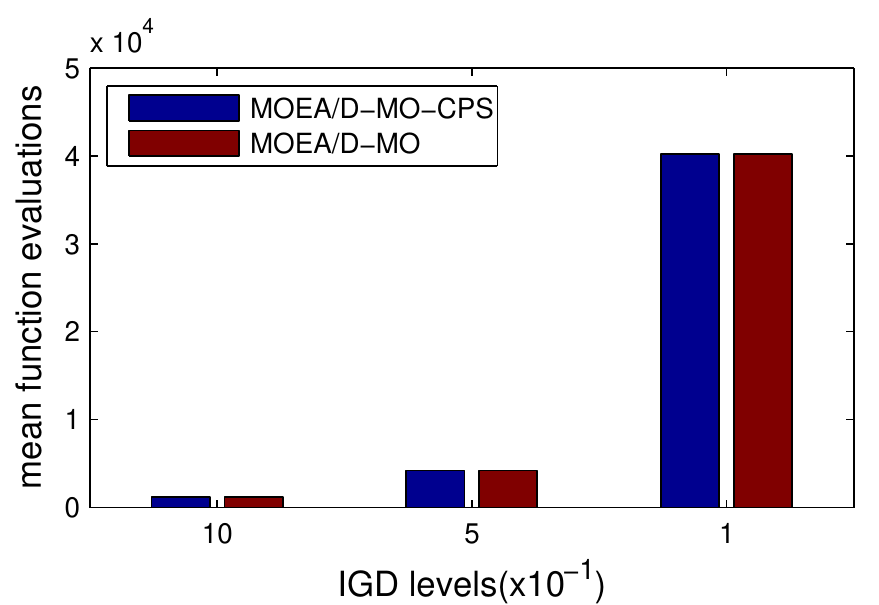}
    \subcaption{ZZJ8}
\end{subfigure}
\begin{subfigure}[t]{0.38\columnwidth}
    \includegraphics[ width=\columnwidth]{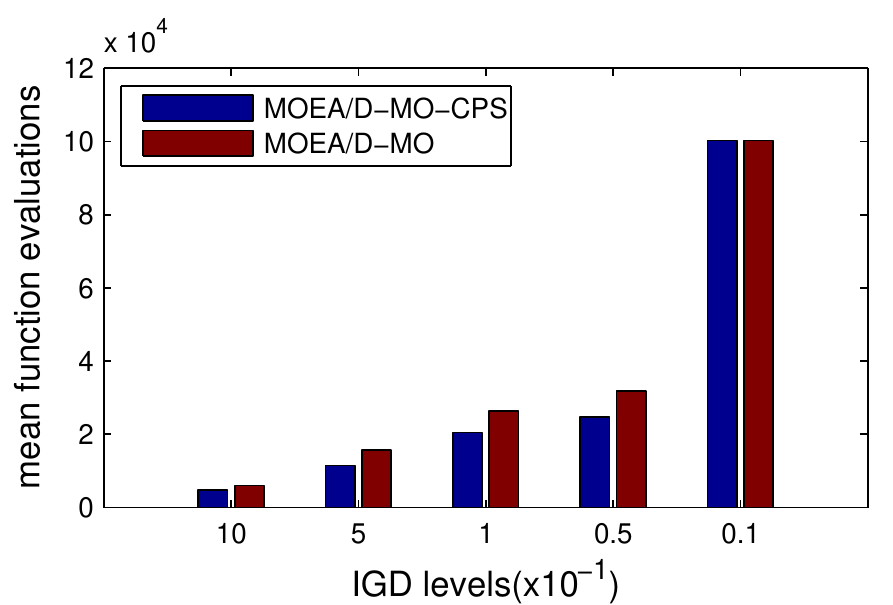}
    \subcaption{ZZJ9}
\end{subfigure}
\begin{subfigure}[t]{0.38\columnwidth}
    \includegraphics[ width=\columnwidth]{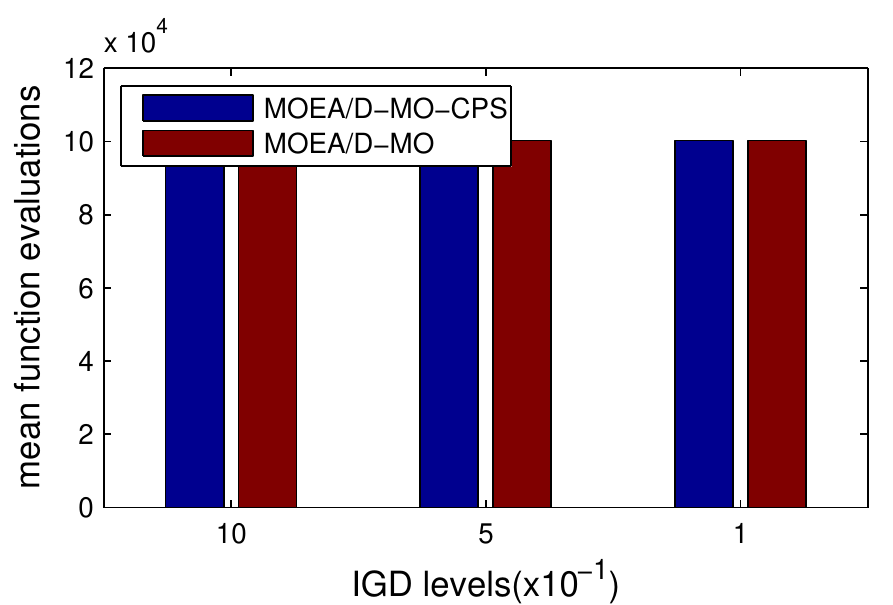}
    \subcaption{ZZJ10}
\end{subfigure}
\caption {The mean FEs required by MOEA/D-MO-CPS and MOEA/D-MO to obtain different IGD values over 30 runs}
\label{fig:moeadbar}
\end{figure*}

\begin{figure*}[htbp]
\centering
\begin{subfigure}[t]{0.64\columnwidth}
    \includegraphics[ width=0.50\columnwidth,height=2.3cm]{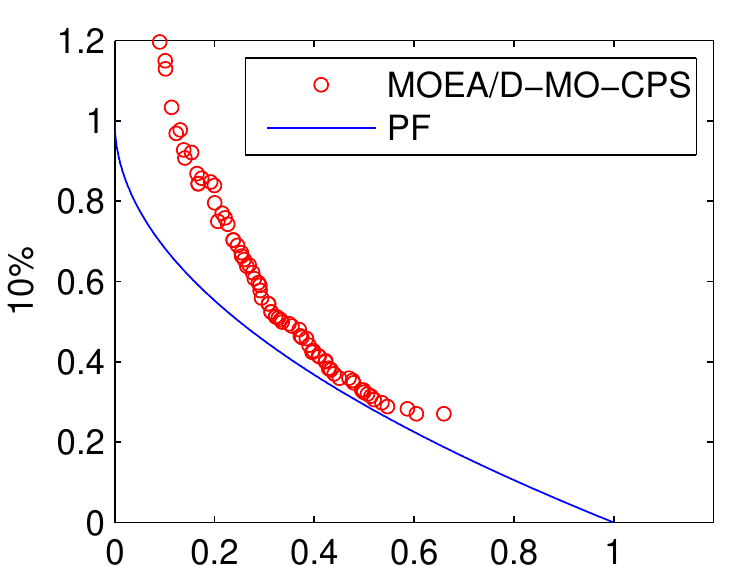}
    \includegraphics[ width=0.46\columnwidth,height=2.3cm]{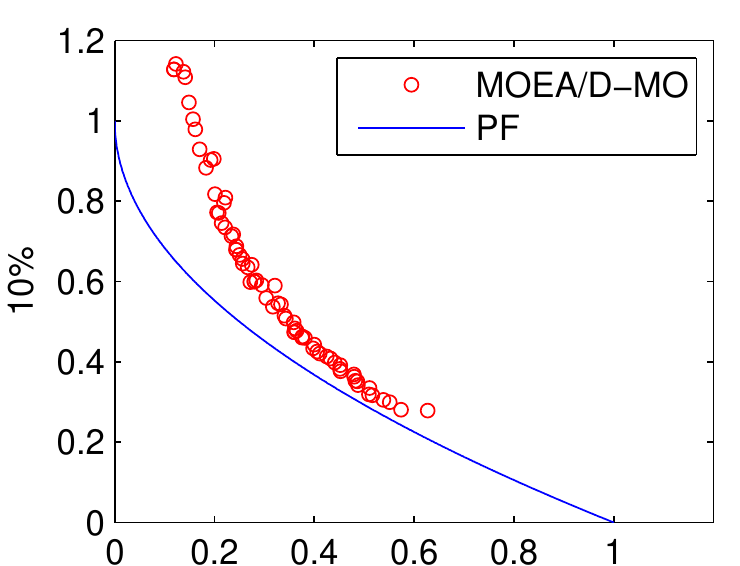}

    \includegraphics[ width=0.50\columnwidth,height=2.3cm]{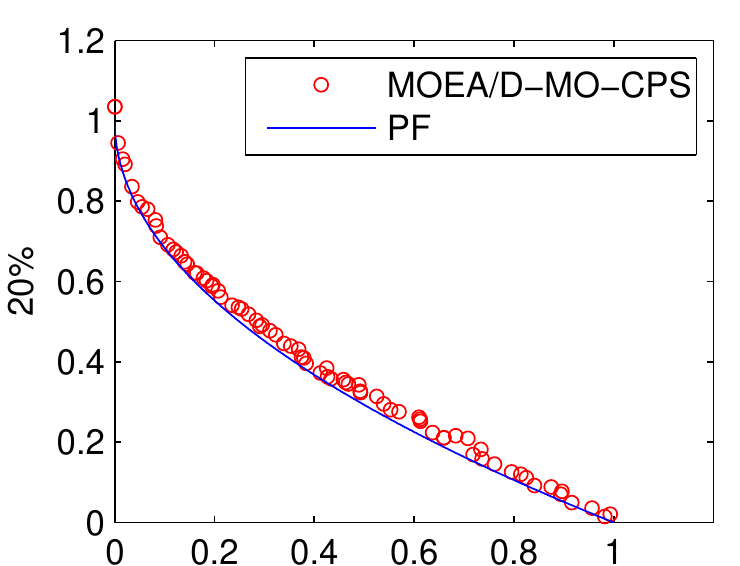}
    \includegraphics[ width=0.46\columnwidth,height=2.3cm]{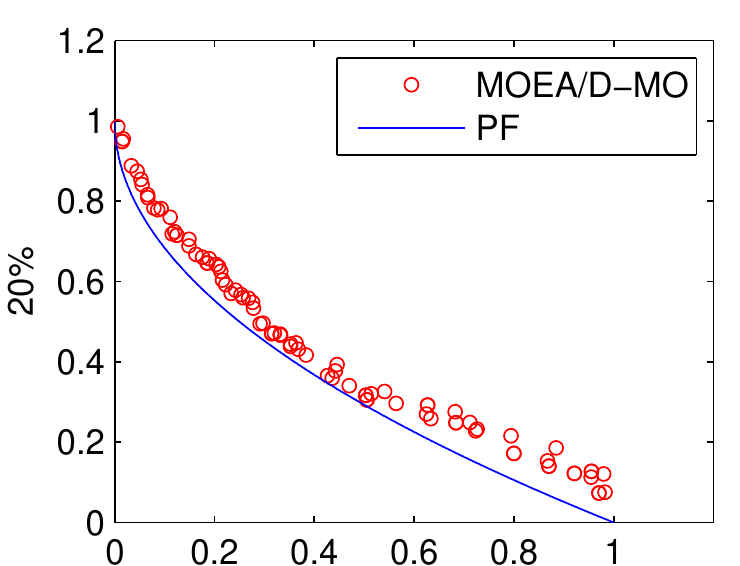}

    \includegraphics[ width=0.50\columnwidth,height=2.3cm]{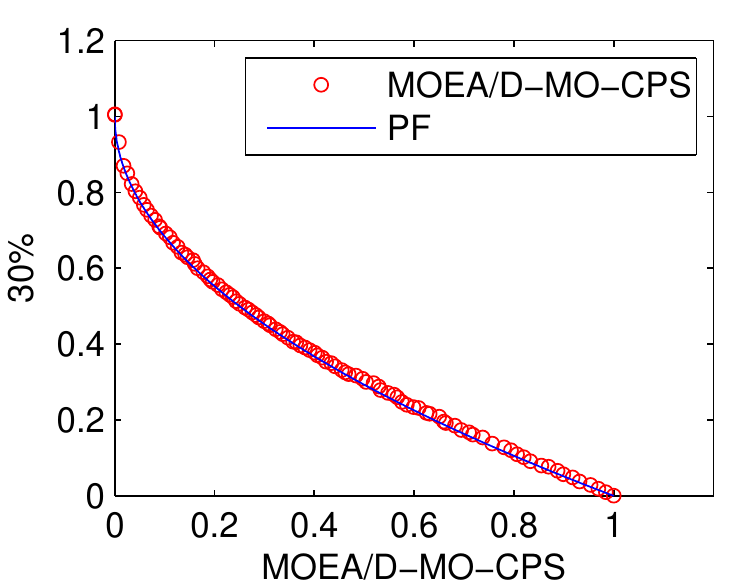}
    \includegraphics[ width=0.46\columnwidth,height=2.3cm]{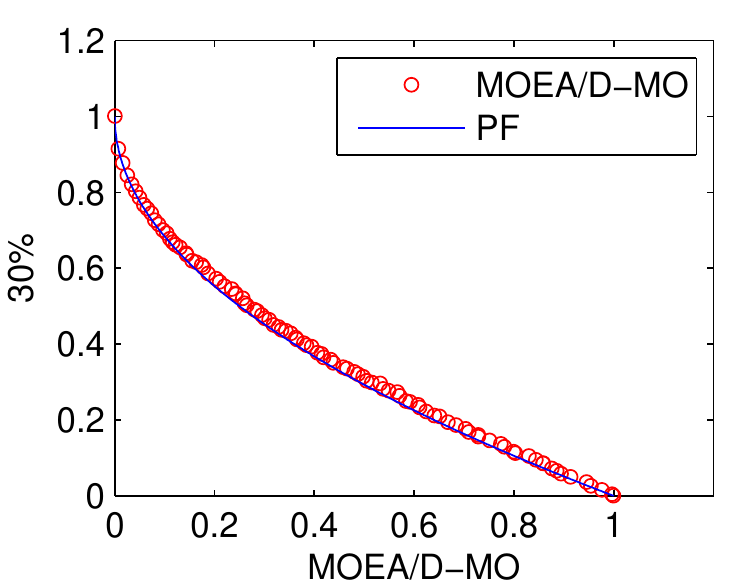}
    \subcaption{ZZJ1}
\end{subfigure}
\begin{subfigure}[t]{0.64\columnwidth}
    \includegraphics[ width=0.50\columnwidth,height=2.3cm]{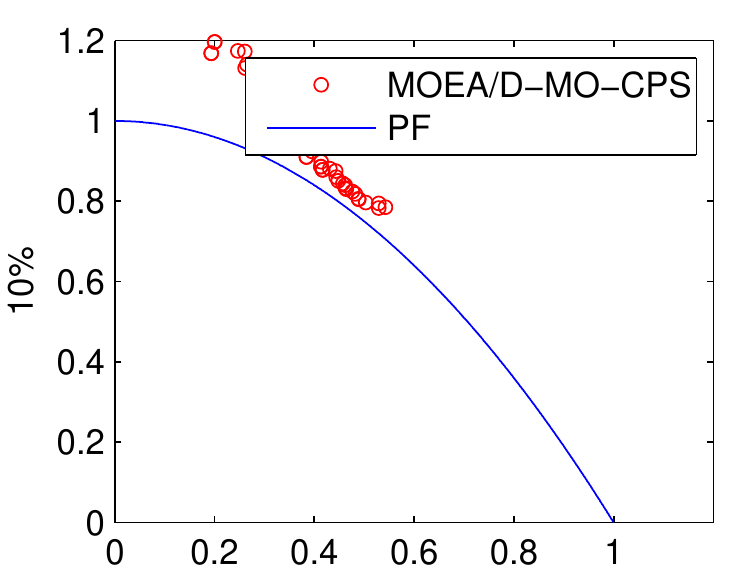}
    \includegraphics[ width=0.46\columnwidth,height=2.3cm]{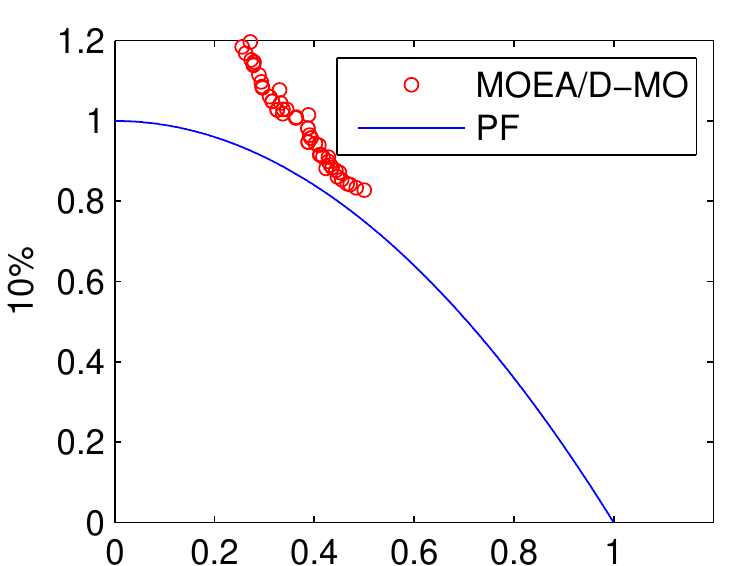}

    \includegraphics[ width=0.50\columnwidth,height=2.3cm]{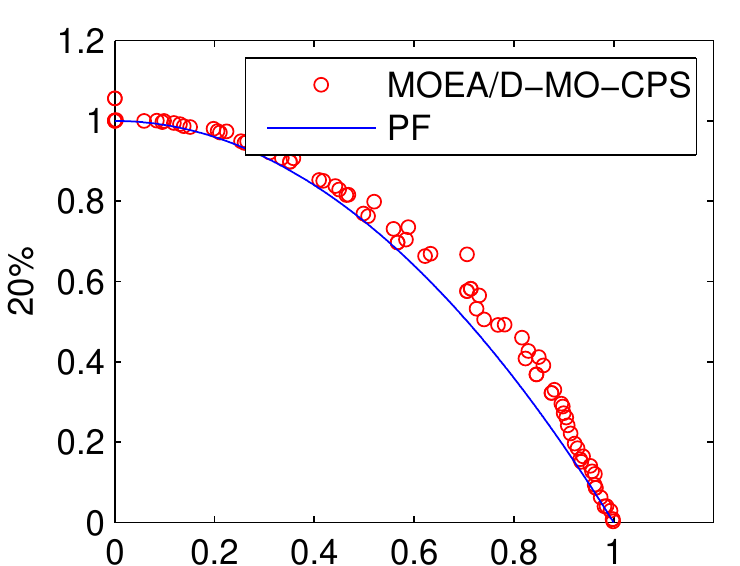}
    \includegraphics[ width=0.46\columnwidth,height=2.3cm]{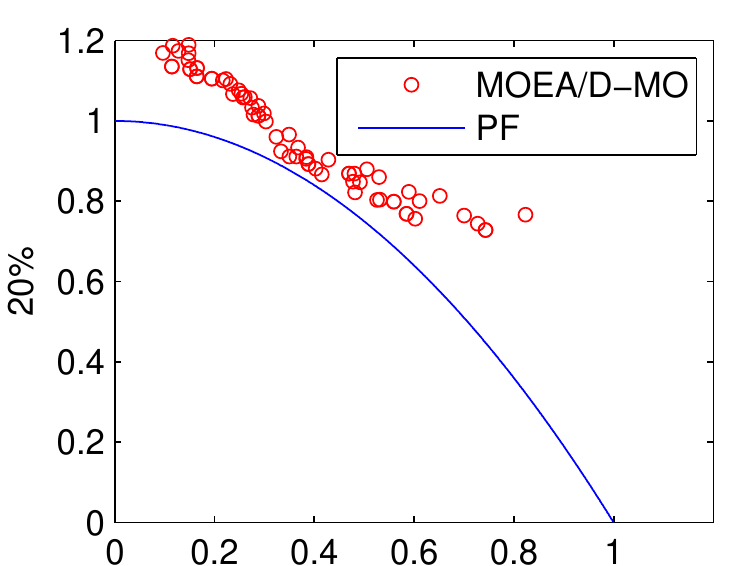}

    \includegraphics[ width=0.50\columnwidth,height=2.3cm]{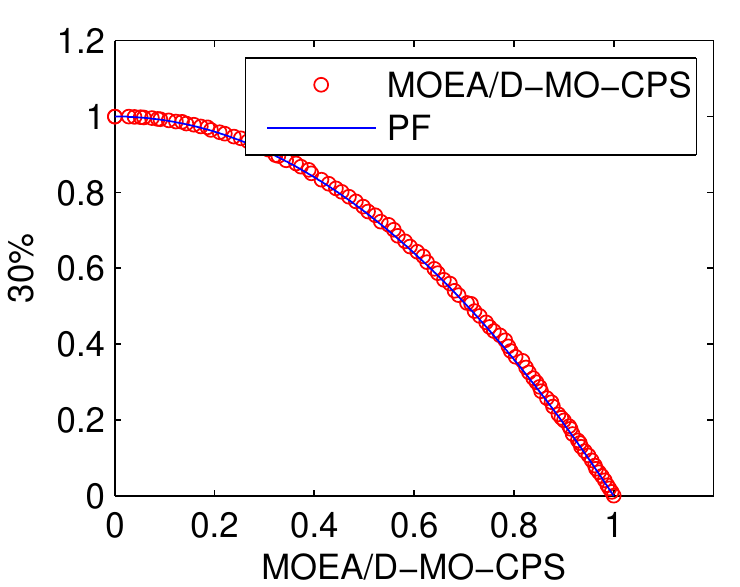}
    \includegraphics[ width=0.46\columnwidth,height=2.3cm]{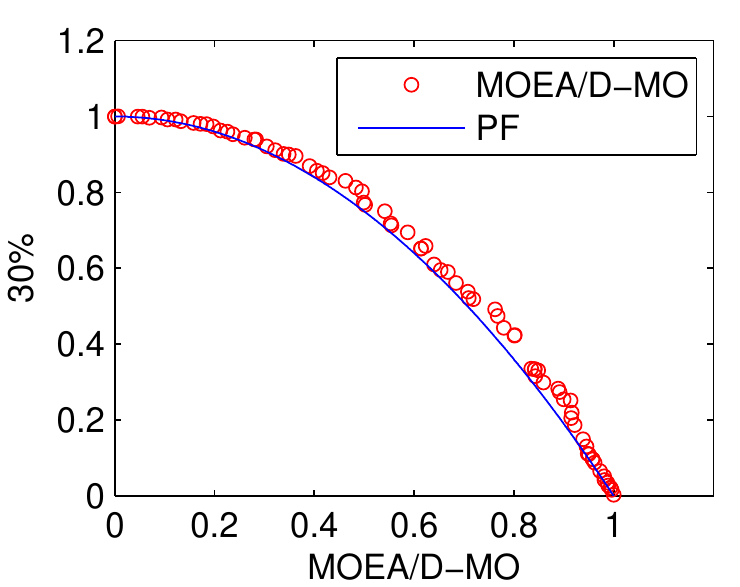}
    \subcaption{ZZJ2}
\end{subfigure}
\begin{subfigure}[t]{0.64\columnwidth}
    \includegraphics[ width=0.50\columnwidth,height=2.3cm]{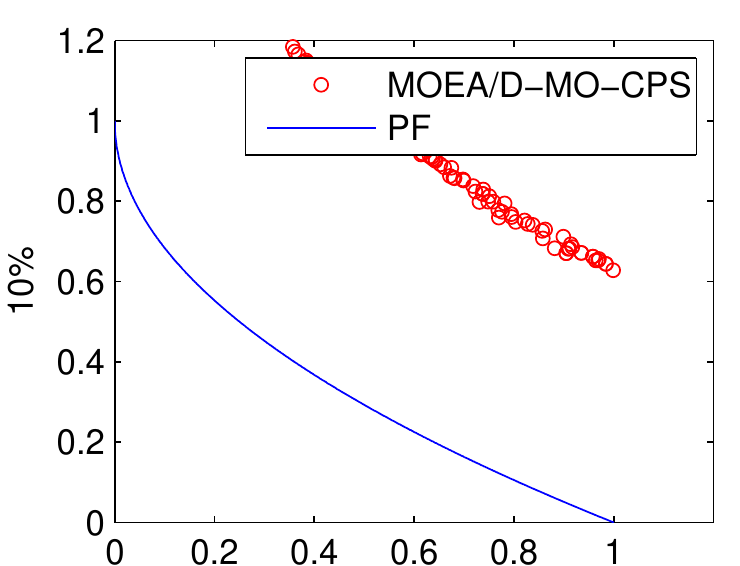}
    \includegraphics[ width=0.46\columnwidth,height=2.3cm]{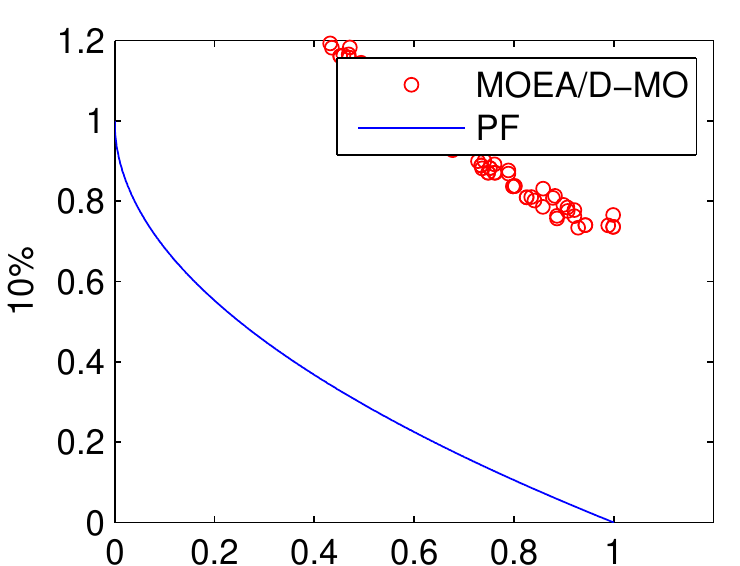}

    \includegraphics[ width=0.50\columnwidth,height=2.3cm]{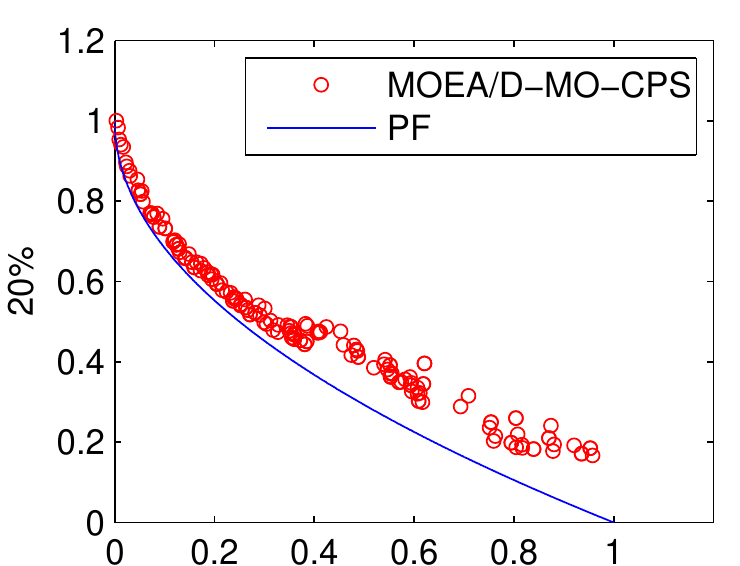}
    \includegraphics[ width=0.46\columnwidth,height=2.3cm]{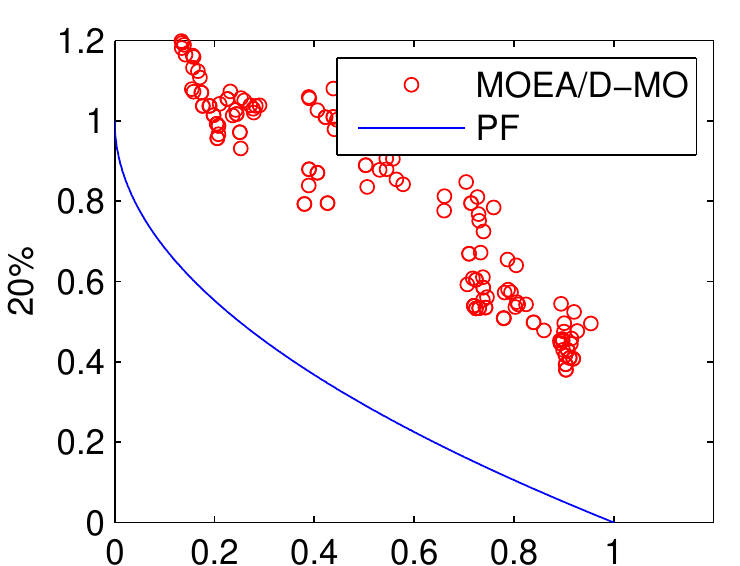}

    \includegraphics[ width=0.48\columnwidth,height=2.3cm]{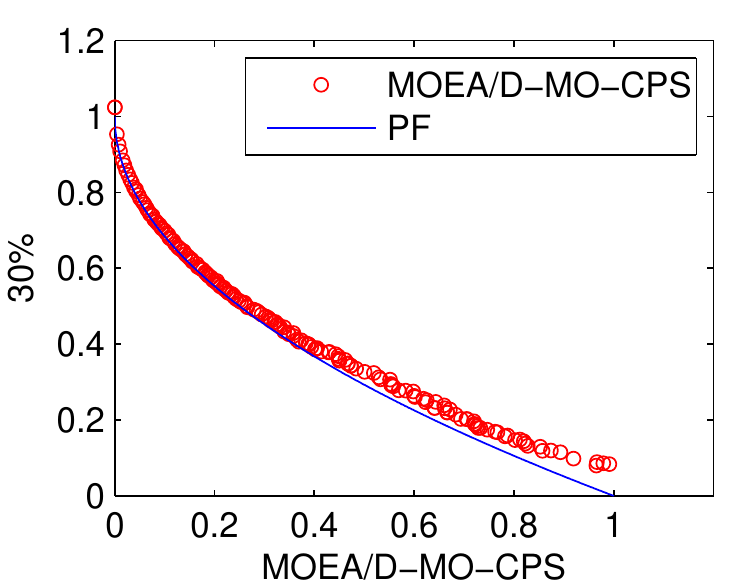}
    \includegraphics[ width=0.48\columnwidth,height=2.3cm]{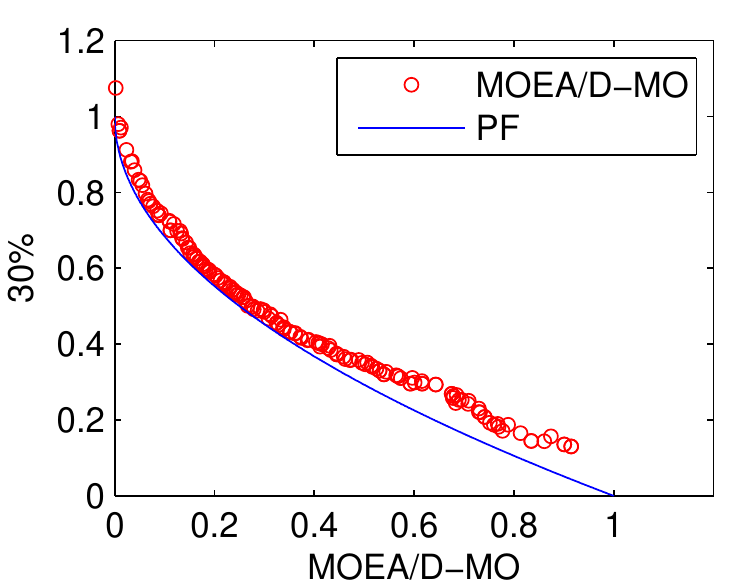}
    \subcaption{ZZJ9}
\end{subfigure}
\caption {The median (according to the IGD metric values) approximations obtained by MOEA/D-MO-CPS and MOEA/D-MO after 10\%, 20\%, 30\% of the max FEs on (a) ZZJ1, (b) ZZJ2, and (c) ZZJ9}
\label{fig:moeadplot}
\end{figure*}

Table~\ref{fig:moead} shows the statistical results of IGD and $I^-_H$ metric} values obtained by MOEA/D-MO-CPS and MOEA/D-MO on ZZJ1-ZZJ10 over 30 runs. The statistical tests according to both the IGD and the $I^-_H$ metric metrics are consistent with each other. It shows on ZZJ1-ZZJ5, MOEA/D-MO-CPS outperforms MOEA/D-MO, and on ZZJ6-ZZJ10, the two algorithms performs similar with each other. This suggests that with the given FEs, MOEA/D-MO-CPS works no worse than MOEA/D-MO. The reason is that CPS helps MOEA/D-MO to obtain better results on some instances.

Fig.~\ref{fig:moead} presents the run time performance in terms of the IGD values obtained by MOEA/D-MO-CPS and MOEA/D-MO on the 10 instances. The curves obtained by the two algorithms in Fig.~\ref{fig:moead} show that on most of the instances MOEA/D-MO-CPS converges faster than MOEA/D-MO. Only on ZZJ6, ZZJ8, and ZZJ10, MOEA/D-MO-CPS converges slower than MOEA/D-MO. Fig.~\ref{fig:moeadbar} plots the FEs required by the two algorithms to obtain some levels of IGD values on the 10 instances. It suggests that to obtain the same IGD values, MOEA/D-MO-CPS uses fewer computational resources than MOEA/D-MO on most of the instances.

To do a visual comparison, Fig.~\ref{fig:moeadplot} plots the final obtained populations of the median runs according to the IGD values after 10\%, 20\%, and 30\% of the max FEs for ZZJ1, ZZJ2 and ZZJ9. The figure indicates that MOEA/D-MO-CPS can achieve better results than MOEA/D-MO with the same computational costs on most of the instances.

\subsubsection{More Discussion}
The experiments in the previous sections clearly show that CPS can successfully improve the performances of the three algorithms on most of the given test instances according to both the quality of the final obtained solutions and the algorithm convergence. This section investigates why CPS works well.

%The \emph{Generational Distance (GD)}, i.e., the distance from a solution to a reference PF, is used to measure the quality of solutions.

\begin{figure*}[htbp]
\centering
\begin{subfigure}[t]{0.38\columnwidth}
    \includegraphics[ width=\columnwidth]{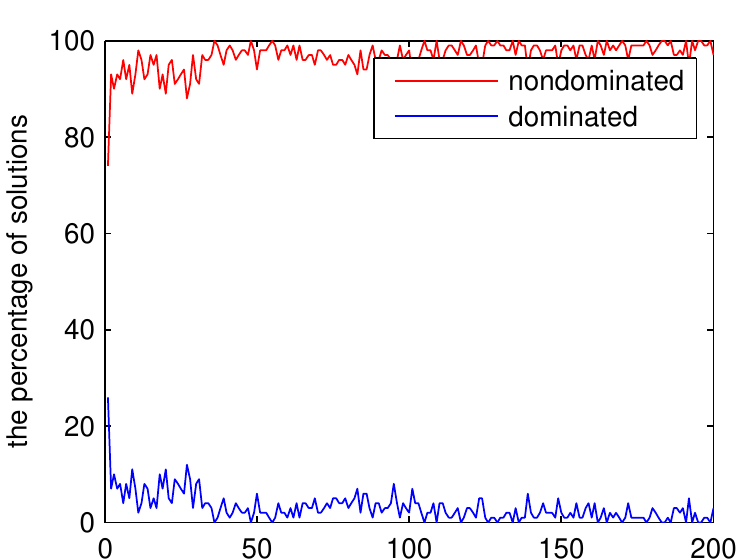}
    \subcaption{ZZJ1}
\end{subfigure}
\begin{subfigure}[t]{0.38\columnwidth}
    \includegraphics[ width=\columnwidth]{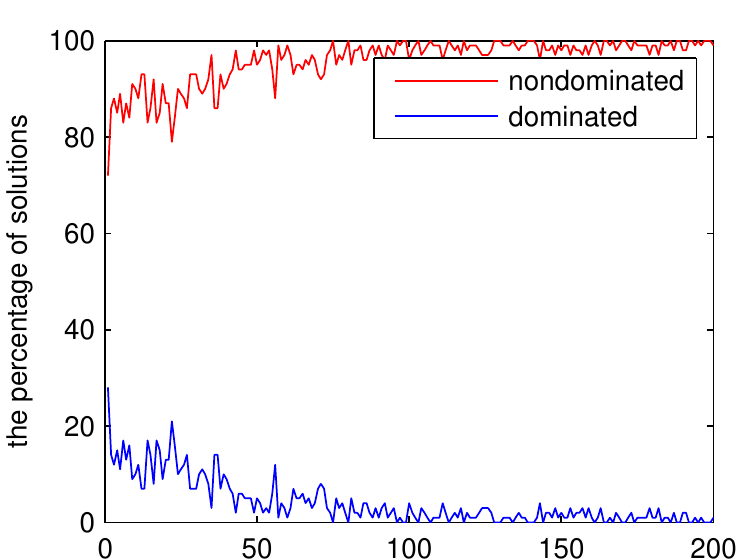}
    \subcaption{ZZJ2}
\end{subfigure}
\begin{subfigure}[t]{0.38\columnwidth}
    \includegraphics[ width=\columnwidth]{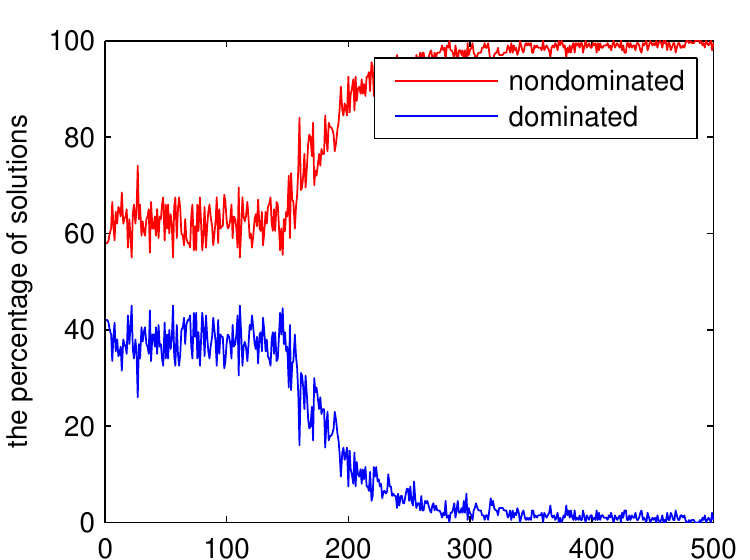}
    \subcaption{ZZJ3}
\end{subfigure}
\begin{subfigure}[t]{0.38\columnwidth}
    \includegraphics[ width=\columnwidth]{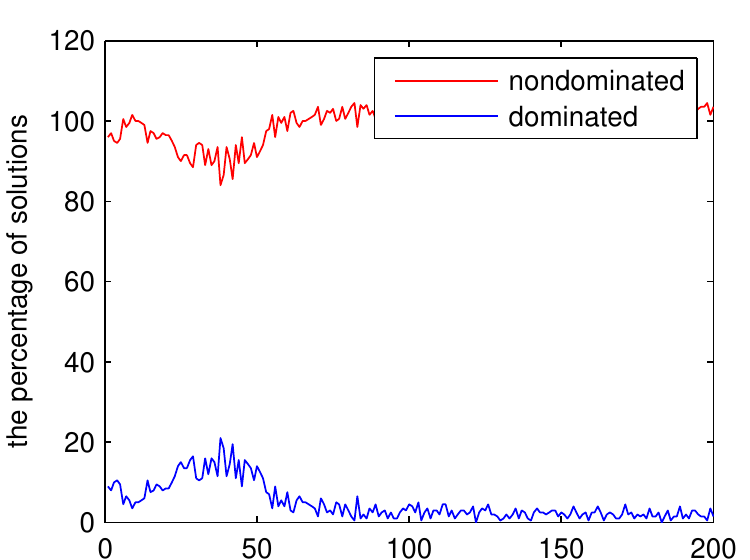}
    \subcaption{ZZJ4}
\end{subfigure}
\begin{subfigure}[t]{0.38\columnwidth}
    \includegraphics[ width=\columnwidth]{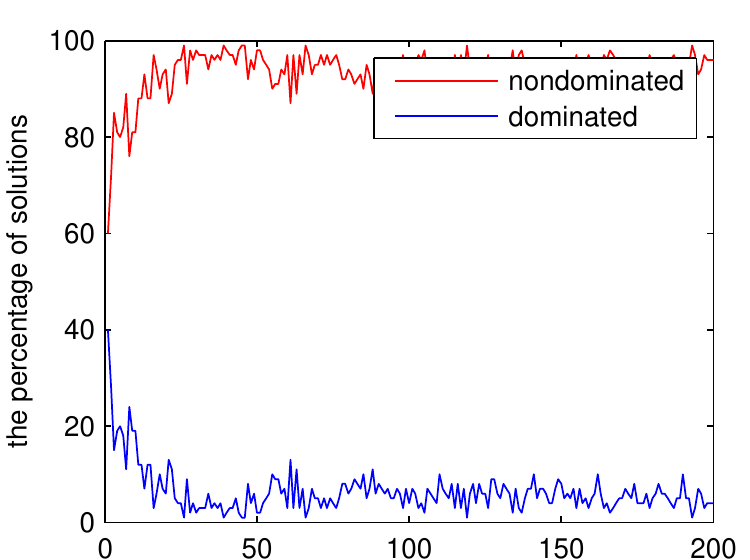}
    \subcaption{ZZJ5}
\end{subfigure}
\begin{subfigure}[t]{0.38\columnwidth}
    \includegraphics[ width=\columnwidth]{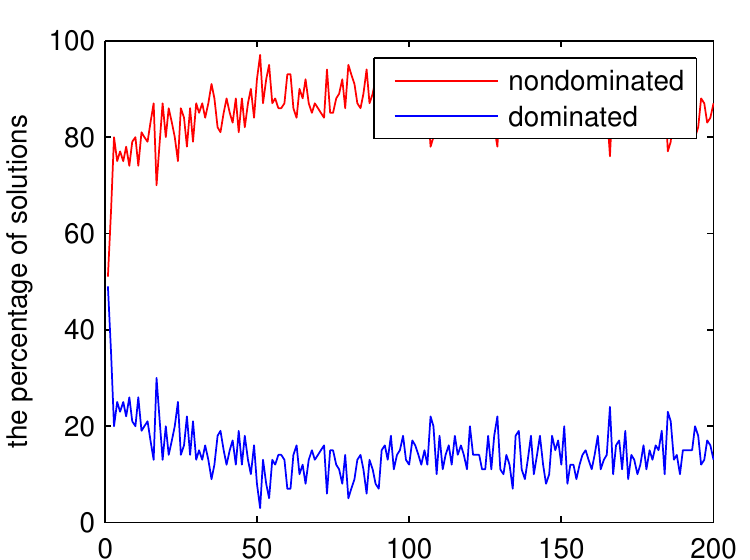}
    \subcaption{ZZJ6}
\end{subfigure}
\begin{subfigure}[t]{0.38\columnwidth}
    \includegraphics[ width=\columnwidth]{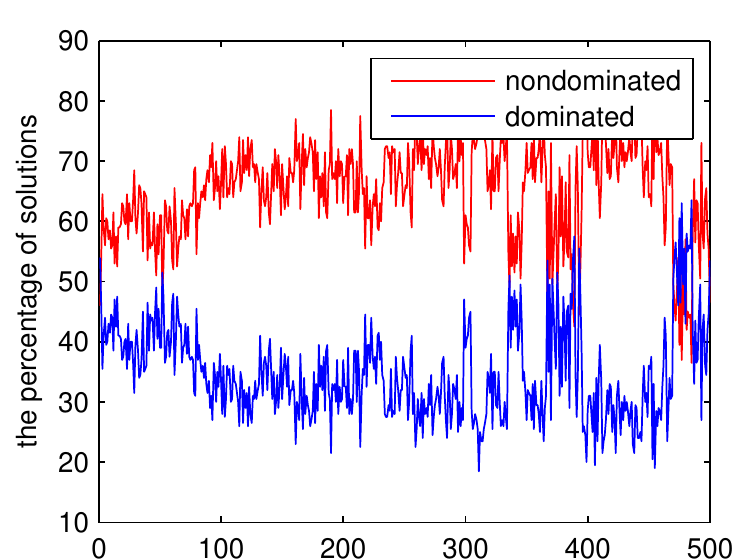}
    \subcaption{ZZJ7}
\end{subfigure}
\begin{subfigure}[t]{0.38\columnwidth}
    \includegraphics[ width=\columnwidth]{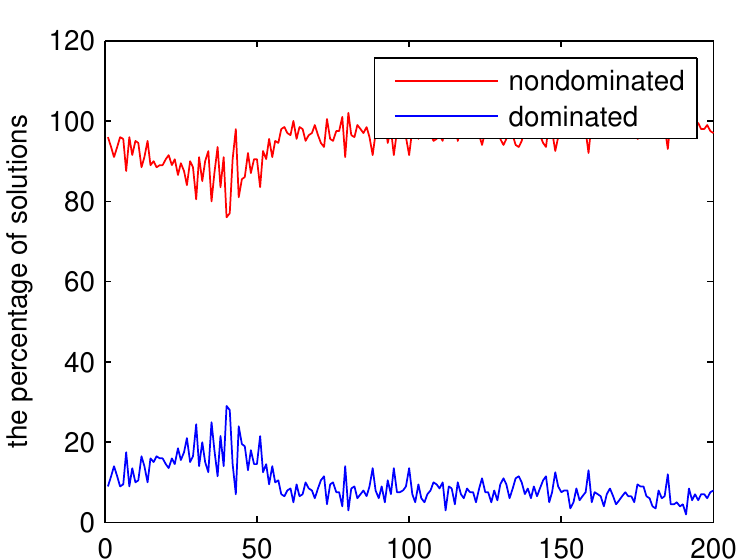}
    \subcaption{ZZJ8}
\end{subfigure}
\begin{subfigure}[t]{0.38\columnwidth}
    \includegraphics[ width=\columnwidth]{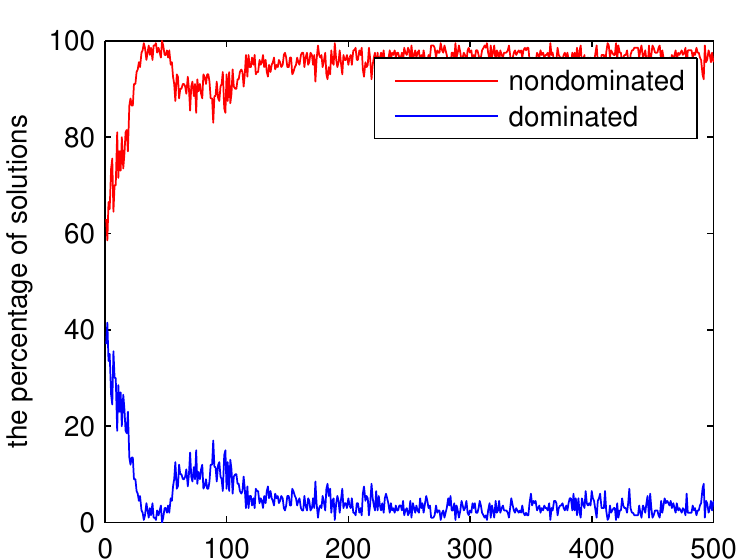}
    \subcaption{ZZJ9}
\end{subfigure}
\begin{subfigure}[t]{0.38\columnwidth}
    \includegraphics[ width=\columnwidth]{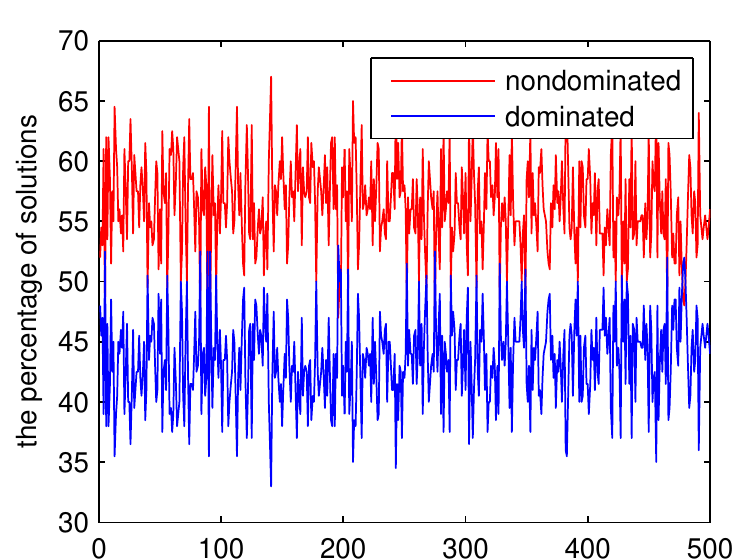}
    \subcaption{ZZJ10}
\end{subfigure}
\caption {The proportion of nondominated and dominated solutions selected by KNN in each generation for RM-MEDA}
\label{fig:rmnd}
\end{figure*}

\begin{figure*}[htbp]
\centering
\begin{subfigure}[t]{0.38\columnwidth}
    \includegraphics[ width=\columnwidth]{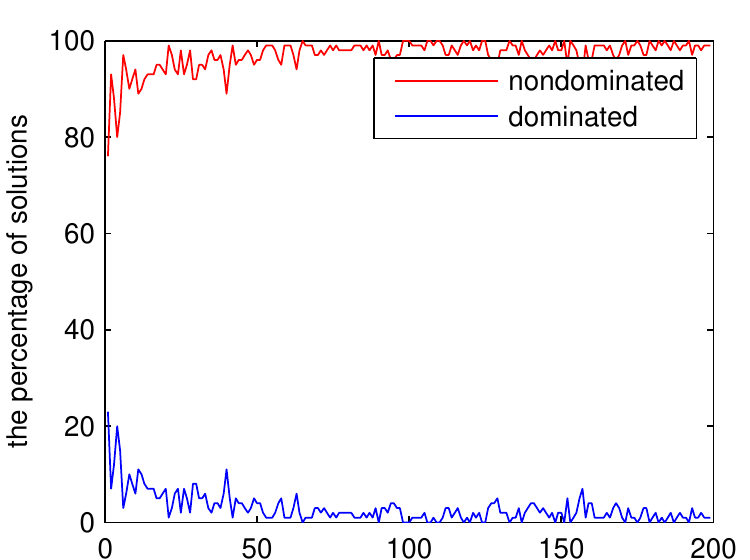}
    \subcaption{ZZJ1}
\end{subfigure}
\begin{subfigure}[t]{0.38\columnwidth}
    \includegraphics[ width=\columnwidth]{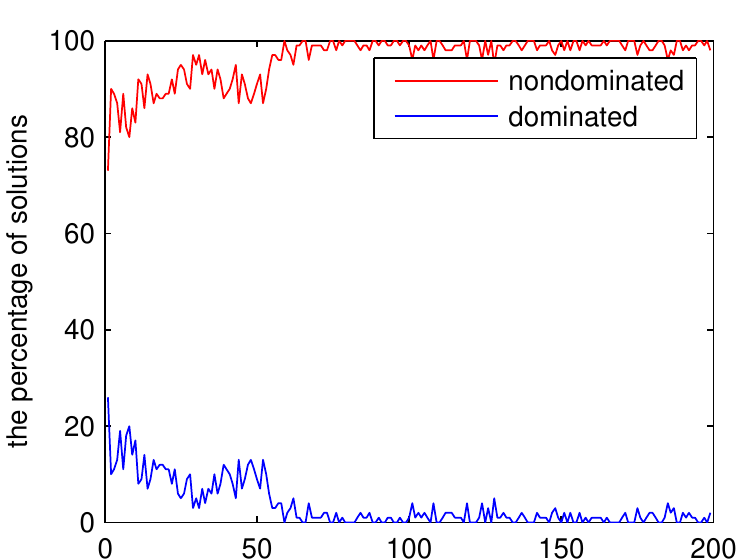}
    \subcaption{ZZJ2}
\end{subfigure}
\begin{subfigure}[t]{0.38\columnwidth}
    \includegraphics[ width=\columnwidth]{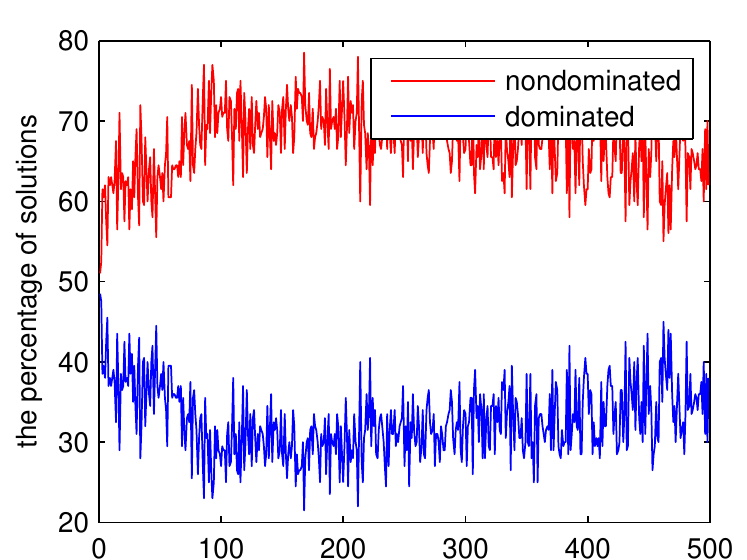}
    \subcaption{ZZJ3}
\end{subfigure}
\begin{subfigure}[t]{0.38\columnwidth}
    \includegraphics[ width=\columnwidth]{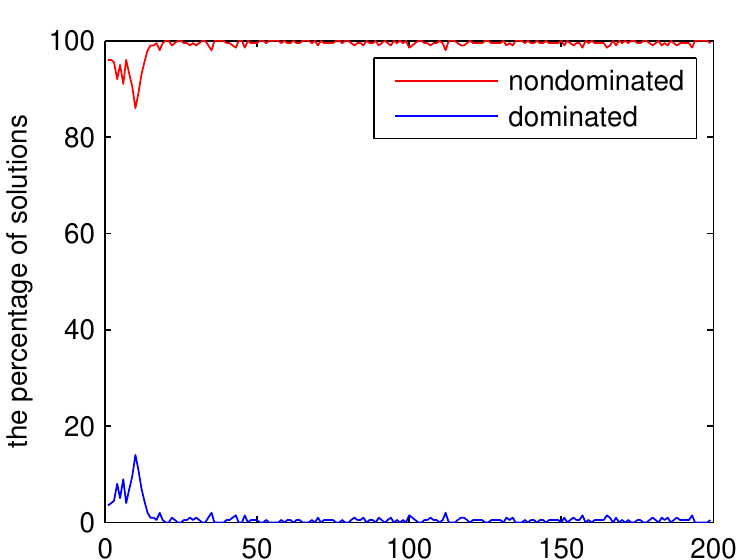}
    \subcaption{ZZJ4}
\end{subfigure}
\begin{subfigure}[t]{0.38\columnwidth}
    \includegraphics[ width=\columnwidth]{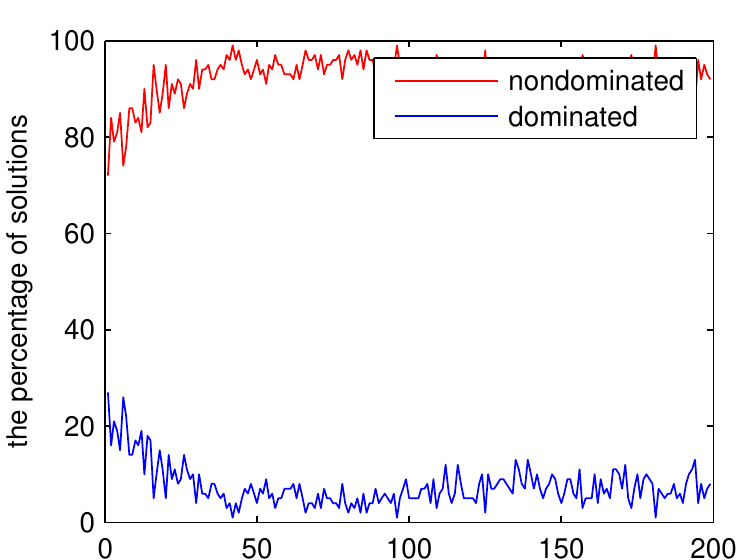}
    \subcaption{ZZJ5}
\end{subfigure}
\begin{subfigure}[t]{0.38\columnwidth}
    \includegraphics[ width=\columnwidth]{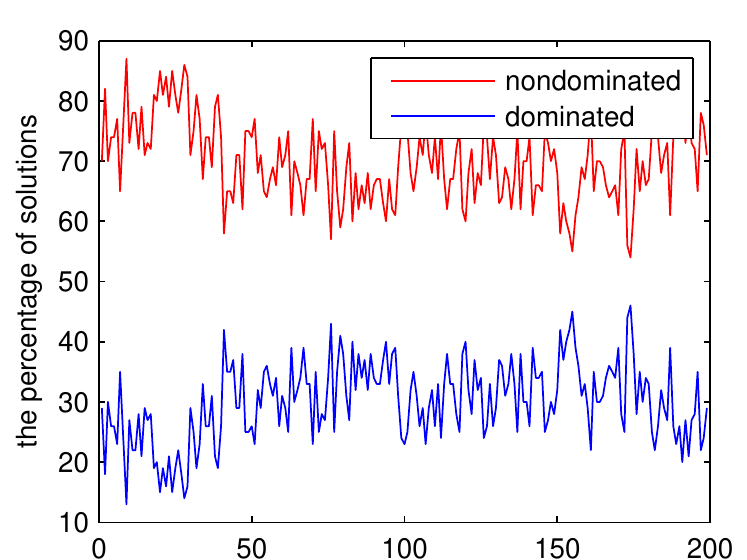}
    \subcaption{ZZJ6}
\end{subfigure}
\begin{subfigure}[t]{0.38\columnwidth}
    \includegraphics[ width=\columnwidth]{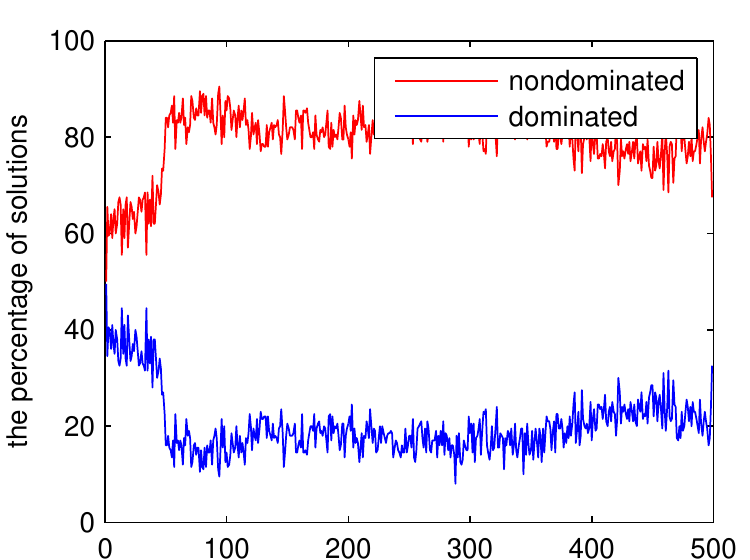}
    \subcaption{ZZJ7}
\end{subfigure}
\begin{subfigure}[t]{0.38\columnwidth}
    \includegraphics[ width=\columnwidth]{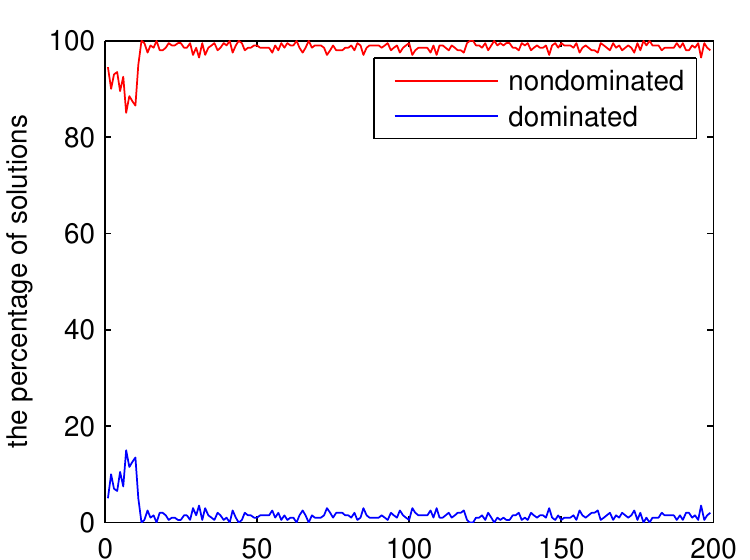}
    \subcaption{ZZJ8}
\end{subfigure}
\begin{subfigure}[t]{0.38\columnwidth}
    \includegraphics[ width=\columnwidth]{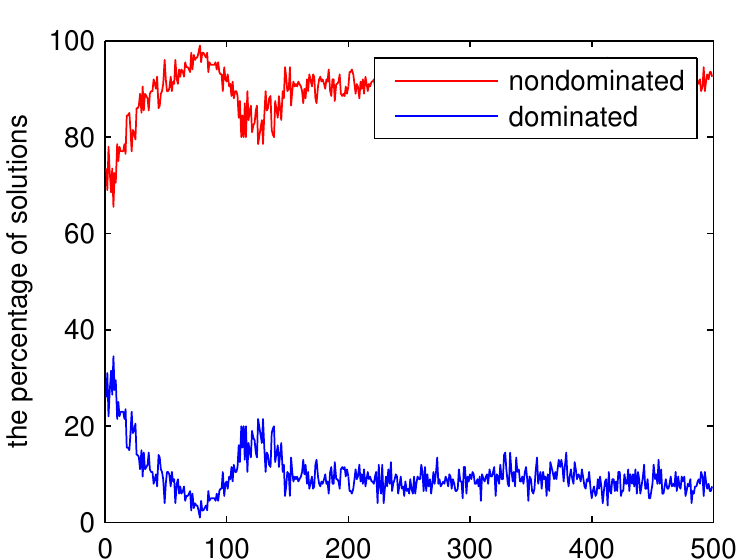}
    \subcaption{ZZJ9}
\end{subfigure}
\begin{subfigure}[t]{0.38\columnwidth}
    \includegraphics[ width=\columnwidth]{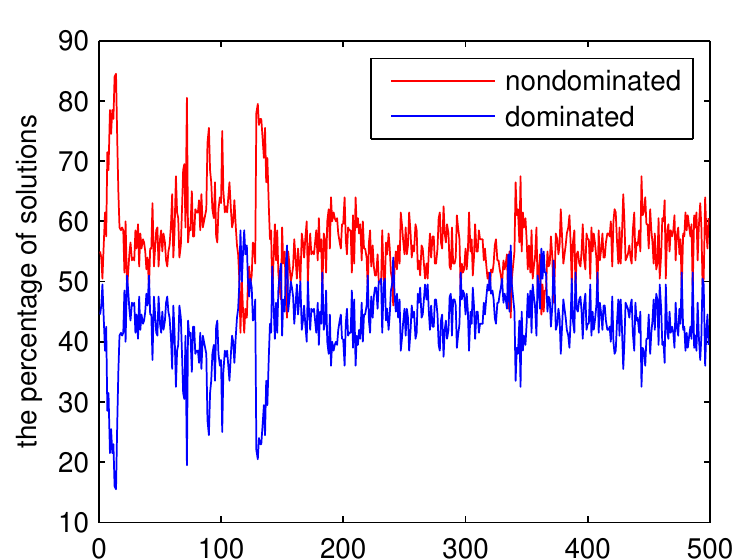}
    \subcaption{ZZJ10}
\end{subfigure}
\caption {The proportion of nondominated and dominated solutions selected by KNN in each generation for SMS-EMOA}
\label{fig:smsnd}
\end{figure*}

\begin{figure*}[htbp]
\centering
\begin{subfigure}[t]{0.38\columnwidth}
    \includegraphics[ width=\columnwidth]{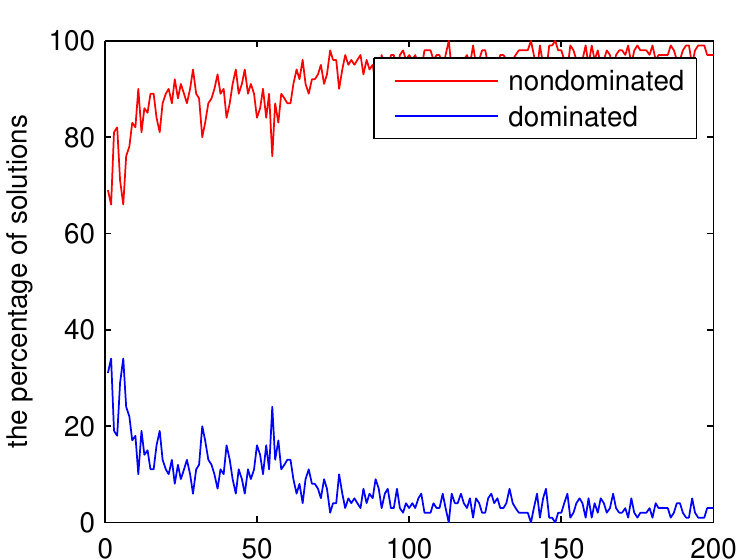}
    \subcaption{ZZJ1}
\end{subfigure}
\begin{subfigure}[t]{0.38\columnwidth}
    \includegraphics[ width=\columnwidth]{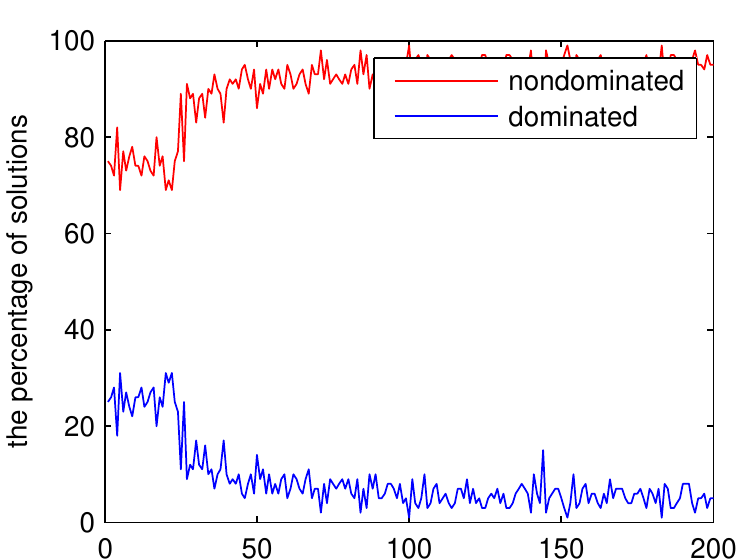}
    \subcaption{ZZJ2}
\end{subfigure}
\begin{subfigure}[t]{0.38\columnwidth}
    \includegraphics[ width=\columnwidth]{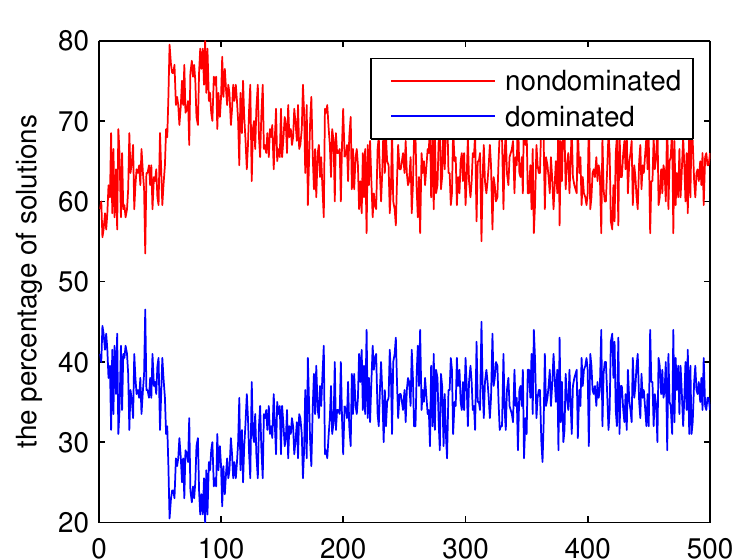}
    \subcaption{ZZJ3}
\end{subfigure}
\begin{subfigure}[t]{0.38\columnwidth}
    \includegraphics[ width=\columnwidth]{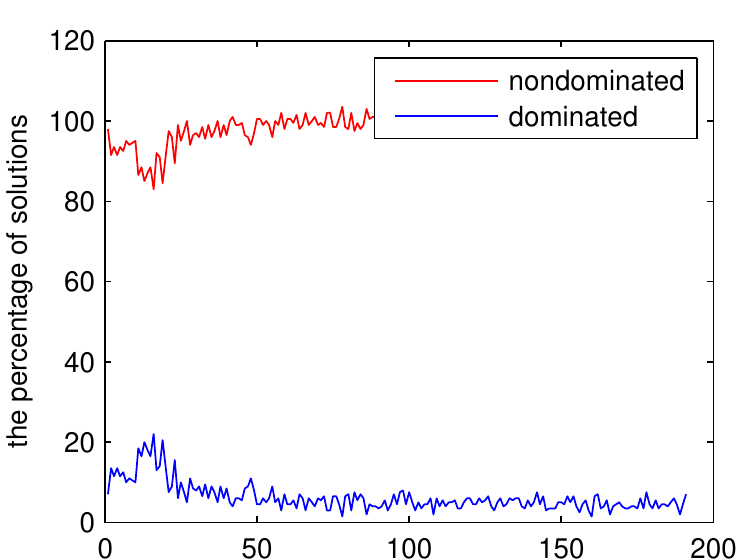}
    \subcaption{ZZJ4}
\end{subfigure}
\begin{subfigure}[t]{0.38\columnwidth}
    \includegraphics[ width=\columnwidth]{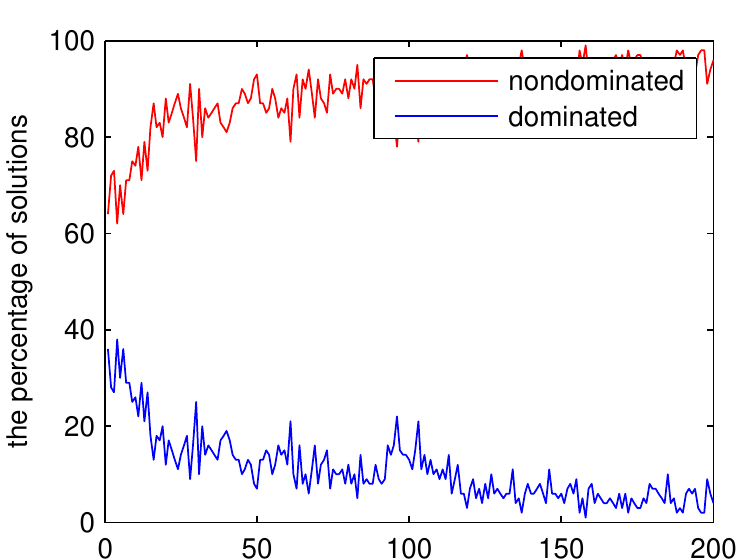}
    \subcaption{ZZJ5}
\end{subfigure}
\begin{subfigure}[t]{0.38\columnwidth}
    \includegraphics[ width=\columnwidth]{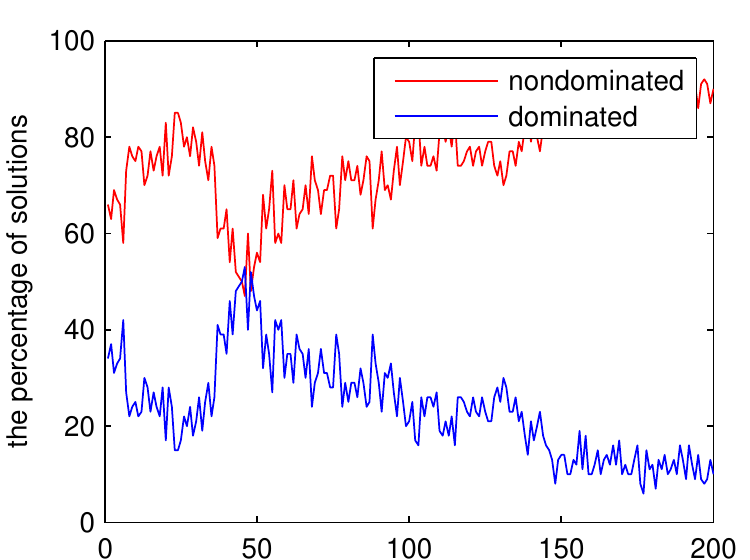}
    \subcaption{ZZJ6}
\end{subfigure}
\begin{subfigure}[t]{0.38\columnwidth}
    \includegraphics[ width=\columnwidth]{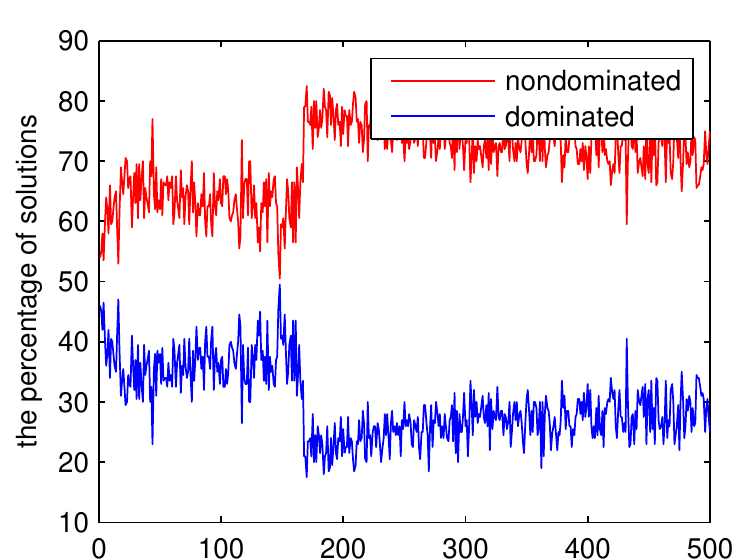}
    \subcaption{ZZJ7}
\end{subfigure}
\begin{subfigure}[t]{0.38\columnwidth}
    \includegraphics[ width=\columnwidth]{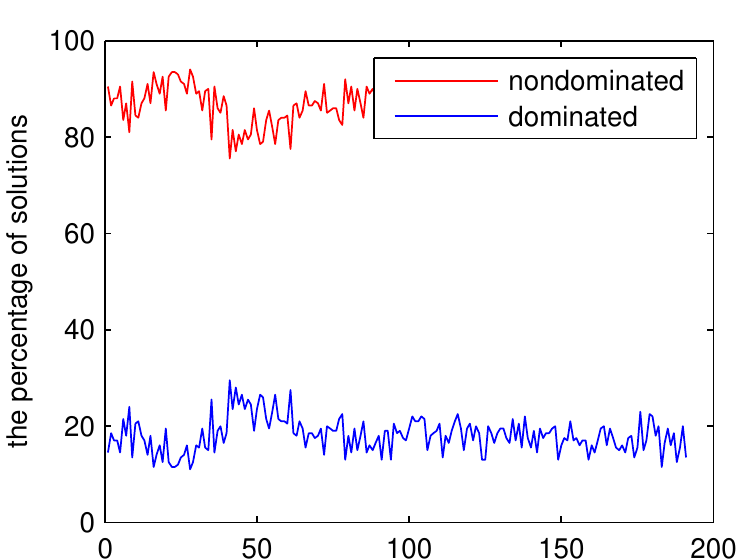}
    \subcaption{ZZJ8}
\end{subfigure}
\begin{subfigure}[t]{0.38\columnwidth}
    \includegraphics[ width=\columnwidth]{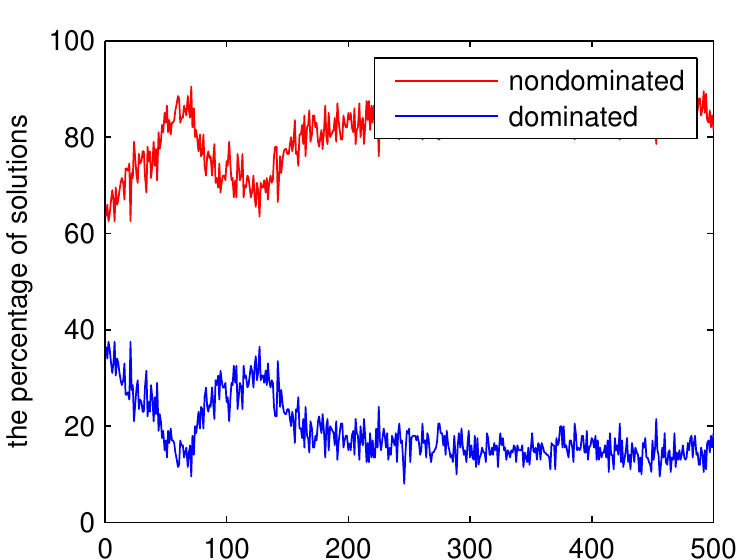}
    \subcaption{ZZJ9}
\end{subfigure}
\begin{subfigure}[t]{0.38\columnwidth}
    \includegraphics[ width=\columnwidth]{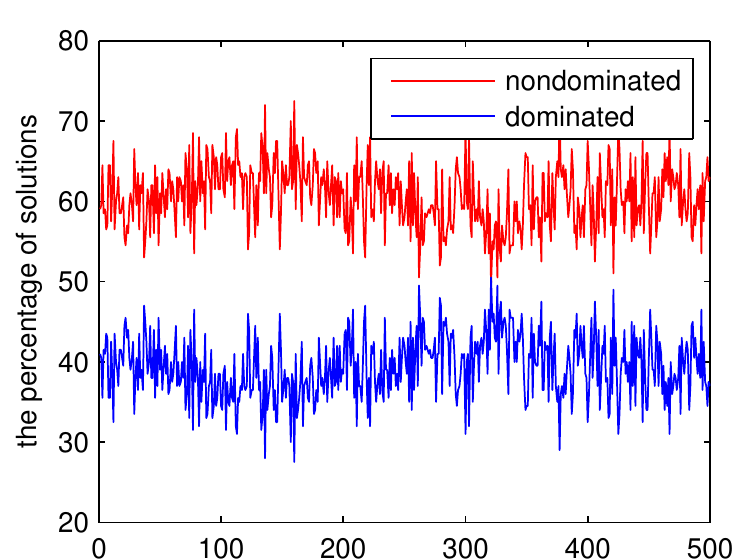}
    \subcaption{ZZJ10}
\end{subfigure}
\caption {The proportion of nondominated and dominated solutions selected by KNN in each generation for MOEA/D-MO}
\label{fig:moeadnd}
\end{figure*}

We conduct the following experiment. For each parent, $M=3$ candidate offspring solutions are generated and evaluated, and their domination relationships are also calculated. CPS is used to select an offspring solution. The proportion of offspring solutions, which selected by CPS are nondominated ones and dominated ones for each parent, are recorded for each generation. The statistical results are shown in Fig.~\ref{fig:rmnd}, Fig.~\ref{fig:smsnd} and Fig.~\ref{fig:moeadnd}.

From Fig.~\ref{fig:rmnd}, Fig.~\ref{fig:smsnd}, Fig.~\ref{fig:moeadnd}, we can see that for RM-MEDA, SMS-EMOA and MOEA/D-MO, $60.00\%$-$100.00\%$ offspring solutions chosen by CPS are nondominated ones. In most cases, the proportion is during $80.00\%$-$100.00\%$. And among them, at the beginning of the generation, the proportion of nondominated solutions and dominated solutions are close. Along the increment of generation, the proportions of the nondominated solutions are increased while the dominated ones are decreased. This suggests that statistically, CPS has a good ability to obtain good solutions among the offsprings of for each parent. It shows that CPS can guide the search without estimating the objective values of the candidate offspring solutions. The reason might be that the classification model can successfully detect the boundary between the positive and negative training sets, and thus to eliminate bad candidate offspring solutions without function evaluations.

\subsection{Sensitivity to Control Parameters}

\subsubsection{Sensitivity of the size of $P_{+}$ and $P_{-}$}
\begin{figure}[htbp]
 \centering
\includegraphics[ width=\columnwidth, height=5cm]{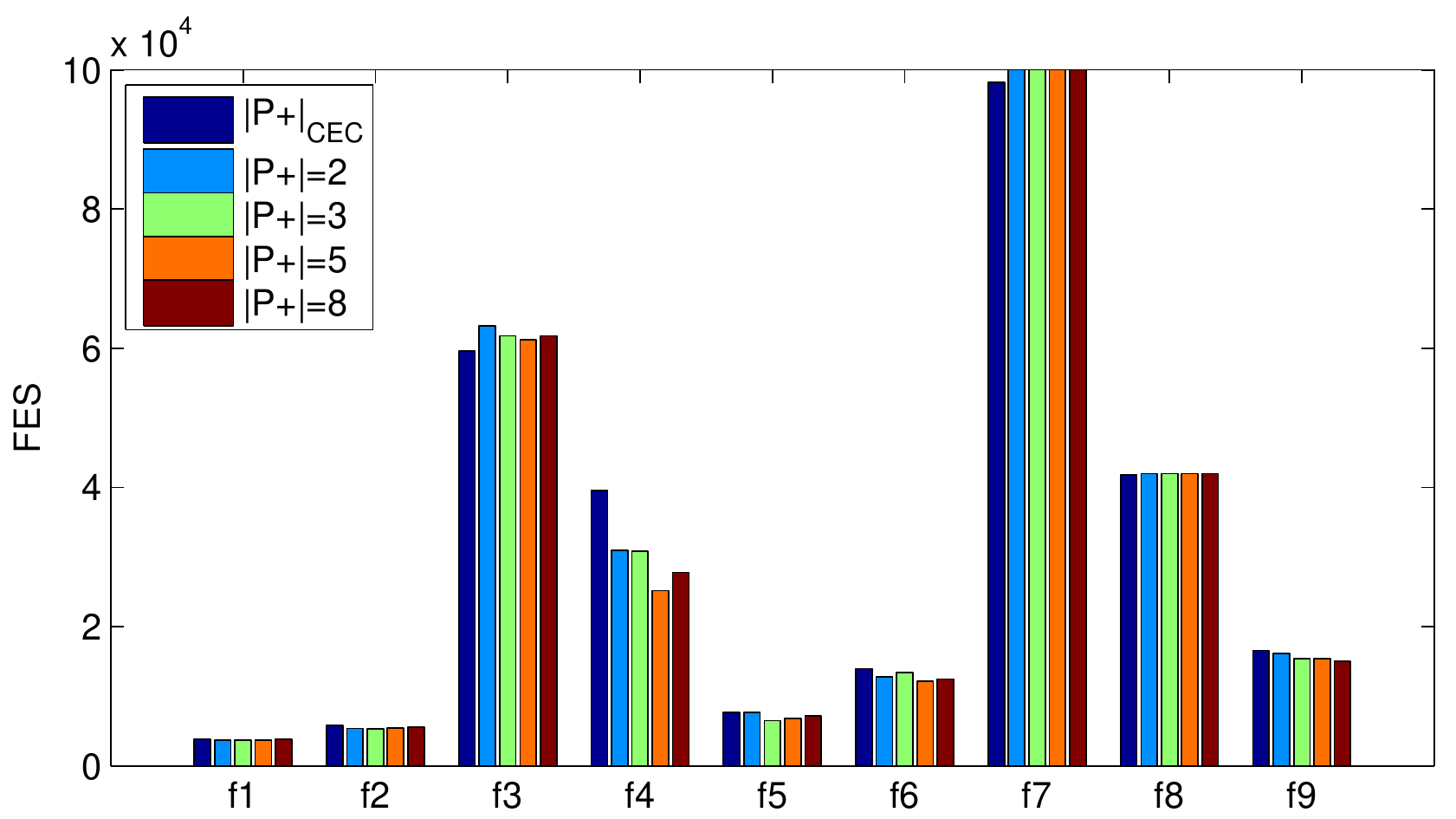}
\caption {The influence of the size of $P_{+}$ and $P_{-}$ on $ZZJ1-ZZJ9$}
\label{fig:gs}
\end{figure}

This section studies the influence of the size of $P_{+}$ and $P_{-}$. $|P_+|=|P_-|=2N$, $3N$, $5N$, $8N$, and $10N$ are studied. The data preparation strategy in our previous work ~\cite{2015ZhangZZ}, denoted as $|P_+|_{CEC}$~\footnote{$|P_+|=|P_-|=0.5N$.}, is also compared here. RM-MEDA is used as the basic optimizer and all the other parameters are the same as in Section~\ref{sec41}. The mean FEs that required to achieve $IGD=5\times 10^{-2}$ have been recorded and are shown in Fig.~\ref{fig:gs}.

The statistical results suggest that RM-MEDA-CPS with all the size values perform similarly on $ZZJ1$ and $ZZJ8$; RM-MEDA-CPS with $|P_+|=|P_-|=5N$ works the best on $ZZJ4$ and $ZZJ6$, and on $ZZJ3, ZZJ5, ZZJ9$, it works the second best. Our previous strategy $|P_+|_{CEC}$~\cite{2015ZhangZZ} performs best on $ZZJ3$ and $ZZJ7$, but it also achieves the worst results on $ZZJ2$, $ZZJ4$, $ZZJ5$, $ZZJ6$ and $ZZJ9$. These results show that CPS with $|P_+|=|P_-|=5N$ performs the best in most cases.

\subsubsection{Sensitivity of the number of $M$}
\begin{figure*}[htbp]
\centering
\begin{subfigure}[t]{0.38\columnwidth}
    \includegraphics[ width=\columnwidth]{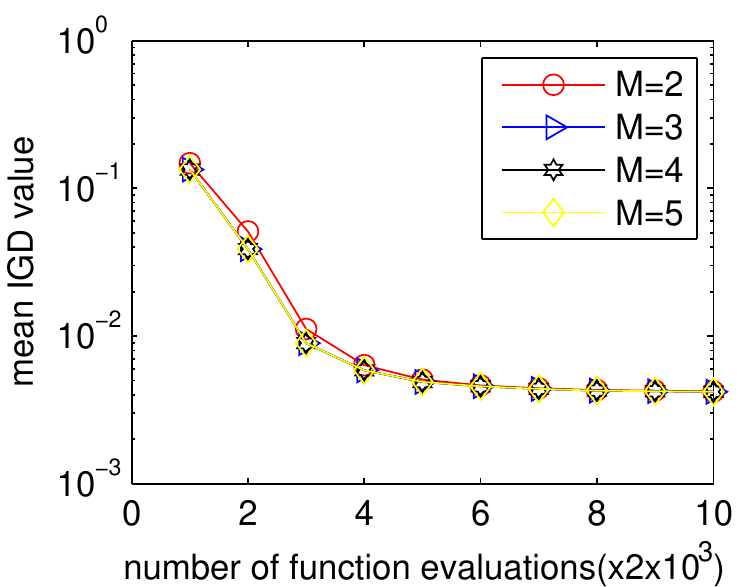}
    \subcaption{ZZJ1}
\end{subfigure}
\begin{subfigure}[t]{0.38\columnwidth}
    \includegraphics[ width=\columnwidth]{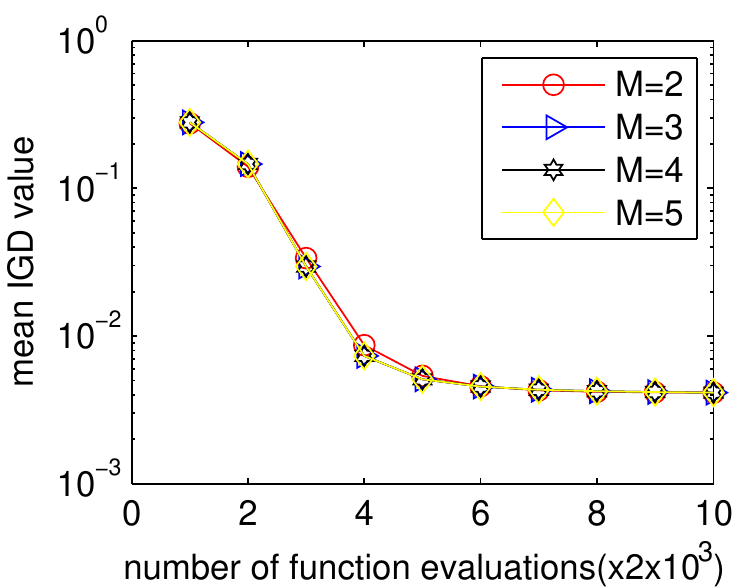}
    \subcaption{ZZJ2}
\end{subfigure}
\begin{subfigure}[t]{0.38\columnwidth}
    \includegraphics[ width=\columnwidth]{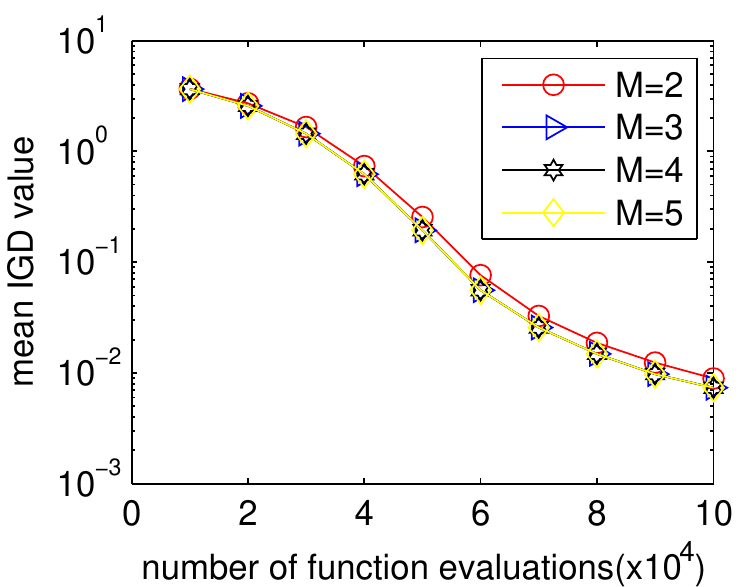}
    \subcaption{ZZJ3}
\end{subfigure}
\begin{subfigure}[t]{0.38\columnwidth}
    \includegraphics[ width=\columnwidth]{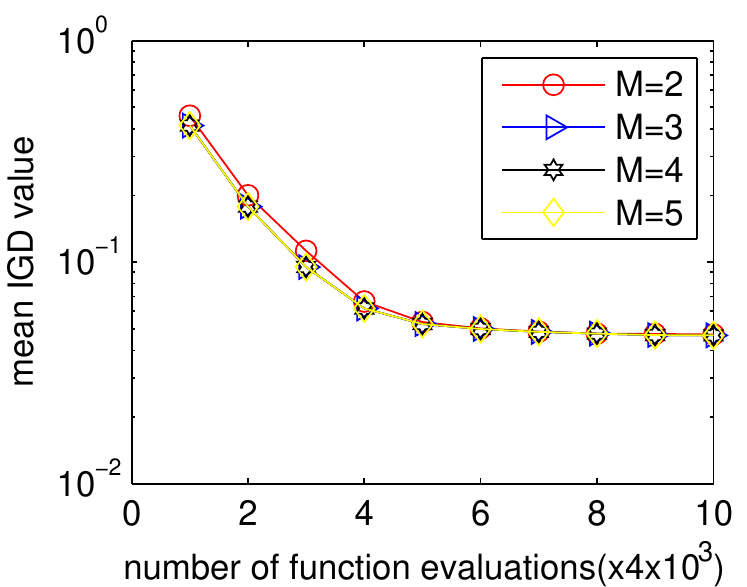}
    \subcaption{ZZJ4}
\end{subfigure}
\begin{subfigure}[t]{0.38\columnwidth}
    \includegraphics[ width=\columnwidth]{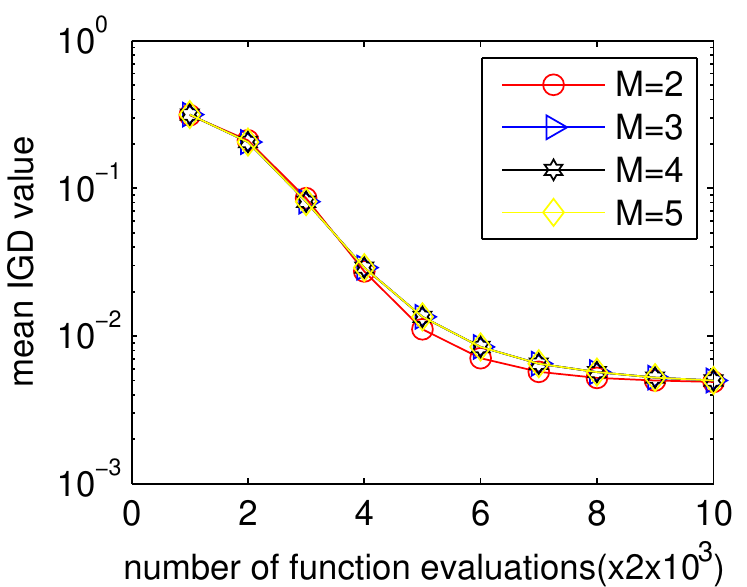}
    \subcaption{ZZJ5}
\end{subfigure}
\begin{subfigure}[t]{0.38\columnwidth}
    \includegraphics[ width=\columnwidth]{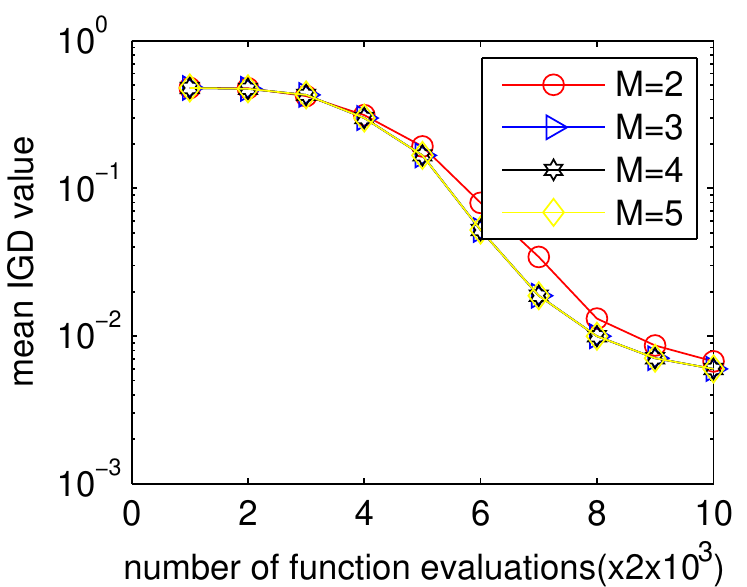}
    \subcaption{ZZJ6}
\end{subfigure}
\begin{subfigure}[t]{0.38\columnwidth}
    \includegraphics[ width=\columnwidth]{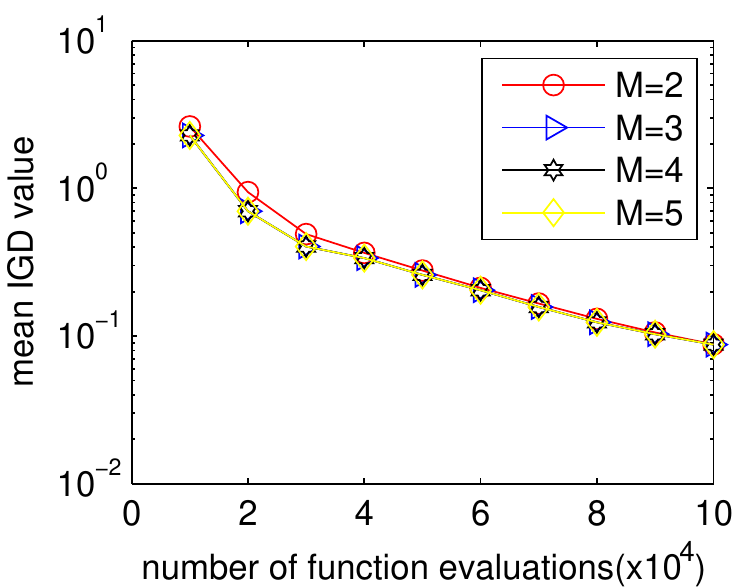}
    \subcaption{ZZJ7}
\end{subfigure}
\begin{subfigure}[t]{0.38\columnwidth}
    \includegraphics[ width=\columnwidth]{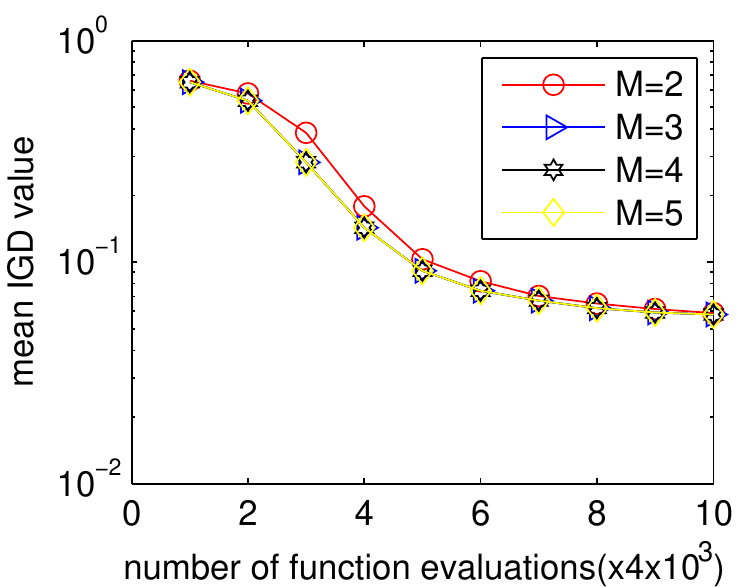}
    \subcaption{ZZJ8}
\end{subfigure}
\begin{subfigure}[t]{0.38\columnwidth}
    \includegraphics[ width=\columnwidth]{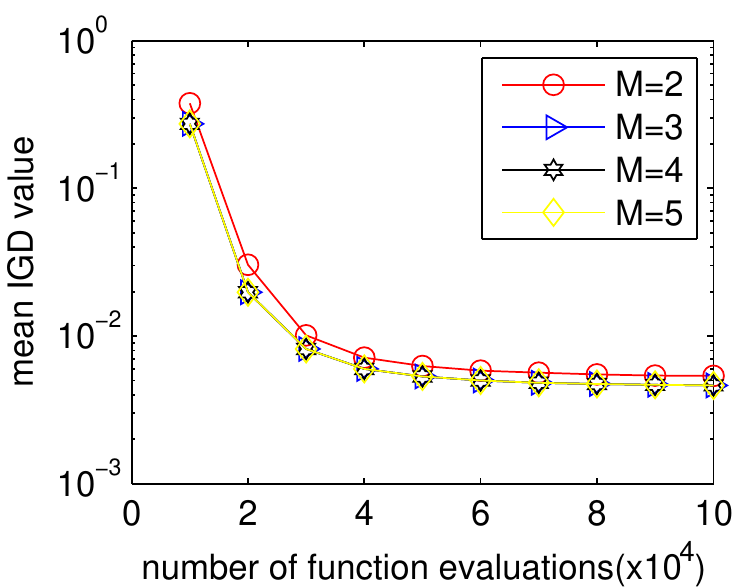}
    \subcaption{ZZJ9}
\end{subfigure}
\begin{subfigure}[t]{0.38\columnwidth}
    \includegraphics[ width=\columnwidth]{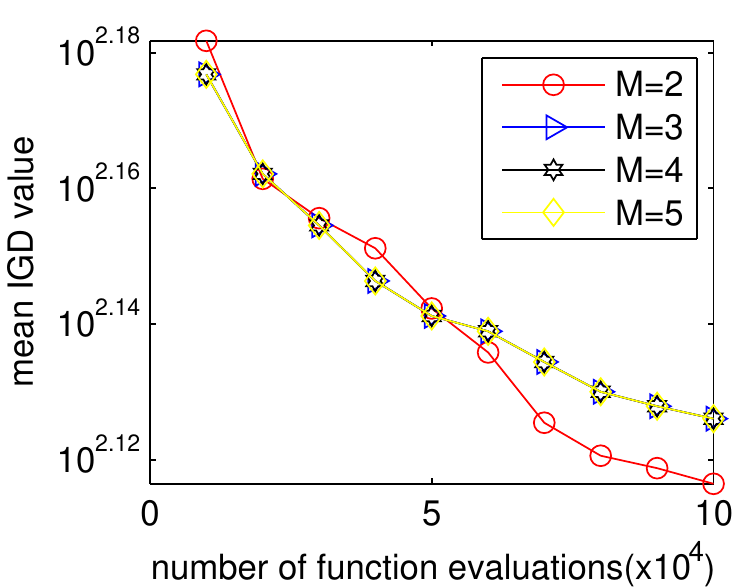}
    \subcaption{ZZJ10}
\end{subfigure}
\caption {The IGD value versus FEs obtained by RM-MEDA with different values of $M$ over 30 runs}
\label{fig:mn}
\end{figure*}

This section studies the influence of the number of candidate offspring solutions. RM-MEDA-CPS with $M=2, 3, 4, 5$ is employed to the test suite. All the other parameters are the same as in Section~\ref{sec41}. The IGD values versus FEs have been recorded and are shown in Fig.~\ref{fig:mn}.

The statistical results suggest that RM-MEDA-CPS with $M=2$ does not work well as RM-MEDA-CPS with other values. Furthermore, RM-MEDA-CPS with $M=3, 4, 5$ works similarly on all of the instances. This results show that the CPS strategy is not sensitive to $M$. However, considering the computational cost required by offspring generation, $M=3$ might be a proper choice in practice.

\section{Experimental Results on the LZ Test Suite}

\begin{table*}[htbp]
\scriptsize
\centering \caption{The statistical results of IGD and $I^-_H$ metric values obtained by RM-MEDA-CPS and RM-MEDA, MOEA/D-MO-CPS and MOEA/D-MO, SMS-EMOA-CPS and SMS-EMOA on LZ1-LZ9}\label{tab:lz}
\begin{tabular}{l|c|cc|cc|cc}\hline\hline
instance&\multicolumn{1}{c|}{metric}&\multicolumn{1}{c}{RM-MEDA-CPS}&\multicolumn{1}{c|}{RM-MEDA}&\multicolumn{1}{c}{SMS-EMOA-CPS}&\multicolumn{1}{c}{SMS-EMOA}&\multicolumn{1}{c}{MOEA/D-MO-CPS}&\multicolumn{1}{c|}{MOEA/D-MO}\\
\hline
$LZ1$	&$IGD$&1.54e-03$_{2.93e-05}(+)$	&1.58e-03$_{3.23e-05}$	&1.36e-03$_{2.04e-05}(+)$	&1.45e-03$_{3.06e-05}$ &1.35e-03$_{8.45e-06}(+)$	&1.38e-03$_{1.46e-05}$		\\
&$I^-_H$&1.52e-03$_{7.34e-05}(+)$	&1.63e-03$_{9.00e-05}$	&1.13e-03$_{4.45e-05}(+)$	&1.36e-03$_{5.89e-05}$ &1.12e-03$_{2.61e-05}(+)$	&1.20e-03$_{3.89e-05}$		\\
$LZ2$	&$IGD$&4.16e-02$_{3.09e-03}(+)$	&4.66e-02$_{3.08e-03}$	&3.52e-02$_{3.81e-03}(+)$	&3.81e-02$_{3.72e-03}$ &3.70e-03$_{5.51e-04}(+)$	&4.62e-03$_{1.00e-03}$		\\
&$I^-_H$&7.66e-02$_{4.80e-03}(+)$	&8.45e-02$_{5.63e-03}$	&6.14e-02$_{7.01e-03}(+)$	&6.75e-02$_{6.96e-03}$ &9.00e-03$_{2.68e-03}(+)$	&1.10e-02$_{2.76e-03}$		\\
$LZ3$	&$IGD$&1.38e-02$_{1.99e-03}(+)$	&1.53e-02$_{1.67e-03}$	&2.96e-02$_{3.00e-03}(+)$	&3.76e-02$_{2.45e-03}$ &3.23e-03$_{8.14e-04}(+)$	&3.62e-03$_{8.17e-04}$		\\
&$I^-_H$&2.45e-02$_{4.39e-03}(+)$	&2.61e-02$_{3.11e-03}$	&4.32e-02$_{4.32e-03}(+)$	&5.89e-02$_{4.10e-03}$ &6.84e-03$_{2.21e-03}(\sim)$	&7.46e-03$_{2.34e-03}$		\\
$LZ4$	&$IGD$&2.20e-02$_{4.70e-03}(+)$	&2.50e-02$_{4.89e-03}$	&2.98e-02$_{3.20e-03}(+)$	&3.81e-02$_{2.35e-03}$ &3.58e-03$_{9.30e-04}(+)$	&4.93e-03$_{1.35e-03}$		\\
&$I^-_H$&3.56e-02$_{7.95e-03}(+)$	&4.19e-02$_{8.13e-03}$	&4.43e-02$_{7.07e-03}(+)$	&5.76e-02$_{3.73e-03}$ &7.43e-03$_{3.02e-03}(+)$	&9.94e-03$_{3.58e-03}$		\\
$LZ5$	&$IGD$&1.42e-02$_{2.13e-03}(\sim)$	&1.53e-02$_{3.06e-03}$	&2.85e-02$_{3.96e-03}(+)$	&3.51e-02$_{2.52e-03}$ &7.23e-03$_{9.81e-04}(\sim)$	&7.72e-03$_{1.09e-03}$		\\
&$I^-_H$&2.44e-02$_{3.99e-03}(\sim)$	&2.62e-02$_{4.67e-03}$	&4.11e-02$_{3.81e-03}(+)$	&5.28e-02$_{3.23e-03}$ &1.37e-02$_{2.80e-03}(\sim)$	&1.42e-02$_{2.94e-03}$		\\
$LZ6$	&$IGD$&1.64e-01$_{7.31e-02}(+)$	&2.08e-01$_{6.38e-02}$	&9.71e-02$_{6.67e-03}(+)$	&1.00e-01$_{4.61e-03}$ &5.84e-02$_{9.76e-03}(\sim)$	&5.96e-02$_{9.81e-03}$		\\
&$I^-_H$&3.35e-01$_{1.27e-01}(+)$	&4.15e-01$_{9.47e-02}$	&1.76e-01$_{5.86e-03}(+)$	&2.07e-01$_{8.16e-03}$ &1.26e-01$_{1.96e-02}(\sim)$	&1.30e-01$_{1.85e-02}$		\\
$LZ7$	&$IGD$&1.49e+00$_{3.09e-01}(\sim)$	&1.38e+00$_{4.46e-01}$	&2.87e-01$_{1.61e-02}(+)$	&3.17e-01$_{2.11e-02}$ &1.42e-01$_{1.04e-01}(\sim)$	&1.17e-01$_{1.07e-01}$		\\
&$I^-_H$&1.08e+00$_{1.11e-01}(\sim)$	&1.04e+00$_{1.58e-01}$	&4.68e-01$_{2.49e-02}(+)$	&5.37e-01$_{3.35e-02}$ &2.29e-01$_{1.84e-01}(\sim)$	&1.92e-01$_{1.77e-01}$		\\
$LZ8$	&$IGD$&1.61e-02$_{8.82e-03}(\sim)$	&2.23e-02$_{1.40e-02}$	&1.40e-01$_{2.56e-02}(+)$	&1.71e-01$_{1.76e-02}$ &1.40e-02$_{1.26e-02}(\sim)$	&1.50e-02$_{1.30e-02}$		\\
&$I^-_H$&5.55e-02$_{2.47e-02}(+)$	&7.12e-02$_{3.24e-02}$	&2.23e-01$_{3.63e-02}(+)$	&2.67e-01$_{2.59e-02}$ &2.39e-02$_{2.41e-02}(\sim)$	&2.50e-02$_{2.33e-02}$		\\
$LZ9$	&$IGD$&4.40e-02$_{5.53e-03}(\sim)$	&4.47e-02$_{3.77e-03}$	&3.89e-02$_{6.93e-03}(\sim)$	&4.18e-02$_{5.56e-03}$ &4.79e-03$_{1.07e-03}(+)$	&6.13e-03$_{1.42e-03}$		\\
&$I^-_H$&7.56e-02$_{7.67e-03}(+)$	&8.06e-02$_{6.13e-03}$	&6.60e-02$_{8.57e-03}(+)$	&7.57e-02$_{8.07e-03}$ &9.04e-03$_{2.42e-03}(+)$	&1.16e-02$_{2.60e-03}$		\\
\hline
$+/-/\sim$ &$IGD$	&5/0/4	& &8/0/1	& &5/0/4	&\\
$+/-/\sim$ &$I^-_H$ &7/0/2		& &9/0/0	 & &4/0/5	&\\
\hline\hline
\end{tabular}
\end{table*}

In this section, we apply the three algorithms and their CPS variants on the LZ test suite~\cite{2009LiZ}, of which the PSs have complicated shapes. The parameters are as follows. The number of decision variables is $n=30$ for all the test instances. The algorithms are executed $30$ times independently on each instance and stop after $150,000$ FEs on LZ1-LZ5 and LZ7-LZ9, and $297,500$ FEs on LZ6. The population size is set as $N=300$ on LZ1-LZ5 and LZ7-LZ9, and $595$ on LZ6 respectively. All the other parameters are the same as in Section~\ref{sec41}.

Table~\ref{tab:lz} shows the mean and std. IGD and $I^-_H$ metric values obtained by RM-MEDA-CPS, RM-MEDA, SMS-EMOA-CPS, SMS-EMOA, MOEA/D-MO-CPS and MOEA/D-MO on $LZ1-LZ9$ after 30 runs.

The experimental results in Table \ref{tab:lz} suggests that CPS is able to improve the performances of the original algorithms according to both the IGD and $I^-_H$ metric values. It is clear that the CPS based versions work no worse than the original versions on all of the test instances. More precisely according to the IGD metric, (a) RM-MEDA-CPS performs better than RM-MEDA on 5 instances, and on the other 4 instances, they work similarly; (b) SMS-EMOA-CPS achieves better results than SMS-EMOA does on 8 instances, and they achieve similar results on the other instance; and (c) MOEA/D-MO-CPS outperforms MOEA/D-MO on 5 instances, and they get similar results on the other 4 instances. According to the $I^-_H$ metric, (a) RM-MEDA-CPS wins on 7 instances, and on the other 2 instances, they work similar; (b) SMS-EMOA-CPS performs better than SMS-EMOA on all 9 instances; and (c) MOEA/D-MO-CPS outperforms MOEA/D-MO on 4 instances, and they get similar results on the other 4 instances.

%%%%%%%%%%%%%%%%%%%%%%%%%%%%%%%%%%%%%%%%%%%%%%%%%%%%%%
%% SEC5
%%%%%%%%%%%%%%%%%%%%%%%%%%%%%%%%%%%%%%%%%%%%%%%%%%%%%%
\section{Conclusion}
\label{sec5}

This paper proposes a \emph{classification based preselection (CPS)} to improve the performance of \emph{multiobjective evolutionary algorithms (MOEAs)}. In CPS based MOEAs, some solutions are chosen to form a training data set, and then a classifier is built based on the training data set in each generation. Each parent solution generates a set of candidate offspring solutions, and chooses a promising one as the real offspring solution based on the classifier.

The CPS strategy is applied to three types of MOEAs. The three algorithms and their original versions are empirically compared on two test suites. The experimental results suggest that the CPS can successfully improve the performance of the original algorithms.

There is still some work that is worth further investigating in the future. Firstly, the efficiency of CPS could be analyzed, secondly, more suitable data preparation strategies and more classification models should be tried, and thirdly, some strategies should be used to analyze CPS performance.

%\bibliography{20170424_CPS}

\end{document}